\documentclass{article} %
\PassOptionsToPackage{varqu}{inconsolata}
\usepackage{colm2024_conference}

\usepackage{booktabs}
\usepackage{arydshln}

\usepackage{graphicx}
\usepackage{enumitem}
\usepackage{wrapfig}
\usepackage{algorithm}
\usepackage{algpseudocode}

\usepackage{microtype}
\usepackage{amsmath}
\usepackage{amssymb}
\usepackage{colortbl}
\usepackage[utf8]{inputenc}
\usepackage[T1]{fontenc}
\newcommand{\dq}{{\fontencoding{T1}\fontfamily{cmtt}\selectfont\char34}}
\newcommand{\sq}{{\fontencoding{T1}\fontfamily{cmtt}\selectfont\char39}}
\definecolor{lightgray}{rgb}{0.9,0.9,0.9}
\usepackage{caption}
\usepackage{subcaption}
\usepackage{xcolor}
\usepackage{setspace}
\usepackage{url}
\usepackage{multirow}
\usepackage{tabularx}
\usepackage{pgfplots}
\pgfplotsset{compat=1.18}
\usepackage{tikz}
\usetikzlibrary{er,positioning,bayesnet}
\usepackage{makecell}
\usepackage{tipa}
\usepackage{siunitx}
\usepackage{nicefrac}
\usepackage{tocloft}
\usepackage{listings}
\usepackage[raster,skins]{tcolorbox}
\usepackage{xltabular}
\usepackage{adjustbox}
\usepackage{xurl}
\usepackage{xspace}

\usepackage{amsmath,amsfonts,bm}

\def\eqref#1{equation~\ref{#1}}

\def\1{\bm{1}}

\DeclareMathAlphabet{\mathsfit}{\encodingdefault}{\sfdefault}{m}{sl}
\SetMathAlphabet{\mathsfit}{bold}{\encodingdefault}{\sfdefault}{bx}{n}

\lstdefinestyle{domainex}{%
  basicstyle=\ttfamily\fontsize{5.2pt}{6.2pt}\selectfont,
  breaklines=true,
  breakatwhitespace=false,
  columns=fullflexible,
  frame=none,
  aboveskip=0pt,
  belowskip=0pt,
  xleftmargin=0pt,
  xrightmargin=0pt,
  keepspaces=true,
  showstringspaces=false,
}

\newcommand{\method}{Qwen-AgentWorld\xspace}
\newcommand{\bench}{AgentWorldBench\xspace}
\colorlet{colorscale}{cyan!40!red}
\colorlet{colorctrl}{cyan!40!black}
\definecolor{colordecouple}{HTML}{d2e2f0}
\definecolor{colorunify}{HTML}{dad8e8}
\definecolor{lightblue}{RGB}{220,235,250}
\tcbset{
  takeawaysbox/.style={
    title=Takeaways,
    colback=lightblue!80,
    colframe=black,
    fonttitle=\bfseries\small,
    coltitle=white,
    colbacktitle=black,
    enhanced,
    attach boxed title to top left={xshift=2.5mm,yshift=-2.5mm},
    boxed title style={rounded corners, size=small, colframe=black, colback=black},
    width=\linewidth,
    arc=3.5mm
  }
}

\title{Qwen-AgentWorld:\\ Language World Models for General Agents}

\author{
\bf Qwen Team
}

\begin{document}

\maketitle

\begin{abstract}

A world model predicts environment dynamics based on current observations and actions, serving as a core cognitive mechanism for reasoning and planning.
In this work, we investigate how world modeling based on language models can further push the boundaries of general agents.
(i) We first focus on building foundation models for agentic environment simulation.
We introduce \textbf{\method-35B-A3B} and \textbf{\method-397B-A17B}, the first language world models capable of simulating agentic environments covering 7 domains via long chain-of-thought reasoning.
Leveraging more than 10M environment interaction trajectories of 7 domains in real-world environments, we develop \method through a three-stage training pipeline:
CPT injects general-purpose world modeling capabilities from the state transition dynamics and augmented professional corpora, SFT activates next-state-prediction reasoning, and RL sharpens simulation fidelity through a tailored framework with hybrid rubric-and-rule rewards.
To evaluate language world models, we present \textbf{\bench}, a comprehensive benchmark constructed from real-world interactions of 5 frontier models on 9 established benchmarks, such as Tool Decathlon, Terminal-Bench 1.0 \& 2.0, and OSWorld-Verified, which evaluates world modeling quality through ground-truth grounded rubric judging across 5 dimensions.
Empirical results demonstrate that \method significantly outperforms existing frontier models.
(ii) Beyond foundation models, we further investigate two complementary paradigms through which world modeling enhances general agents.
First, as a \emph{decoupled} environment simulator, \method supports scalable and controllable simulation of thousands of real-world environments for agentic RL, yielding gains that surpass real-environment training alone.
Second, as a \emph{unified} agent foundation model, world-model training acts as a highly effective warm-up that improves downstream performance across 7 agentic benchmarks.

\end{abstract}

\vfill

\begin{figure}[hbp]
    \vspace{-4mm}
    \centering
    \includegraphics[width=.96\linewidth]{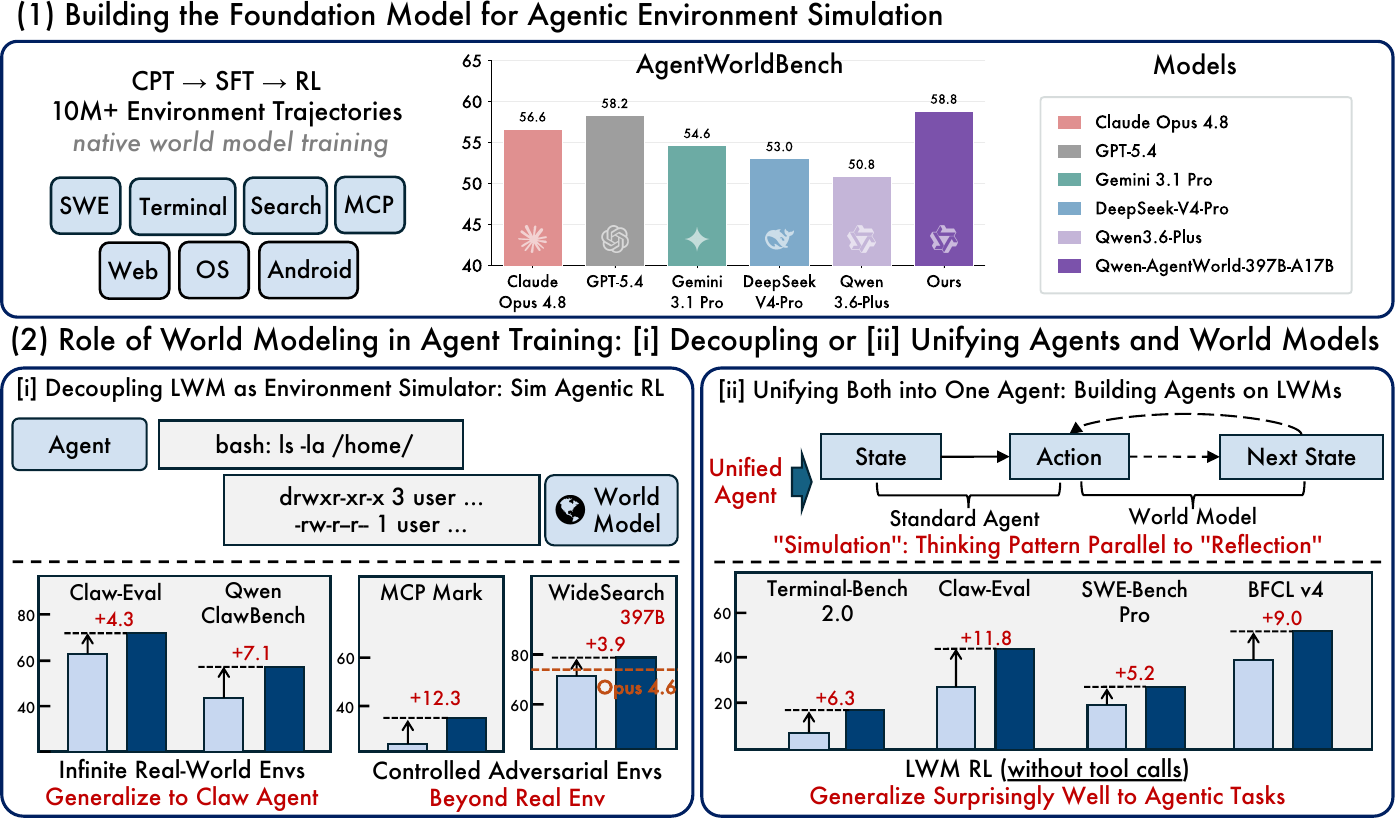}
    \caption{
      Overview of \method.
      \textbf{Top:} \method is a unified \emph{native} language world model across seven domains.
      \textbf{Bottom:} We explore two complementary strategies for applying world modeling to enhance language agents (mainly using the 35B-A3B model as agent):
      {\fboxsep=1.5pt\colorbox{colordecouple}{\textbf{Decouple}}} and {\fboxsep=1.5pt\colorbox{colorunify}{\textbf{Unify}}},
      where the world model serves as the environment simulator and agent foundation model, respectively. 
      }
    \label{fig:teaser}
\end{figure}

\vfill

\newpage

\makeatletter
\def\addcontentsline#1#2#3{%
  \addtocontents{#1}{\protect\contentsline{#2}{#3}{\thepage}{\@currentHref}}}
\makeatother
\setcounter{tocdepth}{3}
\setcounter{secnumdepth}{3}
\tableofcontents
\newpage

\section{Introduction}
\label{sec:intro}

World models have been widely recognized as a foundation toward general intelligence~\citep{genie3, worldlabs2025marble, xiang2025pan, ali2025world}, with a growing consensus that learning to predict the world is prerequisite to acting effectively within it~\citep{lecun2022path, hafner2023mastering, hafner2025training, assran2025v}.
\citet{richens2025general} further prove a stronger claim: any agent capable of generalizing across a sufficiently broad range of tasks must have learned a world model, establishing world models not merely as useful but as necessary for general-purpose agents.

\looseness=-1 Yet the language environments in which LLM agents operate still lack a general-purpose world model.
In the agent–environment interaction loop, two complementary components are essential: the policy~(states $\rightarrow$ actions) and the world model~((states, actions) $\rightarrow$ subsequent states).
However, current research on LLM agents has focused almost exclusively on the policy side.
We argue that world modeling is a crucial missing piece in the path to general agents.
\textbf{In this work, we explore both how to achieve language world modeling and how to apply it to advance general agents.}
We first study world modeling in language models to devlop a foundation model, the Language World Model (LWM), for agentic environment simulation.
We then investigate how world modeling can improve general agents through two complementary paradigms: either decoupling the agent from the world model or unifying them into a single framework.

\definecolor{boxbrown}{HTML}{862D1A}
\definecolor{boxwarm}{HTML}{F7F4ED}
\begin{tcolorbox}[
  enhanced,
  colback=boxwarm!30,
  colframe=boxbrown!80,
  colbacktitle=boxwarm,
  coltitle=boxbrown!90!black,
  fonttitle=\small,
  title={\textbf{Why Language World Models When Real Environments Exist?}\\{Not for Cost Reduction, but as a Complementary Axis for Pushing the Frontier}},
  boxrule=0.5pt,
  arc=1.5pt,
  left=8pt, right=8pt, top=4pt, bottom=4pt,
  toptitle=3pt, bottomtitle=3pt,
]
\textbf{(1) Decoupling:} Using the world model as a simulator facilitates turn-level scalability and controllability.
(i) Scalability:
LWM enables turn-level scaling of diverse environments without requiring dedicated infrastructure (e.g., sandboxes or GUI virtual machines), spanning extreme scenarios, real-world tasks~\citep{openclaw2026,patwardhan2025gdpval}, and high-value professional domains where real execution is infeasible due to irreversible operations, proprietary deployments, or the absence of public implementations.
(ii) Controllability:
LWM offers precise controllability, enabling more diverse and challenging environments that systematically expose agent weaknesses through targeted perturbations rare or absent in real environments.
For instance, it can return partial results that force the agent to take additional interaction steps to retrieve the complete information.
Training against these targeted perturbations helps agents handle edge cases that real-environment training alone cannot cover, ultimately surpassing agents trained solely in real environments (\S\ref{sec:app:controllable}).

\smallskip
\textbf{(2) Unifying:} A capable general agent should possess both decision-making and world-modeling abilities.
World modeling serves as a foundation for stronger agents, as it enables agents to predict future states to refine action selection, whereas traditional agent training has focused only on state-to-action decision-making.
Intuitively, an agent capable of predicting environment feedback prior to committing to an action can in principle perform no worse than its counterpart lacking such capacity~\citep{richens2025general}.
Next state prediction can thus be internalized as a meta-level thinking pattern similar to ``reflection'' but oriented toward the future (\S\ref{sec:app:foundation}).
Furthermore, accurate next-state prediction requires reasoning, knowledge, instruction following, and long-context handling (Appendix~\ref{sec:analysis:reasoning}), capabilities that are themselves foundational to general agents.
\end{tcolorbox}

We present \textbf{\method}, the first language world model that simulates seven agent environments through long chain-of-thought reasoning: MCP, Search, Terminal, Software Engineering, Android, Web, and OS.
For the three GUI domains, environment observations are represented as accessibility trees and UI view hierarchies rather than pixel frames.
\method is a \emph{native} world model trained through three stages: CPT injects state-transition dynamics and world knowledge, SFT activates next-state-prediction thinking patterns, and RL with hybrid rubric-and-rule rewards sharpens simulation fidelity.
To evaluate LWM, we construct \textbf{\bench}, a comprehensive benchmark across all seven domains.
The benchmark is built from real environment interactions of frontier models such as Claude Opus 4.6 on widely used agent benchmarks such as Terminal-Bench 1.0 \& 2.0~\citep{merrill2026terminal} and OSWorld-Verified~\citep{xie2024osworld} ensuring entirely out-of-distribution evaluation.
\bench evaluates simulation quality through open-ended rubric judging across five dimensions.
We additionally design rule-based verifiers for deterministic checks on targeted simulation capabilities.
Empirical evaluations demonstrate that \method achieves superior performance over existing frontier models.
We further investigate two complementary paradigms by which world modeling improves general agents:

\begin{itemize}[leftmargin=1.5em,itemsep=2pt,topsep=0pt]

\item \textbf{Environment Simulator~(\S\ref{sec:app:simulator}).}
We demonstrate that \method can simulate $4k$ real-world OpenClaw environments for agentic RL, yielding gains on Claw-Eval~\citep{ye2026claw} and QwenClawBench~\citep{qwenclawbench}.
Moreover, the controllability provides significant advantages complementary to real-world interaction, leading to substantial gains on Tool Decathlon~\citep{li2025tool}, MCPMark~\citep{wu2025mcpmark}, and WideSearch~\citep{wong2025widesearch}.

\item \textbf{Agent Foundation Model~(\S\ref{sec:app:foundation}).}
Comprehensive experiments on Terminal-Bench 2.0~\citep{merrill2026terminal}, SWE-Bench Verified~\citep{jimenez2024swe}, SWE-Bench Pro~\citep{deng2025swe}, BFCL~v4~\citep{patil2025berkeley}, Claw-Eval~\citep{ye2026claw}, QwenClawBench~\citep{qwenclawbench}, and WideSearch~\citep{wong2025widesearch} demonstrate that LWM training serves as a warm-up or auxiliary training stage.
It acquaints agents with environment dynamics and next-state prediction before downstream agentic RL, thereby providing a critical foundation for bootstrapping stronger agent performance.

\end{itemize}

\begin{figure}[t]
    \centering
    \includegraphics[width=\linewidth]{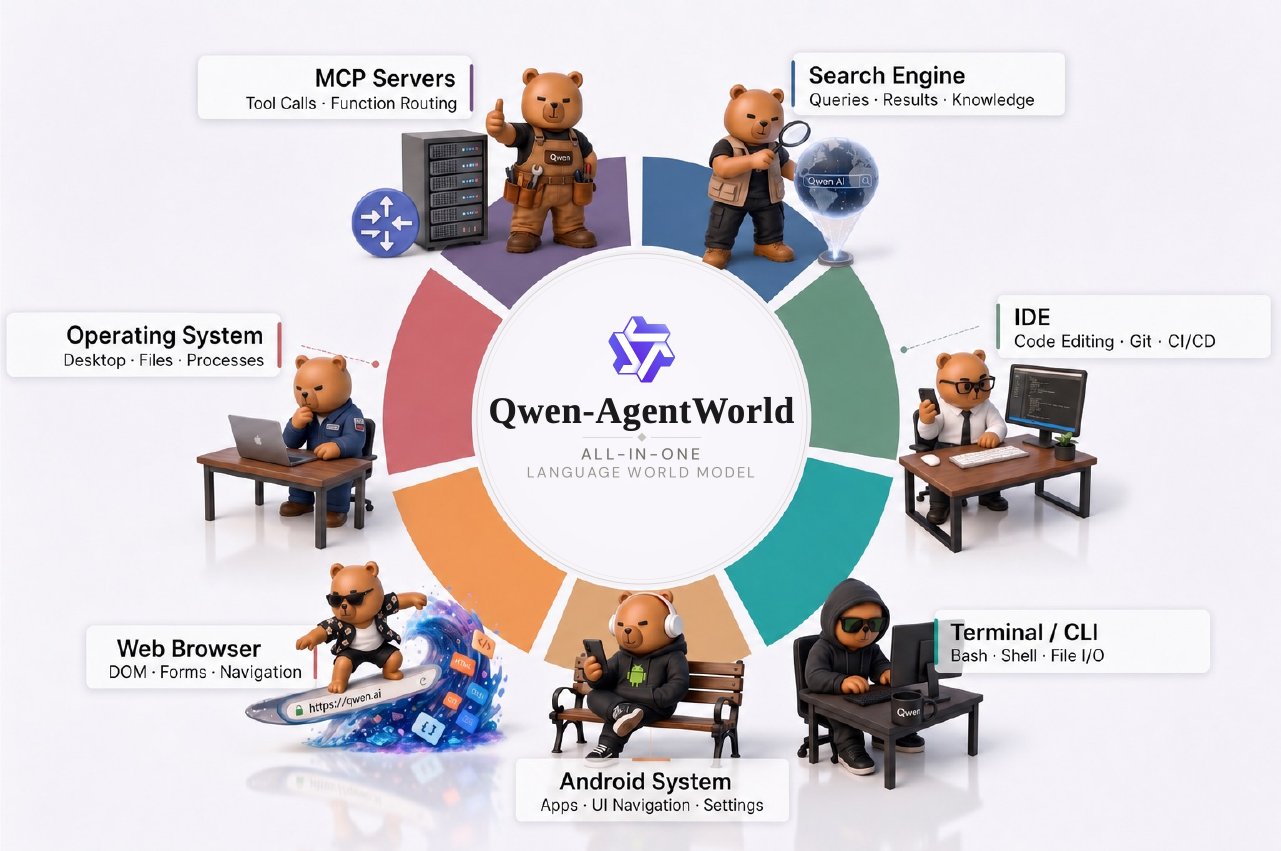}
    \caption{\method unifies seven categories of interactive environment simulation within a single language world model.
    }
    \label{fig:all_in_one}
\end{figure}

\section{Preliminaries}
\label{sec:prelim}

This section formalizes the language world model (LWM) studied throughout this work.
We train the world model based on language models using the broad world-knowledge corpora, and unify seven domains under a shared textual representation that enables cross-domain generalization for language world modeling.
We begin by establishing terminology (\S\ref{sec:prelim:terminology}), then present the unified trajectory schema shared across seven domains and training stages (\S\ref{sec:prelim:schema}) and formalize the world modeling objective (\S\ref{sec:prelim:formulation}).

\subsection{Terminology}
\label{sec:prelim:terminology}

We use the following terms consistently throughout this report.
\textbf{LWM training} refers to training the world model itself through three stages: continual pre-training (\textbf{LWM CPT}, \S\ref{sec:pipeline:cpt}), supervised fine-tuning (\textbf{LWM SFT}, \S\ref{sec:pipeline:sft}), and reinforcement learning (\textbf{LWM RL}, \S\ref{sec:pipeline:rl}).
Separately, \textbf{Sim~RL} trains a policy agent via RL using a LWM as an environment simulator (\S\ref{sec:app:simulator}), while \textbf{Real~RL} trains the same agent against a live, real-world environment (e.g., an actual search engine or a running terminal).

Throughout this report, \textbf{trajectory} refers to an \textbf{environment trajectory}, which is structurally a multi-turn dialogue between the agent and the environment, represented as a sequence of (action, observation) pairs.
In contrast, an \textbf{agentic trajectory} is the agent's single completion for a task, a multi-step tool-integrated reasoning trace that interleaves the agent's internal thinking and action selection with the environment's observations.
Environment trajectories can be extracted from agentic trajectories by stripping agent reasoning and retaining only (action, observation) pairs, or collected directly from raw interaction logs.

\begin{figure*}[t]
\centering
\begin{tcolorbox}[
  colback=gray!3, colframe=black!50, boxrule=0.5pt,
  title={\small System Prompt --- Terminal LWM RL},
  fonttitle=\bfseries\small,
  left=4pt, right=4pt, top=2pt, bottom=2pt
]
\scriptsize
\textbf{\color{blue!70!black}[1] Task Description} \hfill \textit{static}\\[2pt]
You are a \textbf{Terminal World Model} --- a precise terminal state simulator. Your task is to predict the exact output of a Linux/Unix terminal after executing a given command or sequence of commands. Your goal is to be as faithful as possible to real terminal behavior while maintaining consistency and logical correctness across the interaction sequence.\\[1pt]
\textit{Given:} (1)~Historical Context (optional); (2)~Current Terminal State; (3)~User Action (keystrokes).\\\textit{Predict:} the exact next terminal state after all actions are executed.\\[1pt]
Core Responsibilities: State Prediction~$\mid$~Context Maintenance~$\mid$~Behavioral Fidelity\\[1pt]
{\color{gray}\textit{[\,\dots\; state transition rules, side-effect tracking, error handling, output formatting \dots\,]}}

\tcbline

\textbf{\color{blue!70!black}[2] Action Space} \hfill \textit{static}\\[2pt]
User actions are JSON arrays of command objects:~~{\ttfamily\char`\{``keystrokes'': ``ls -la\char`\\n'', ``duration'': 0.1\char`\}}\\[1pt]
Keystrokes: {\ttfamily\char`\\n} = execute;~~{\ttfamily C-c} = SIGINT;~~{\ttfamily C-d} = EOF;~~{\ttfamily C-z} = SIGTSTP;~~{\ttfamily C-l} = clear screen\\[1pt]
Shell constructs: pipes, redirects, command chaining;~~Programs: shell builtins, editors (vim/nano), REPLs, pagers

\tcbline

\textbf{\color{red!70!black}[3] Initial State} \hfill \textit{dynamic, injected per trajectory}\\[2pt]
The initial container snapshot of the terminal environment is:\\[1pt]
OS: Ubuntu 22.04.3 LTS (x86\_64)\quad Kernel: Linux 5.15.0-134-generic\\
Packages: Python 3.10.12, pip 23.2.1, Node 18.17.1, gcc 11.4.0, git 2.34.1, Docker 24.0.5\\
Disk: 64\,GB (41\,GB free)\quad RAM: 16\,GB\quad Working dir: {\ttfamily /workspace}\\[1pt]
The initial terminal state is:\\[1pt]
{\ttfamily root@6b254155-5503-4b86-837b-fd0f080ab297:/workspace\char`\#}

\tcbline

\textbf{\color{blue!70!black}[4] Demonstrations} \hfill \textit{static}\\[2pt]
\textbf{Turn~1} \,\textbar\, Action: {\ttfamily\dq{}git clone https://github.com/expressjs/express.git /tmp/express\char`\\n\dq{}}\\
\hphantom{\textbf{Turn~1} \,\textbar\,} Observation: {\ttfamily root@6b254155:/workspace\char`\# git clone https://github.com/expressjs/express.git /tmp/express}\\
\hphantom{\textbf{Turn~1} \,\textbar\, Observation: }{\ttfamily Cloning into \sq{}/tmp/express\sq{}...}\\[1pt]
\textbf{Turn~2} \,\textbar\, Action: {\ttfamily\dq{}\dq{}}, duration=3.0\\
\hphantom{\textbf{Turn~2} \,\textbar\,} Observation: {\ttfamily remote: Enumerating objects: 32841, done.}\\
\hphantom{\textbf{Turn~2} \,\textbar\, Observation: }{\ttfamily remote: Counting objects: 100\char`\% (1205/1205), done.}\\
\hphantom{\textbf{Turn~2} \,\textbar\, Observation: }{\ttfamily Receiving objects: 100\char`\% (32841/32841), 12.76 MiB | 8.53 MiB/s, done.}\\
\hphantom{\textbf{Turn~2} \,\textbar\, Observation: }{\ttfamily Resolving deltas: 100\char`\% (21567/21567), done.}\\
\hphantom{\textbf{Turn~2} \,\textbar\, Observation: }{\ttfamily root@6b254155:/workspace\char`\#}\\[1pt]
{\color{gray}\textit{[\,\dots\; 6 more turns demonstrating directory navigation and build operations \dots\,]}}

\tcbline

\textbf{\color{red!70!black}[5] Simulation Instruction} \hfill \textit{dynamic, injected per trajectory}\\[2pt]
Simulate a system with CUDA~11.8 drivers and only 2\,GB free disk space. {\ttfamily pip install torch==2.1.0} should complete the download phase normally (progress bar reaching 100\char`\%), then fail during unpacking of {\ttfamily libtorch\_cuda.so} with {\ttfamily OSError: [Errno 28] No space left on device}. After the failure, {\ttfamily pip cache list} should report the partially downloaded {\ttfamily .whl} file still present. {\ttfamily df -h} should show {\ttfamily /tmp} at 100\char`\% usage.\\[1pt]
{\color{gray}\textit{See \S\ref{sec:app:controllable} for the controllable simulation capability.}}
\end{tcolorbox}
\caption{
Anatomy of a Terminal domain LWM RL system prompt, showing the five components defined in \S\ref{sec:prelim:schema}. {\color{blue!70!black}Blue} = static (shared across trajectories); {\color{red!70!black}red} = dynamic (injected per trajectory).
}
\label{fig:system_prompt_example}
\end{figure*}

\subsection{Unified Environment Trajectory Schema}
\label{sec:prelim:schema}

Training a single world model across seven domains with state representations as varied as file-system snapshots for Terminal and UI view hierarchies for Android requires a shared format.
We adopt the following environment trajectory schema uniformly across all seven domains and training stages:

\begin{tcolorbox}[
  colback=gray!3, colframe=black!50,
  boxrule=0.5pt, arc=2pt,
  title={\small\bfseries Unified Environment Trajectory Schema},
  top=4pt, bottom=4pt, left=6pt, right=6pt
]
\small
\begin{tabular}{@{}r@{$\;\,:=\;\,$}l@{}}
$\mathrm{system\_prompt}$ & $\mathrm{task\_description} \oplus \mathrm{action\_space} \oplus \mathrm{initial\_state} \oplus \mathrm{demonstrations} \oplus \mathrm{simulation\_instruction}$ \\[4pt]
$\mathrm{turn}_t$ & $(\mathrm{action}_t,\,\mathrm{observation}_t)$ \\[4pt]
$\mathrm{trajectory}$ & $\mathrm{system\_prompt} \oplus [\mathrm{turn}_1, \ldots, \mathrm{turn}_T]$
\end{tabular}
\end{tcolorbox}

In this schema, an \emph{action} is the agent's output at one turn (e.g., a tool call or shell command) and an \emph{observation} is the environment's feedback (e.g., tool response or command output).

Figure~\ref{fig:system_prompt_example} illustrates a fully assembled system prompt from the Terminal domain, with representative excerpts from each component.
The system prompt has five components, whose construction is detailed in \S\ref{sec:pipeline:prompt}.
The \emph{task description} instructs the model to act as a world model for a specific domain and defines the simulation objective.
The \emph{action space} enumerates the available tools or operations and their calling conventions.
The \emph{initial state} specifies the environment's starting configuration before any interaction begins: installed packages, file-system layout, UI screen state, or any other precondition that the trajectory assumes.
Together with the interaction history and current action, a sufficiently detailed initial state substantially constrains the expected observation.
\emph{Demonstrations} are optional few-shot (action, observation) examples.
The \emph{simulation instruction} specifies controllable simulation conditions (e.g., ``hide the answer from the \texttt{web\_search} responses'') and is used primarily for the controllable simulation experiments (\S\ref{sec:app:controllable}).

\begin{figure*}[t]
\centering
\begin{subfigure}[t]{\linewidth}
  \centering
  \includegraphics[width=\linewidth]{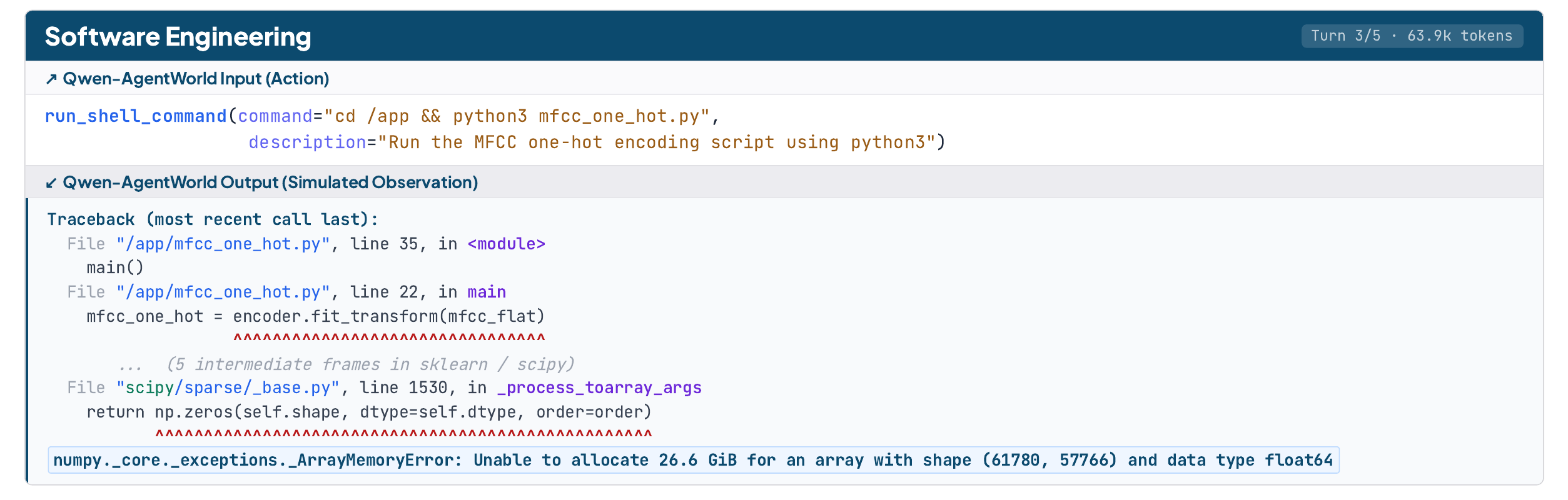}
  \caption{
  \textbf{SWE} (text-based): The agent runs a Python script; the world model predicts the full traceback including an out-of-memory error during one-hot encoding.
  }
  \label{fig:domain_example_swe}
\end{subfigure}
\vspace{4pt}
\begin{subfigure}[t]{\linewidth}
  \centering
  \includegraphics[width=\linewidth]{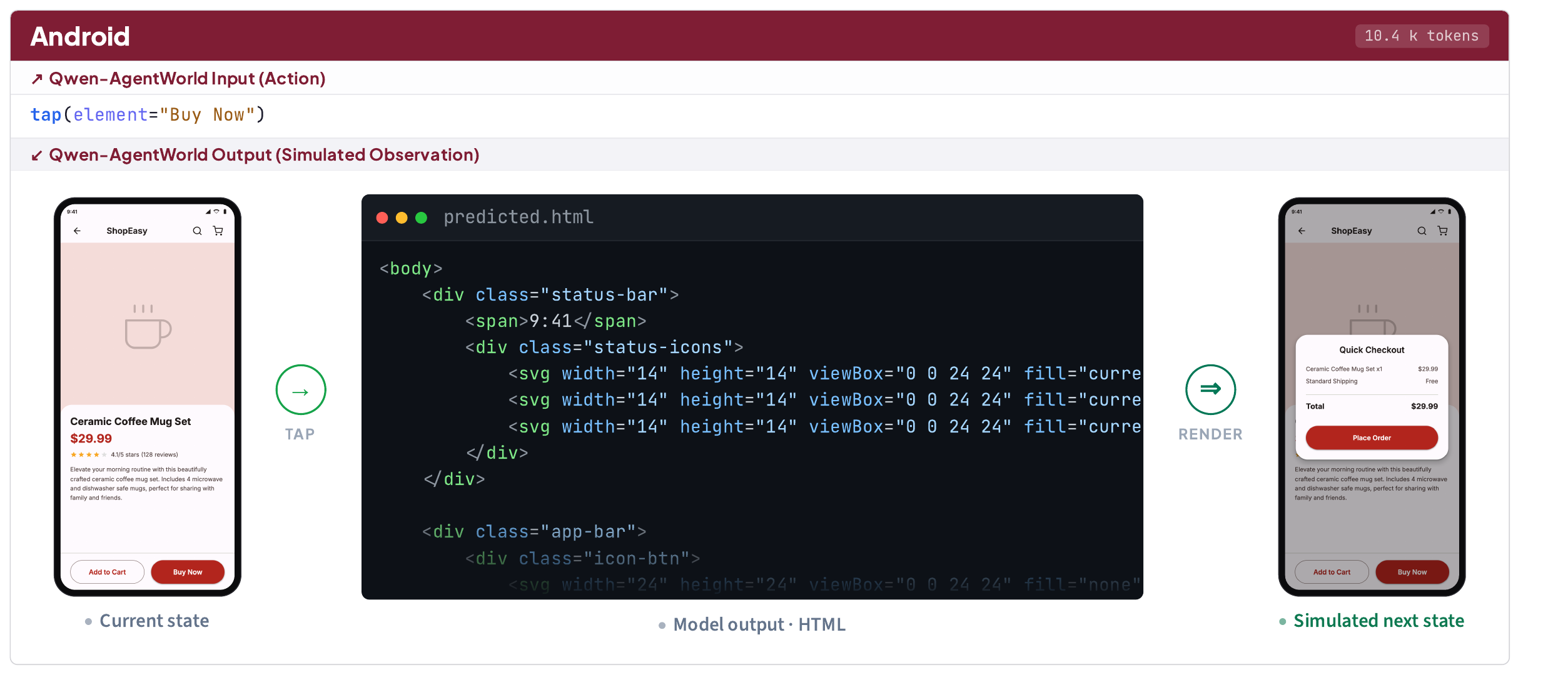}
  \caption{
  \textbf{Android} (GUI): The agent taps ``Buy Now'' on a product page; the world model predicts an HTML representation of the next screen, which is rendered as the predicted checkout sheet.
  }
  \label{fig:domain_example_android}
\end{subfigure}
\caption{Representative interaction examples from a text-based domain (SWE) and a GUI domain (Android), illustrating the breadth of the observation space.}
\label{fig:domain_examples}
\end{figure*}

\subsection{Language World Model}
\label{sec:prelim:formulation}

A language world model (LWM) is a conditional text generator that predicts the next environment observation given the interaction history and the agent's current action.
Let $c$ denote the system prompt, $o_t$ the environment observation at turn $t$, and $a_t$ the agent's action. The LWM $f_\theta$ produces:
\begin{equation}
    \hat{o}_{t+1} = f_\theta(c,\, o_{\leq t},\, a_{\leq t}),
\end{equation}
where the conditioning context comprises $c$, the full interaction history, and the current action. The training target is the ground-truth observation $o_{t+1}$.

Table~\ref{tab:domain_taxonomy} lists all seven domains with their observation and action representations.
In \emph{stateless} environments (e.g., Search), state is carried implicitly in the conversation history, whereas \emph{stateful} environments (e.g., Terminal, OS) maintain an explicit internal state that evolves with each action.
In both cases the LWM operates over the same observation sequence, as formalized by the schema in \S\ref{sec:prelim:schema}.
Yet the diversity of domains ensures that a single world-modeling objective simultaneously exercises reasoning, knowledge, and long-context understanding.
These capabilities are also foundational to general agents (\S\ref{sec:app:foundation}).

\clearpage

\begin{table}[!ht]
\caption{The seven domains covered by \method, with their action, observation, and core capability exercised by next-state prediction.}
\label{tab:domain_taxonomy}
\centering
\small
\resizebox{\linewidth}{!}{%
\begin{tabular}{@{}llll@{}}
\toprule
\textbf{Domain} & \textbf{Action} & \textbf{Observation} & \textbf{Core Capability} \\
\midrule
MCP      & JSON Tool Call               & Tool response (file content, DB, etc.) & Factual world knowledge \\
Search   & Web Search / Web Extractor   & Conversation history (query + results) & Factual world knowledge \\
SWE      & Read / Edit / Bash / \ldots  & Tool output (file content + diffs) & Code execution reasoning \\
Terminal & Bash Commands / Keystrokes   & Terminal output (stdout + shell prompt) & Long-context causal reasoning \\
Android  & Touch / Swipe / Type / ...         & UI view hierarchy + app state & Visual state reasoning \\
Web      & Click / Type / Navigate / ...     & Accessibility tree + browser state & Visual state reasoning \\
OS       & Mouse / Keyboard             & Accessibility tree + window/app state & Visual state reasoning \\
\bottomrule
\end{tabular}
}
\end{table}

Figure~\ref{fig:domain_examples} illustrates this breadth with one representative example from each category.
In SWE (Figure~\ref{fig:domain_example_swe}), the agent runs a Python script and the LWM must reason through memory allocation and array dimensions to predict the resulting out-of-memory traceback.
In Android (Figure~\ref{fig:domain_example_android}), the agent taps a UI element, and the LWM must infer how the interaction transforms the page layout and predict the resulting screen as renderable HTML.
Appendix~\ref{sec:appendix:domain_examples} presents interaction examples for other domains.

\section{Training Recipe}
\label{sec:pipeline}

\begin{figure}[!ht]
    \centering
    \includegraphics[width=\linewidth]{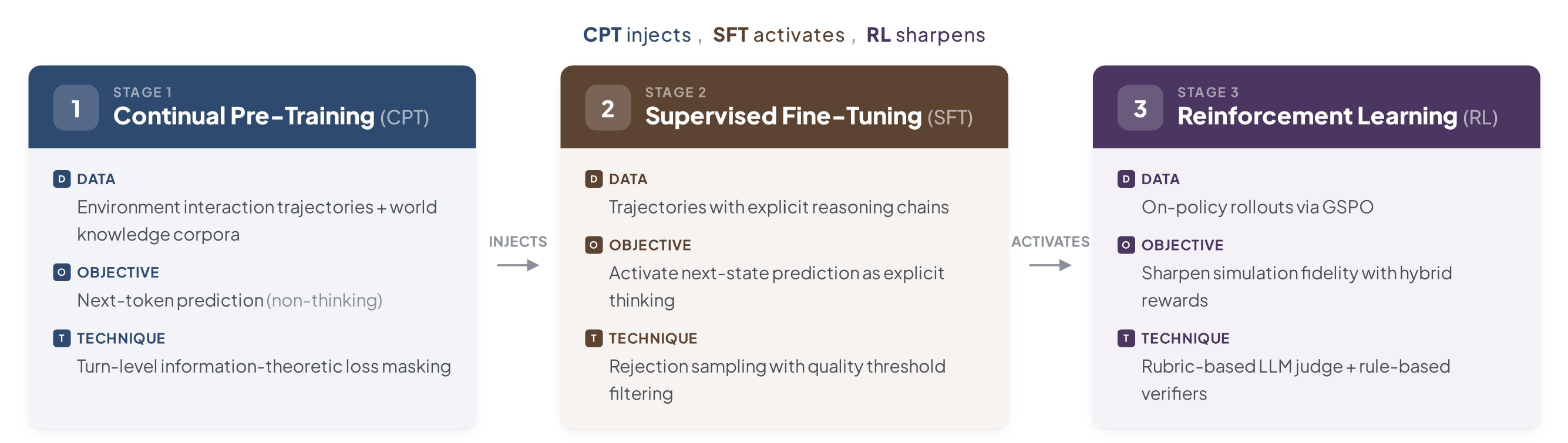}
    \caption{Three-stage training pipeline of \method. Stage~1 CPT injects world knowledge; Stage~2 SFT instills next-state-prediction thinking patterns; Stage~3 RL sharpens output quality.}
    \label{fig:pipeline}
\end{figure}

\method is trained end-to-end with environment modeling as the explicit objective from continual pre-training onward.
As shown in Figure~\ref{fig:pipeline}, we adopt the principle \emph{``CPT injects, SFT activates, RL sharpens''}, yielding a three-stage pipeline: Stage~1 CPT injects environment world knowledge through non-thinking trajectories (\S\ref{sec:pipeline:cpt}); Stage~2 SFT activates next-state prediction as an explicit thinking pattern (\S\ref{sec:pipeline:sft}); Stage~3 RL sharpens output quality with hybrid rubric-and-rule rewards (\S\ref{sec:pipeline:rl}).

\subsection{Training Data}
\label{sec:pipeline:data}

We first describe the data sources and unified processing pipeline that supply all three training stages.

\subsubsection{Environment Trajectories Collection}

No existing public dataset covers this breadth of domains at the volume required for world model training.
We collect environment trajectories from three complementary sources:

\begin{itemize}[leftmargin=1.5em,itemsep=2pt]
\item \textbf{Dedicated Agent Infrastructure.}
We deploy a suite of agent--environment backends: containerized execution sandboxes for code and tool invocation, MCP servers, persistent terminal sessions with full shell state tracking.
For GUI domains, we deploy persistent Android, browser, and desktop OS environments that represent GUI observations as textual accessibility trees and UI view hierarchies for world-model training. These environments run on physical hosts provisioned with Ubuntu, macOS, and Android virtual machines.
On top of this infrastructure, we automatically synthesize task queries spanning each domain's target distribution and let agentic systems execute them end-to-end.
This pipeline runs continuously and is the primary source of scalable, controlled, reproducible interaction data.

\item \textbf{Open Environment Interaction Traces.}
We collect naturally occurring action--environment interaction traces from public sources:
terminal session recordings,
open-source agentic tool-call logs, and execution traces in code repositories.
These raw traces are noisy, structurally heterogeneous, and often incomplete.
Therefore, we build a multi-agent cleaning pipeline in which specialized agents handle fetching, denoising, segmentation, semantic alignment, and quality scoring as separate stages.
Only sequences that pass every stage enter the training pool.
This source captures long-tail interaction patterns (unusual shell workflows, rare API error modes, idiosyncratic tool-call chains), complementing the controlled distribution of the dedicated infrastructure.

\item \textbf{In-House Agentic Trajectories.}
We draw from in-house foundation model SFT agentic trajectories accumulated during routine model development, covering all seven domains.
These trajectories are converted into environment trajectories with format unification (\S\ref{sec:prelim:terminology}).
\end{itemize}

The data pools for the three stages are strictly disjoint.
CPT data draws from dedicated agent infrastructure, open interaction traces, and specialized-domain world knowledge corpora (\S\ref{sec:pipeline:cpt}).
SFT and RL draw exclusively from internally accumulated trajectories.
Since RL amplifies data artifacts through on-policy rollouts, we invest most of the data engineering effort into the downstream pipeline.
Table~\ref{tab:data_stats} summarizes the SFT and RL data statistics.

\begin{table}[!ht]
\caption{SFT and RL training data statistics across all seven domains. Average token counts and turn counts are computed over the RL training pool.}
\label{tab:data_stats}
\centering
\small
\begin{tabular}{@{}lrrrr@{}}
\toprule
\textbf{Domain} & \textbf{SFT} & \textbf{RL Train} & \textbf{Avg.\ tokens} & \textbf{Avg.\ turns} \\
\midrule
MCP      & 179   & 4{,}156  & 62{,}702 & 28.9 \\
Search   & 1{,}042 & 20{,}004 & 18{,}873 & 6.2  \\
Terminal & 1{,}580 & 34{,}125 & 5{,}805  & 12.0 \\
SWE      & 249   & 8{,}181  & 36{,}734 & 24.7 \\
Android  & 1{,}337 & 11{,}498 & 30{,}064 & 19.3 \\
Web      & 1{,}605 & 8{,}716  & 19{,}417 & 10.2 \\
OS       & 1{,}102 & 5{,}628  & 25{,}439 & 12.4 \\
\midrule
Total & 7{,}094 & 92{,}308 & 19{,}443 & 13.4 \\
\bottomrule
\end{tabular}
\end{table}

\subsubsection{Unified Data Processing}

All training data follows the unified environment trajectory schema defined in \S\ref{sec:prelim:schema}: a system prompt specifying the simulation context, followed by alternating user turns (agent actions) and assistant turns (environment observations).
Raw agentic trajectories from heterogeneous sources are normalized into this format through domain-specific handlers and the shared preprocessing steps described below.

\paragraph{Trajectory-to-Turn Expansion.}
We first expand each multi-turn trajectory into multiple turn-level prediction samples.
For a trajectory with $T$ turns, any turn $t$ can serve as a prediction target: the preceding interactions $(\text{turn}_1, \ldots, \text{turn}_{t-1})$ concatenated with the state and action at turn $t$ form the input, and the observation at turn $t$ becomes the target.
Because agent--environment interaction is inherently multi-turn and each observation depends only on its preceding history, every turn within a trajectory is itself a valid prediction instance once paired with its prior context.
For the training split, we randomly sample one turn from each trajectory to diversify the training objectives.

\paragraph{Data Filtering.}
We filter trajectories at both the trajectory and turn levels before training.
At the trajectory level, we drop sequences with fewer than two turns, discard MCP and SWE trajectories that invoke tools absent from the declared action space, and exclude GUI trajectories affected by environment failures, such as missing state files, CAPTCHA challenges, or HTTP errors, that disrupt the causal relationship between actions and subsequent environment states.
At the turn level, we strip empty-action turns caused by pauses or demonstration narration and apply two non-trivial filters described below.

\begin{itemize}[leftmargin=1.5em,itemsep=2pt]
    \item \textbf{Retry-Cycle Skipping.}
    Agents frequently fall into ``garbage output $\to$ error $\to$ retry'' cycles.
    Simply deleting these turns breaks the state chain. Instead, we skip the offending (user, assistant) pairs while preserving the current state so that the next valid turn inherits the correct history.

    \item \textbf{No-Change Turn Filtering} (GUI domains).
    We remove turns whose pre-action and post-action states show no effective change, typically caused by slow system response or network latency.
    If retained, these samples teach the model to copy the previous state as-is regardless of the action taken, undermining its ability to predict how actions change the environment.
\end{itemize}

\paragraph{System Prompt Construction.}
\label{sec:pipeline:prompt}
As defined in \S\ref{sec:prelim:schema}, each system prompt comprises five components: \emph{task description}, \emph{action space}, \emph{initial state}, \emph{demonstrations}, and \emph{simulation instruction}.
The initial state and simulation instructions are optional, while all other fields are always present.
Figure~\ref{fig:system_prompt_example} illustrates a real system prompt from the Terminal domain.
Which components are static (shared across trajectories) and which are dynamic (filled per trajectory) varies by domain:

\begin{itemize}[leftmargin=1.5em,itemsep=2pt]
\item \textbf{Static Components.}
For all domains except MCP and SWE, the action space and demonstrations are predefined: each domain has a fixed set of available operations and a fixed set of canonical action--observation examples.
MCP and SWE trajectories require per-trajectory action spaces because the available tools differ across MCP server instances and code repositories.
\item \textbf{Dynamic Components.}
When present, the initial state and the simulation instruction are dynamic (filled per trajectory).
The initial state captures the environment's starting configuration (installed packages, file-system layout, database contents, UI screen state, or other domain-specific preconditions) and provides the preconditions that, together with the interaction history and current action, determine the expected observation (\S\ref{sec:prelim:schema}).
For GUI domains, the initial state is intentionally diverse.
During training and inference, it may start from portal pages, Google Search, or arbitrary websites and desktop/app states.
The simulation instruction specifies controllable conditions for the trajectory.
For the search domain, each trajectory is annotated with a reverse-engineered simulation instruction containing the target query, reference answer, and a no-leakage constraint.
For terminal and other domains that depend strongly on the initial environment, we augment a subset of the data with output-controlling instructions.
These instructions train the model to simulate according to given directives, establishing ``simulate according to this instruction'' as a learned pattern that directly supports the controllable simulation capability described in \S\ref{sec:app:controllable}, while making the intended response boundary explicit to reduce hallucination during SFT and RL training.
\end{itemize}

\paragraph{System Prompt Template Construction via AutoResearch.}
Crafting effective system prompts requires deep domain expertise and extensive manual iteration: each prompt must encode domain-specific state-transition rules, output formatting constraints, and demonstration patterns.
Rather than hand-crafting these templates, we formulate prompt optimization as an automated research problem~\citep{karpathy2026autoresearch} with a clear objective: maximize the world model's prediction accuracy on held-out real trajectories.
The pipeline proceeds iteratively: in each cycle, an optimizer agent analyzes sampled trajectory data to identify domain-specific patterns and common failure modes, then drafts or revises a candidate system prompt.
The candidate prompt is loaded into the world model, which runs inference on held-out real trajectories. A separate judge model scores the predictions against ground-truth environment responses.
The optimizer agent examines both the scores and concrete prediction errors to pinpoint weaknesses, producing a targeted revision for the next cycle.
Each optimization run executes 10 such propose--evaluate--refine iterations.
We launch 12 runs in parallel, each seeded with a distinct style directive (verbose specification, concise checklist, demonstration-heavy, etc.), yielding 12 template variants (v0--v11) ranging from a minimal $\sim$30-line constraint-style template to a $\sim$1100-line specification-style template.
Human reviewers audit and approve each final version.
The three training pools draw from disjoint template subsets: RL uses v0, CPT uses v1, and SFT randomly samples from v2--v11 per sample to maximize prompt-format diversity.

\subsection{Stage~1: Continual Pre-Training}
\label{sec:pipeline:cpt}

\paragraph{CPT Data.}
The objective of CPT is to model real-world environment behavior while injecting broad world knowledge into the model.
Accordingly, beyond the environment trajectories from the data collection pipeline above, we further incorporate specialized-domain world knowledge corpora spanning a broad range of professional and factual domains: industrial control and manufacturing, cybersecurity, law and regulation, medicine and healthcare, finance, current affairs and encyclopedic knowledge.
These corpora ground the model in factual world knowledge that environment trajectories alone cannot provide: simulating a regulatory compliance platform requires legal knowledge, simulating a hospital information system requires medical knowledge, and simulating search-engine responses on current events requires up-to-date factual coverage.
This breadth of domain knowledge also enables the model to generalizably construct specialized environments beyond the seven training domains (\S\ref{sec:app:scaling}).

\paragraph{Training Objective.}
CPT trains under the standard next-token prediction objective.
Multi-turn environment trajectories are framed as world-modeling tasks: the system prompt defines the simulation context, user turns carry the agent's actions, and assistant turns carry the environment's responses, mapping the language-modeling objective directly onto $p(o_{t+1} \mid o_{\leq t}, a_{\leq t})$.
Because each trajectory is expanded into turn-level prediction samples (\S\ref{sec:pipeline:data}), the model receives supervision at every turn within a trajectory's history.
Specialized-domain world knowledge corpora enter as single-turn data under the same objective.

\paragraph{Turn-Level Information-Theoretic Loss Masking.}
We find that many turns in tool-use trajectories are boilerplate, such as tools that simply echo their input or APIs that mirror request parameters.
The gradients induced by these turns are often low-quality and introduce noise, but the turns cannot simply be removed because subsequent turns depend on them as context.
To this end, we compute four statistics per (action, observation) pair---Overlap (OL), Novelty (Nov), Jaccard (Jac), and length ratio (R)---and assign each turn to one of seven semantic categories with the keep ratios shown in Table~\ref{tab:loss_mask_categories}.
This is designed to identify turns that carry genuine world knowledge and thus have meaningful learning value.
Importantly, categories are determined statistically rather than by tool name, keeping the classification tool-agnostic.
Given the word sets $W_{\text{act}}$ and $W_{\text{obs}}$ extracted from the action and observation (lowercased and deduplicated tokens):
$\text{OL} = |W_{\text{act}} \cap W_{\text{obs}}| \,/\, |W_{\text{act}}|$ measures how much action vocabulary the observation echoes;
$\text{Nov} = |W_{\text{obs}} \setminus W_{\text{act}}| \,/\, |W_{\text{obs}}|$ captures the fraction of genuinely new information in the observation;
$\text{Jac} = |W_{\text{act}} \cap W_{\text{obs}}| \,/\, |W_{\text{act}} \cup W_{\text{obs}}|$ gives the symmetric word-set similarity;
and $R = |\text{obs}| \,/\, |\text{act}|$ is the character-level length ratio.
Masked turns are excluded from the loss computation while their tokens are retained as context for subsequent turns.
This decouples ``learning the next state'' from ``learning the next token'': loss is computed only on turns that carry genuine environment information.
The same filtering applies to any trajectory-format training data, since the four statistics (OL, Nov, Jac, R) are computed from surface-level token overlap and require no domain-specific annotation.

\begin{table}[!ht]
\caption{Seven turn categories for information-theoretic loss masking. Categories are determined from statistical signals rather than tool names. Keep ratio is the fraction of tokens used in loss computation.}
\label{tab:loss_mask_categories}
\centering
\small
\begin{tabular}{@{}lp{4.5cm}p{4cm}r@{}}
\toprule
\textbf{Category} & \textbf{Statistical signature} & \textbf{Intuition} & \textbf{Keep} \\
\midrule
retrieval    & Nov $\geq 60\%$, R $> 1$           & \texttt{read\_file} $\to$ contents       & 100\% \\
expansion    & OL $\geq 50\%$, Nov $\geq 50\%$, R $> 1.5$ & \texttt{fetch} $\to$ page + metadata & 100\% \\
action       & Nov $\geq 50\%$, R $\leq 1$ or short & \texttt{send\_email} $\to$ ``sent''     & 100\% \\
transform    & Nov $< 50\%$, R $< 1$               & long input $\to$ status word            & 50\%  \\
boilerplate  & OL $\geq 50\%$, Nov $< 50\%$        & API echo                                & 10\%  \\
echo         & OL $\geq 70\%$, Nov $< 30\%$        & \texttt{think(x)} $\to$ \texttt{\{thought:x\}} & 5\% \\
other        & uncategorized                       & ---                                     & 100\% \\
\bottomrule
\end{tabular}
\end{table}

\subsection{Stage~2: Supervised Fine-Tuning}
\label{sec:pipeline:sft}

After CPT, the model has learned what tools return and how states evolve, but this knowledge is applied only implicitly through next-token prediction over observation tokens.
Through SFT, we explicitly activate next-state prediction as a reasoning pattern.
We use SFT to teach the model to engage in next-state prediction explicitly to better activate the state-transition knowledge acquired during CPT (identifying what an action requests, recalling the prior state, and anticipating the expected response format).
This explicit reasoning reduces hallucinations and improves state consistency over long trajectories.
During the SFT stage, we still use standard next-token prediction as the training objective, the same loss used in CPT.
We employ a 256k-token context window to accommodate long multi-step trajectories.

\paragraph{SFT Reasoning Trace Curation.}
The SFT stage shifts from the non-thinking regime of CPT to thinking trajectories that contain explicit reasoning chains.
Starting from the SFT pool, we first diversify prompt templates across samples, then generate multi-beam rollouts and apply rejection sampling to curate the final training set.
(i)~\textbf{Prompt template diversification.}
Each SFT sample has its default system prompt replaced by one uniformly sampled from 10 template variants (\S\ref{sec:pipeline:prompt}).
The trajectory's dynamic content (tool definitions, demonstrations, simulation instruction) is preserved and re-injected into the new template.
Token counts are recomputed and samples exceeding 256k are truncated.
This improves the model's generalization across system prompt variations by exposing the same trajectory to diverse prompt formats during SFT.
(ii)~\textbf{Rejection sampling.}
For each query, we generate three rollouts from a general-purpose reasoning model.
The candidates are scored by an independent judge and compared pairwise to select the highest-quality trajectory.
If the winning trajectory's score falls below the minimum threshold, the query is discarded.
Table~\ref{tab:rejection_sampling} reports the rejection sampling statistics.
Starting from 10{,}250 candidate queries, the pipeline retains 7{,}094 trajectories (69.2\% retention rate).

\begin{table}[!ht]
\caption{Rejection sampling statistics per domain. ``Candidates'' is the number of queries with complete rollouts. ``Retain rate'' is the fraction of queries whose best-of-three trajectory exceeds the quality threshold. ``Final SFT'' is the count after filtering.}
\label{tab:rejection_sampling}
\centering
\small
\begin{tabular}{@{}lrrrr@{}}
\toprule
\textbf{Domain} & \textbf{Candidates} & \textbf{Retain rate} & \textbf{Final SFT} & \textbf{Avg.\ turns} \\
\midrule
MCP      & 261    & 68.6\% & 179     & 24.3 \\
Search   & 1{,}466 & 71.1\% & 1{,}042 & 3.3 \\
Terminal & 1{,}826 & 86.5\% & 1{,}580 & 5.9 \\
SWE      & 402    & 61.9\% & 249     & 26.9 \\
Android  & 1{,}975 & 67.7\% & 1{,}337 & 15.9 \\
Web      & 2{,}697 & 59.5\% & 1{,}605 & 3.0 \\
OS       & 1{,}623 & 67.9\% & 1{,}102 & 5.4 \\
\midrule
Total & 10{,}250 & 69.2\% & 7{,}094 & 8.5 \\
\bottomrule
\end{tabular}
\end{table}

\subsection{Stage~3: Reinforcement Learning}
\label{sec:pipeline:rl}

We employ GSPO~\citep{zheng2025group} for the RL stage on the data described above.
RL for LWM poses distinctive challenges due to the difficulty and open-ended nature of environment feedback prediction.
Moreover, because the context is substantially longer than the target output, LWM RL exhibits an extreme prompt--output asymmetry: the prompt consists of the full trajectory history up to the prediction turn and often extends to tens of thousands of tokens, whereas the output, a single predicted observation, typically contains only a few hundred to a few thousand tokens.
As a result, per-sample compute cost is dominated by prompt processing rather than generation.
We cap the prompt at 128k tokens for the RL pool, filtering out trajectories whose history exceeds this limit.
The threshold covers the vast majority of trajectories across all seven domains.

We systematically investigate how to achieve stable RL for world modeling. Specifically, we focus on two aspects: reward design (\S\ref{sec:pipeline:rl:reward}) and training stability (\S\ref{sec:pipeline:rl:stability}).
Appendix~\ref{sec:appendix:dynamics} reports the training dynamics of \method-35B-A3B, tracking all five rubric dimensions throughout RL training, which reveals that they improve at markedly different rates.
Factuality shows the largest relative improvement (11.3\%) yet remains the lowest-scoring dimension throughout, confirming that factual world knowledge is the hardest aspect of environment simulation.

\subsubsection{Reward Design}
\label{sec:pipeline:rl:reward}

We systematically explore how to design effective rewards for world-model RL. The reward combines two complementary signals that address different failure modes.

\begin{itemize}[leftmargin=1.5em,itemsep=2pt]
    \item \textbf{Five-Dimensional Rubric (LLM Judge).}
    Using rubrics as structured rewards for RL has been shown effective in non-verifiable domains~\citep{gunjal2025rubrics}; recursive rubric refinement~\citep{shen2026rethinking} can further improve judge and reward quality.
    Each predicted observation is scored by an LLM judge on the five-dimensional rubric defined in \S\ref{sec:bench:protocol}, each on a 1--5 scale.
    The total reward equals the mean $\times\,5$, yielding a range of $[5,\,25]$.
    If the judge fails to produce a valid score, the reward defaults to~0.
    Each domain uses a tailored judge system prompt with content-type classification (\S\ref{sec:bench:protocol}) to reduce false negatives from irreproducible details such as timestamps and PIDs.
    We invoke the RL judge via asynchronous distributed calls.

    \item \textbf{Rule-Based Verifier.}
    A subset of the data carries executable verifier code that produces a binary 0/1 correctness signal, scaled to $[0,\,25]$ to align with the rubric's range.
    Rule-based rewards serve as an objective anchor, effectively mitigating reward hacking induced by open-ended rewards.

\end{itemize}

We combine the two signals at 9{:}1 (rubric{:}rule), balancing multi-dimensional rubric feedback with strict binary correctness.
We use a multi-strategy JSON parser and strict tag extraction to ensure that only the predicted observation reaches the judge, preventing self-praise in the reasoning from affecting the score.

\subsubsection{Training Stability}
\label{sec:pipeline:rl:stability}

We identify three failure modes through systematic ablation and describe solutions that proved effective.

\paragraph{Reward Collapse from Multi-Turn Expansion.}
We find that when training trajectories are expanded into multiple samples following the procedure described in \S\ref{sec:pipeline:data}, the resulting training instances share a long common prefix, which in turn causes training to collapse quickly.
This is related to the ``Echo Trap'' identified in multi-turn agent RL~\citep{wang2025ragen}, where reward variance collapses and the policy degenerates.
We resolve this by restricting expansion to exactly one turn per trajectory in the RL pool, giving each training sample a unique prediction target with no shared-prefix overlap.

\paragraph{Reward Shaping.}
The choice of reward formulation has a strong effect on convergence~\citep{zhu2025planner}.
We compare the above open-ended reward design against two alternatives.
\emph{Reference-Reward} presents the judge with the ground-truth observation and asks it to choose, in a pairwise A/B test, whether the policy's predicted observation or the initial policy checkpoint's output is closer to the ground truth, yielding a binary 0/1 signal. This design converges slowly: the binary reward is sparse, and when both outputs are plausible but differ in surface form, the judge's preference becomes unstable, injecting noise into the gradient.
\emph{Turing-Test Reward} asks a judge whether the predicted observation could plausibly have come from a real environment.
This reward barely converges, primarily because the false-negative rate is too high.
When the model's generation is very close to or even identical to the ground truth, asking the judge to determine which one is more likely to have come from a real environment introduces an unreliable training signal regardless of which answer is chosen.
Overall, the five-dimensional rubric combined with the rule-based verifier converges stably.
By decomposing the assessment into orthogonal dimensions and reserving a binary anchor for cases with executable ground truth, this design provides consistent, informative gradients even when the observation space is multimodal.

\paragraph{Reward Hacking Through Self-Praise.}
The policy can learn to exploit the judge's specific biases by inserting self-praising phrases into the predicted observation to inflate scores without improving simulation fidelity~\citep{wang2026reward}.
We observe this behavior concretely: the policy learns to embed qualitative affirmations (e.g., ``operation completed successfully with all fields correctly populated'') or echo back key terms from the judge prompt that trigger higher rubric scores.
We apply three mitigations.
First, the rule-based verifier grounds a fraction of the reward in binary correctness that cannot be influenced by judge model biases, providing a stable anchor to counteract reward hacking.
Second, the content-type classification in the judge prompt narrows the scope of each rubric dimension, so that deterministic content is judged by exact match. Self-praise in a region classified as deterministic earns zero credit.
Third, through the tag extraction procedure described above, we strictly enforce a clear separation of the model’s predicted content.
This not only ensures that the thinking block is never exposed to the judge, but also explicitly indicates which parts correspond to the final prediction and which belong to the policy output.

\section{\bench}
\label{sec:bench}

We construct \textbf{\bench} to evaluate language world models across seven agent interaction domains.
When an LWM simulates environments, predicting each turn's observation requires comprehending the full interaction history accumulated across all prior turns.
As trajectories grow longer, maintaining fidelity to this context becomes increasingly challenging, making \bench a naturally grounded long-context benchmark.
Ensuring simulation quality further requires that a world-model benchmark verify whether predicted environment responses match reality across format, factuality, state consistency, and domain-specific conventions.

\subsection{Benchmark Construction}
\label{sec:bench:construction}

\bench is constructed from agentic trajectories of frontier models on established benchmarks across all seven domains, converted into environment trajectories (\S\ref{sec:prelim:schema}) with ground-truth observations obtained from the real environments.

\paragraph{Construction Principles.}
\bench is built on four construction principles:
(i)~\emph{Widely-Used Queries:} all task queries are drawn from established high-quality agentic benchmarks rather than self-constructed tasks, so that the task distribution aligns with the scenarios that current agent development targets;
(ii)~\emph{Frontier-Agent Trajectories:} all trajectories are generated by frontier-model agents, whose actions (long reasoning chains, tool-call compositions, and error-recovery sequences) are high-quality and sufficiently complex to stress-test world-model fidelity at the frontier scale;
(iii)~\emph{Real Observations:} every trajectory is paired with ground-truth observations from real environment execution, providing a reference for evaluation;
(iv)~\emph{Out-of-Distribution:} training data and benchmark queries are partitioned at the data-source level, so that \bench probes generalization rather than memorization;

\begin{figure*}[t]
\centering
\includegraphics[width=\linewidth]{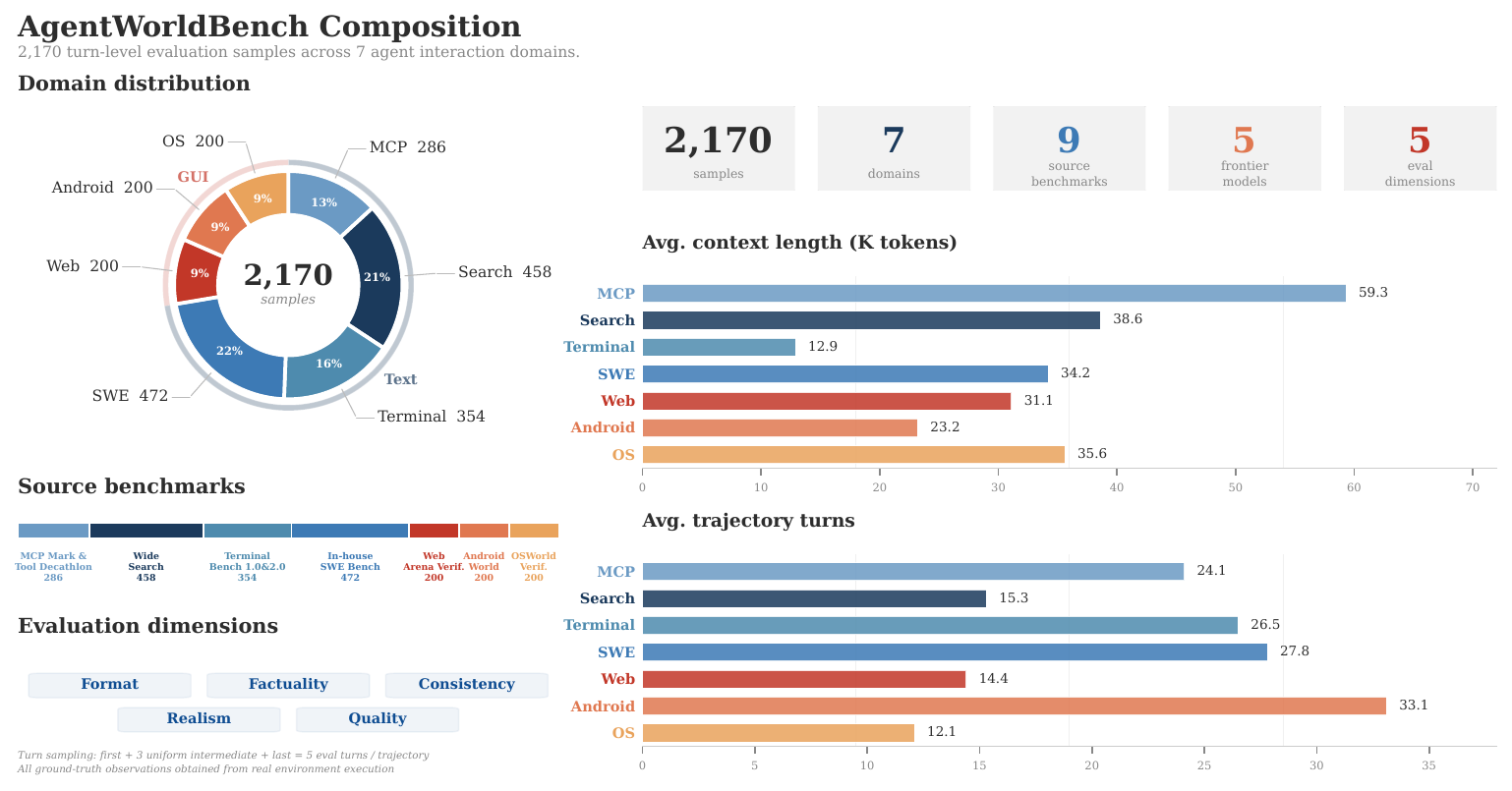}
\caption{
    Overview of \bench composition.
    \textbf{Left:} Domain distribution across seven domains, source benchmarks mapped to each domain, and the five evaluation dimensions (Format, Factuality, Consistency, Realism, Quality).
    \textbf{Right:} Summary statistics, per-domain average context length and trajectory depth.
    All ground-truth observations are obtained from real environment execution.
}
\label{fig:bench_overview}
\end{figure*}

\paragraph{Data Collection.}
Figure~\ref{fig:bench_overview} summarizes the source benchmarks and trajectory statistics for each domain.
We deploy 5 frontier agents on 9 established benchmark query sets across all 7 domains, execute them against real environments to produce agentic trajectories, and then extract environment trajectories (\S\ref{sec:prelim:schema}).
For text-based domains, an early subset (Terminal-Bench~1.0 and the in-house SWE benchmark) was collected with Claude Sonnet~4.5.
All remaining text-domain trajectories use Claude Opus~4.6 exclusively, ensuring that \bench evaluates whether an LWM can faithfully simulate environments to support the workflows of the most capable agents.
For GUI domains, we further diversify the agent pool by adding 3 strong Qwen-family models, expanding coverage to 5 frontier models and producing more varied action sequences for comprehensive evaluation.

\paragraph{Turn-level Sampling Strategy.}
\label{sec:bench:construction:turn_sampling}

Due to the large volume of raw agentic trajectories, we subsample the source trajectories for the in-house SWE benchmark and the 3 GUI benchmarks before applying the turn-level expansion described below.
When trajectories are expanded into turn-level evaluation samples, all text domains use an asymmetric sampling strategy: each trajectory keeps the first and last turns, plus three uniformly sampled intermediate turns, for five evaluation turns in total.
The first turn tests initial simulation without interaction history.
Errors at this turn propagate recursively through the state chain, so first-turn fidelity acts as an anchor.
The last turn has the longest context and the strongest dependence on accumulated prior state. It is the primary probe for long-context simulation fidelity.
The three intermediate turns provide broader coverage of mid-trajectory behavior (tool-call chaining, incremental state updates, error recovery) and dilute the boundary bias.
For GUI domains, since certain operations are overly simplistic (e.g., entering text into an input field), we selectively sampled the more challenging turns from the trajectories during the expansion process.
After the trajectory-to-turn expansion, we further retain a uniform 50\% random sample to form the final benchmark, balancing evaluation coverage against computational cost.

\paragraph{Benchmark Statistics.}
\label{sec:bench:construction:stats}
Figure~\ref{fig:bench_overview} shows the domain coverage and evaluation framework of \bench, which contains 2,170 evaluation samples in total.
The four text-based domains collectively account for 72.4\% of the benchmark, with SWE (21.8\%) and Search (21.1\%) contributing the largest shares, followed by Terminal (16.3\%) and MCP (13.2\%). The three GUI domains each contribute 9.2\%.
The average context length varies across domains: MCP samples average 59,300 tokens because each sample embeds the full tool-definition schema in its system prompt, whereas Terminal samples are the shortest at 12,900 tokens, reflecting self-contained shell sessions.

\subsection{Evaluation Protocol}
\label{sec:bench:protocol}

\bench evaluates simulation quality through an open-ended rubric: an LLM judge scores each predicted observation on five dimensions: \textbf{Format}, \textbf{Factuality}, \textbf{Consistency}, \textbf{Realism}, and \textbf{Quality}.
The primary score is the mean across the five dimensions, scaled to $[0, 100]$.
Format measures whether the output obeys the structural conventions of the domain (JSON schema compliance for MCP, shell prompt patterns for Terminal).
Factuality measures whether stated facts (file contents, search results, tool return values) are correct.
Consistency measures whether the output is internally coherent and coherent with prior turns.
Realism measures whether the simulation matches the behavioral characteristics of the real environment as evidenced by the ground truth, including response patterns, style conventions, and value plausibility.
Quality measures completeness and conciseness relative to the ground truth: critical information must not be omitted, and the output should not be excessively verbose or abbreviated compared to the reference.

\paragraph{Reference-Grounded Judging.}
The judge receives the ground-truth environment observation alongside the predicted observation and scores each dimension by comparing the two.
This reference-grounded design converts the evaluation from an open-ended quality judgment into a factual comparison, substantially narrowing the space for judge hallucination or misjudgment: the judge does not need to independently reason about what a correct environment response should look like, because the real output serves as an unambiguous reference.
This is also the primary reason for the high cross-judge consistency reported below: when scoring against a concrete reference rather than abstract quality criteria, different judge models converge on the same rankings.

\paragraph{Domain-Aware Rubrics.}
Each domain has its own judge prompt that applies the five dimensions in domain-specific terms, ensuring that evaluation not only compares against the reference objectively but also enforces the professional standards of each domain (full prompts in Appendix~\ref{sec:appendix:judge_prompts}).
For example, Terminal judges verify shell prompt patterns and cross-turn state tracking (working directory, environment variables, file-system mutations), MCP judges enforce JSON schema compliance and resource lifecycle consistency, Search judges apply a fact-priority hierarchy where contradicting the reference answer scores zero on Factuality regardless of surface plausibility, and SWE judges enforce tool execution correctness (success/failure status, exit codes) and file-system state persistence across turns.

\paragraph{Differentiated Matching Criteria.}
Not all content in an environment observation requires exact matching, and treating all content uniformly produces excessive false negatives.
We classify content into three types before evaluation.
\emph{Deterministic content} (echo output, file reads, computation results) must match exactly. A \texttt{cat} command that returns the wrong file contents is unambiguously wrong.
\emph{Pre-existing environment content} (preinstalled software versions, file contents not created by the trajectory) requires only format and plausibility verification, because a simulator cannot reproduce the exact patch version of \texttt{gcc} in a particular sandbox.
\emph{Runtime metadata} (timestamps, PIDs, memory addresses, session tokens) requires only format and range verification. A simulated PID of 42731 is as acceptable as the real 18204, provided both are valid.
This three-way split lets the judge reward correct structural and semantic behavior without penalizing irreproducible details.

\paragraph{Judge Selection.}
\label{sec:bench:judge}
We use a double-blind Turing test as a calibration tool to select the judge model and tune the judge prompt.
Given the interaction history up to a prediction turn and the current action, the judge receives two candidate observations in randomized order, one from the real environment and one from the world model, and must identify which came from the real environment.
We randomize the presentation order to control for position bias and use real observations rather than outputs from another world model to avoid distributional confounds.
The rationale is that a judge with high Turing-test accuracy must attend to fine-grained differences between faithful and plausible-but-wrong simulation, the same discriminative ability required for accurate rubric scoring.
We iteratively refine domain-specific judge prompts via autoresearch~\citep{karpathy2026autoresearch} using Turing-test accuracy as the optimization signal, analyzing per-domain error patterns to adjust rubric definitions, differentiated matching boundaries, and scoring anchors until the prompts achieve stable accuracy.

With calibrated prompts in place, we evaluate whether the choice of judge model affects evaluation outcomes.
We compare three frontier models as judge candidates, Gemini 3 Flash~\citep{gemini3flash}, Claude Sonnet 4.5~\citep{sonnet4.5}, and GPT-5.2~\citep{gpt-5.2}, by having each independently score predictions from all models in Table~\ref{tab:main_results} across all seven domains of \bench{}.
Absolute scores show a systematic spread: Gemini 3 Flash is the most lenient and GPT-5.2 the most stringent rater across all domains.
Despite these absolute differences, the model-level rankings are highly consistent across judges: the pairwise Spearman rank correlations are $\rho = 0.92$--$0.99$ (all $p < 10^{-5}$), indicating that the three judges agree on which models produce higher-fidelity predictions despite assigning different absolute scores.
The per-dimension ordering is also preserved: all three judges rank Format and Consistency as the two strongest dimensions while identifying Factuality as the weakest, with a shared ordering of Realism $>$ Quality $>$ Factuality across all judges.
We attribute this cross-judge ranking consistency to the reference-grounded design: each judge scores against the real environment output, reducing the task to factual comparison rather than subjective quality assessment.
Since the relative ranking is stable across judges, we select GPT-5.2 as the judge for its highest average Turing-test accuracy.

\section{Experiments}
\label{sec:experiment}

\begin{figure*}[t]
    \centering
    \includegraphics[width=\linewidth]{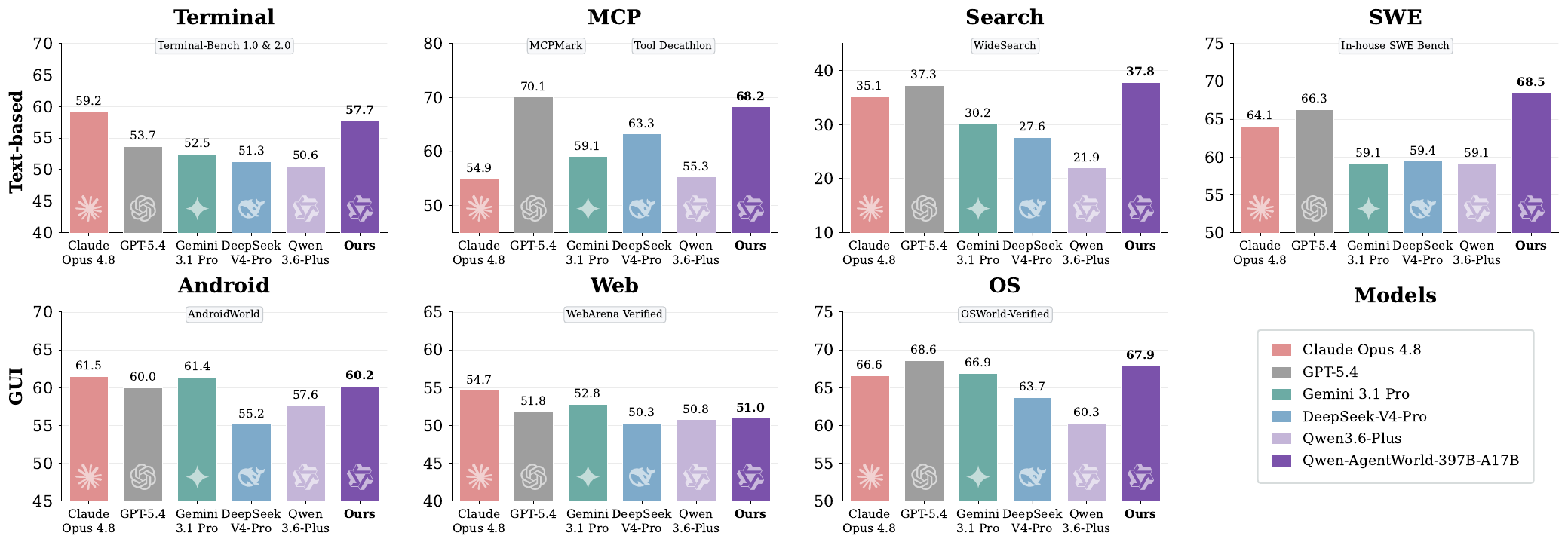}
    \caption{Main results on \bench: five-dimensional rubric mean per domain. \method-397B-A17B achieves the highest overall average among all evaluated models, with consistent advantages on text-based domains and competitive performance on GUI domains.}
    \label{fig:bench_results_bar}
    \vspace{-4mm}
\end{figure*}

We evaluate \method on \bench across all seven domains, covering experimental setup (\S\ref{sec:exp:setup}), main evaluation results (\S\ref{sec:exp:main}), and cross-domain generalization (\S\ref{sec:exp:crossdomain}).
We additionally report supplementary rule-based verification results in Appendix~\ref{sec:appendix:rulebased} that corroborate the main findings.
As previewed in Figure~\ref{fig:bench_results_bar}, \method-397B-A17B achieves the highest overall score.

\subsection{Setup}
\label{sec:exp:setup}

\paragraph{Models.}
We evaluate \method-35B-A3B and \method-397B-A17B, both trained with the three-stage recipe of \S\ref{sec:pipeline}.

\paragraph{Baselines.}
We compare against 14 baselines spanning frontier proprietary models, open-weight models, and Qwen-family models without world-model training:
\begin{itemize}[nosep,leftmargin=*]
  \item \textit{Frontier Proprietary}: Claude Opus~4.8~\citep{opus4.8}, Claude Opus~4.6~\citep{opus4.6}, Claude Sonnet~4.6~\citep{sonnet4.6}, GPT-5.4~\citep{gpt-5.4}, Gemini~3.1~Pro~\citep{gemini3.1pro}.
  \item \textit{Open-Weight}: DeepSeek-V4-Pro~\citep{deepseekai2026deepseekv4}, Kimi~K2.6~\citep{team2026kimi}, GLM-5.1~\citep{zeng2026glm}, MiniMax-M2.7~\citep{minimax2026minimax}.
  \item \textit{Qwen Family (no LWM training)}: Qwen3.5-35B-A3B, Qwen3.5-397B-A17B~\citep{qwen35blog}, Qwen3.6-35B-A3B~\citep{qwen36_35b_a3b}, Qwen3.6-Plus~\citep{qwen36plus}, Qwen3.6-Max-Preview~\citep{qwen36_max_preview}.
\end{itemize}
The Qwen baselines isolate the effect of world-model training: they share the same base architecture but lack the three-stage CPT$\to$SFT$\to$RL pipeline.
In the result tables, the Qwen3.5 base checkpoints are placed alongside \method for direct comparison of the training effect.

\paragraph{Settings.}
All models use temperature~$=$~0.6 with thinking mode enabled where supported.
The maximum generation length and context length are set to the limits supported by each model.
Proprietary models use official API endpoints.
Open-weight models are deployed with SGLang.

\begin{table*}[t]
\caption{\bench main results: five-dimensional rubric mean ($\uparrow$) per domain. The highest and second-best scores per domain are shown in \textbf{bold} and \underline{underlined}, respectively.}
\label{tab:main_results}
\centering
\small
\setlength{\tabcolsep}{3.8pt}
\resizebox{.89\linewidth}{!}{%
\begin{tabular}{@{}ll cccc ccc c@{}}
\toprule
& \multirow{2}{*}{\textbf{Model}} & \multicolumn{4}{c}{\textbf{Text}} & \multicolumn{3}{c}{\textbf{GUI}} & \multirow{2}{*}{\textbf{Avg.}} \\
\cmidrule(lr){3-6} \cmidrule(lr){7-9}
& & \textbf{MCP} & \textbf{Search} & \textbf{Term.} & \textbf{SWE} & \textbf{Android} & \textbf{Web} & \textbf{OS} & \\
\midrule
\multirow{5}{*}{\rotatebox{90}{\scriptsize\textit{Frontier}}}
& Claude Opus~4.8          & 54.93 & 35.14 & \textbf{59.18} & 64.10 & \underline{61.50} & \textbf{54.66} & 66.62 & 56.59 \\
& Claude Opus~4.6          & 69.90 & 29.30 & 57.51 & 64.55 & \textbf{61.74} & 51.42 & \textbf{70.20} & 57.80 \\
& Claude Sonnet~4.6        & \underline{70.00} & 28.79 & 56.98 & 64.52 & 58.03 & 50.78 & 63.17 & 56.04 \\
& GPT-5.4                  & \textbf{70.10} & \underline{37.26} & 53.69 & \underline{66.29} & 60.00 & 51.80 & \underline{68.58} & \underline{58.25} \\
& Gemini~3.1~Pro           & 59.07 & 30.21 & 52.47 & 59.07 & 61.40 & \underline{52.83} & 66.92 & 54.57 \\
\midrule
\multirow{4}{*}{\rotatebox{90}{\scriptsize\textit{Open-weight}}}
& DeepSeek-V4-Pro          & 63.27 & 27.61 & 51.26 & 59.44 & 55.17 & 50.32 & 63.70 & 52.97 \\
& Kimi~K2.6                & 65.23 & 27.48 & 52.54 & 58.77 & 58.93 & 50.20 & 60.80 & 53.42 \\
& GLM-5.1                  & 67.60 & 22.46 & 47.32 & 52.07 & 59.10 & 51.50 & 59.13 & 51.31 \\
& MiniMax-M2.7             & 55.82 & 27.30 & 41.62 & 37.44 & 52.40 & 50.52 & 57.73 & 46.12 \\
\midrule
\multirow{3}{*}{\rotatebox{90}{\scriptsize\textit{Qwen}}}
& Qwen3.6-35B-A3B          & 42.96 & 18.78 & 43.81 & 40.71 & 51.88 & 46.53 & 55.48 & 42.88 \\
& Qwen3.6-Plus             & 55.28 & 21.94 & 50.58 & 59.08 & 57.65 & 50.78 & 60.33 & 50.81 \\
& Qwen3.6-Max-Preview      & 67.01 & 24.71 & 50.86 & 57.11 & 57.74 & 48.58 & 60.95 & 52.42 \\
\midrule
\multirow{4}{*}{\rotatebox{90}{\scriptsize\textit{Ours}}}
& Qwen3.5-35B-A3B          & 57.87 & 25.98 & 46.13 & 47.58 & 53.18 & 47.10 & 56.27 & 47.73 \\
& \method-35B-A3B          & 64.79 & 36.69 & 53.96 & 65.63 & 58.17 & 49.55 & 65.92 & 56.39 \\
\noalign{\vskip 2pt}
\cdashline{2-10}[6pt/3pt]
\noalign{\vskip 2pt}
& Qwen3.5-397B-A17B        & 68.31 & 30.81 & 55.30 & 64.44 & 54.90 & 48.55 & 60.85 & 54.74 \\
& \method-397B-A17B        & 68.24 & \textbf{37.82} & \underline{57.73} & \textbf{68.49} & 60.20 & 50.98 & 67.89 & \textbf{58.71} \\
\bottomrule
\end{tabular}%
}
\end{table*}

\subsection{Main Results}
\label{sec:exp:main}

Table~\ref{tab:main_results} reports the five-dimensional rubric mean across all seven domains.
Each predicted observation is scored on five rubric dimensions---\textit{Format}, \textit{Factuality}, \textit{Consistency}, \textit{Realism}, and \textit{Quality}---using a 1--5 scale, and then normalized to 0--100.
\method-397B-A17B achieves the highest overall average (58.71), surpassing GPT-5.4 (58.25) and all other frontier models.
On text-based domains, \method-397B-A17B leads with an average of 58.07, outperforming GPT-5.4 (56.84) by 1.23 points.
The advantage is most pronounced on Terminal (57.73 vs.\ 53.69) and SWE (68.49 vs.\ 66.29), the two domains where predictions require accurate modeling of code execution state and tool API behavior.
On GUI domains, Claude Opus~4.8 (60.93) and Claude Opus~4.6 (61.12) lead, followed by GPT-5.4 (60.47) and Gemini~3.1~Pro (60.04), with \method-397B-A17B ranking fifth (59.69).
The gap reflects an advantage from multimodal pre-training that text-only world modeling does not fully capture.

\paragraph{Effect of World-Model Training.}
Comparing \method with its base checkpoints isolates the contribution of the three-stage pipeline.
At the 397B scale, the overall average rises from 54.74 to 58.71.
At 35B, the gain is 8.66 points (47.73 to 56.39), lifting \method-35B-A3B above Claude Sonnet~4.6 (56.04) by 0.35 points.
The improvement is consistent across both text and GUI domains: at 397B, the text-domain average rises by 3.35 and the GUI-domain average by 4.92.
These gains are not explained by the base model's general capability alone, since the Qwen3.6 checkpoints without LWM training (Qwen3.6-Plus: 50.81, Qwen3.6-Max-Preview: 52.42) score well below \method despite sharing the same architecture family.

\paragraph{Domain-Level Observations.}
Search is the most challenging domain for all models: the best score (37.82) is roughly half the best score on SWE (68.49) or MCP (70.10).
Search requires modeling constantly evolving web content, and factual consistency across long retrieval chains remains difficult for all models.
On MCP, Claude Opus~4.6 and GPT-5.4 tightly at the top, reflecting their tool-use specialization.

\subsection{Cross-Domain Generalization}
\label{sec:exp:crossdomain}

\begin{figure*}[t]
    \centering
    \includegraphics[width=\linewidth]{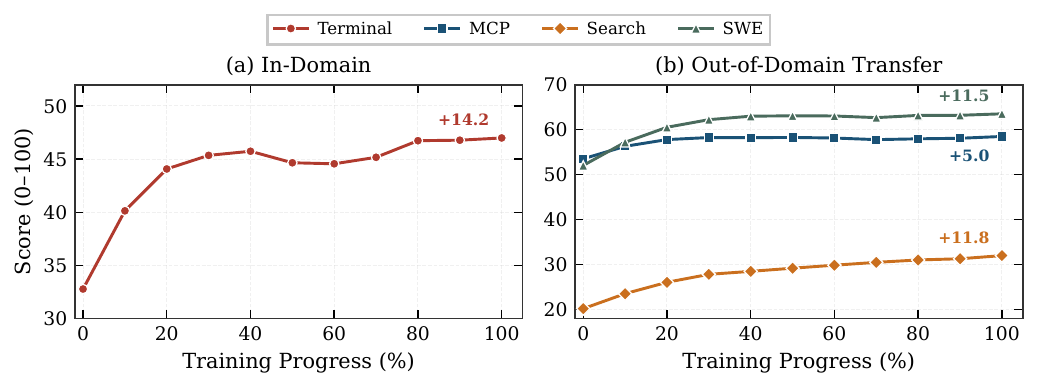}
    \caption{Cross-domain generalization when training Stage~3 (RL) on Terminal data alone. (a)~Terminal (in-domain) improves by +14.2 points over the SFT baseline. (b)~All three held-out domains improve without receiving any domain-specific training signal: MCP~(+5.0), SWE~(+11.5), and Search~(+11.8).}
    \label{fig:transfer_text}
    \vspace{-4mm}
\end{figure*}

This experiment tests whether \method learns generalizable language world knowledge or domain-specific environment behaviors.
We train Stage~3 (RL) on Terminal data alone and evaluate on all four text-based domains every 10 steps.
Figure~\ref{fig:transfer_text} shows the performance when training Stage~3 on Qwen3.5-35B-A3B-SFT, an in-house Qwen checkpoint that has undergone CPT and general-purpose SFT but no world-model training, using Terminal LWM data alone.
Terminal improves from 32.8 to 47.0 (+14.2) within 100 RL steps.
All three held-out text-based domains improve in parallel: SWE gains +11.5 (52.0 $\to$ 63.5), Search gains +11.8 (20.2 $\to$ 32.0), and MCP gains +5.0 (53.5 $\to$ 58.5) despite already starting from a high baseline.
The transfer is non-trivial: Terminal shell commands and MCP tool calls differ in syntax, state representation, and response structure, yet the gains emerge within the first 10 RL steps and remain stable throughout training.
This pattern suggests that RL reinforces generalizable world knowledge, how environments respond to actions, how errors propagate, how state transitions compose across turns, rather than domain-specific output formats.

\section{Applications}
\label{sec:applications}

We investigate two complementary paradigms through which \method enhances general agents.
\method enables infinite environment scaling and controllable simulation~(\S\ref{sec:app:simulator}), allowing agent training to scale to domains and conditions beyond what real environments can provide.
Moreover, LWM training increases the upper bound of downstream agent performance as an agent foundation model~(\S\ref{sec:app:foundation}).

\subsection{Application I: Environment Simulator}
\label{sec:app:simulator}

\begin{tcolorbox}[takeawaysbox]
\small
As a standalone simulator, \method provides \emph{\textcolor{colorscale}{scalability}} and \emph{\textcolor{colorctrl}{controllability}} that real environments cannot offer:
\begin{itemize}[leftmargin=1.2em,itemsep=1pt,topsep=2pt]
\item \textbf{Zero-shot environment generalization.} \method simulates $4k$ OpenClaw environments for agentic RL entirely absent from training, yielding gains of +4.3 on Claw-Eval and +7.1 on QwenClawBench with no domain-specific adaptation.
\item \textbf{Controlled simulation matters.} Controllable perturbations expose agent weaknesses that real-world rarely produce, lifting MCPMark by +12.3 and WideSearch by +16.3, far exceeding uncontrolled Sim RL.
\item \textbf{Surpassing real-environment training.} Controllable Sim RL exceeds Real RL trained using a live search engine (50.3\% vs.\ 45.6\%), while shaping more targeted agent behavior through adversarial snippet design.
\item \textbf{Fictional worlds work.} Agents trained in fully invented, self-consistent worlds generalize to real search tasks, while structurally preventing the agent from confusing simulated facts with real-world knowledge.
\item \textbf{State is the bottleneck.} Sim RL effectiveness depends on providing the world model with a sufficiently detailed initial state; without it, simulation fidelity degrades and downstream gains diminish.
\end{itemize}
\end{tcolorbox}

In this application, \method serves as a standalone environment simulator: the policy agent and the world model are separate models.
\method has two properties that real environments do not provide: \emph{\textcolor{colorscale}{scalability}} (\S\ref{sec:app:scaling}) and \emph{\textcolor{colorctrl}{controllability}} (\S\ref{sec:app:controllable}).

\subsubsection{Generalizable Environment Scaling}
\label{sec:app:scaling}

Leveraging the world knowledge acquired during CPT and the environment modeling capabilities learned across seven training domains, \method can generalizably simulate a wide range of realistic, highly specialized real-world environments.
We study this capability on OpenClaw~\citep{openclaw2026}, an open-source general-purpose real-world AI agent platform whose tasks are drawn from real user multi-step digital workflows rather than curated benchmarks, spanning scheduling, coding, email triage, browser automation, and file management.
Since OpenClaw is entirely out of distribution, it provides a natural testbed for evaluating whether a language world model can generalize to new interactive environment families.
In this setting, \method synthesizes diverse OpenClaw-style environments from a small set of real interaction traces, without any domain-specific adaptation.

\paragraph{Experimental Setup.}
We begin with a small pool of real Claw Agent trajectories as anchors for realistic workflows.
Each trajectory is distilled into a reusable \textbf{seed scenario} that captures the task-relevant \textbf{initial state} (e.g., installed applications, file-system layout, account configuration, and relevant workspace contents) together with the corresponding \textbf{user query}.
From these seeds, we synthesize $4k$ simulated training environments along two complementary axes.
On the environment side, we vary concrete states while preserving the underlying workflow structure; on the task side, we rephrase intents, adjust difficulty, and compose multi-step goals so that the agent sees diverse but realistic OpenClaw-style tasks.
This expansion evaluates the core value of environment scaling: whether a world model can transform scarce real-environment evidence into broad, realistic interaction coverage for downstream agent training.
For each synthesized task, we generate a rubric-based verifier that describes the expected terminal conditions, providing an automatic reward signal for Sim RL.
We use \method-397B-A17B as the simulator to test its generalizable environment simulation capability, with Qwen3.6-Plus as a baseline for ablation.
Based on Qwen3.5-35B-A3B~\citep{qwen35blog}, we conduct multi-turn Sim RL with up to 50 interaction turns and evaluate on Claw-Eval~\citep{ye2026claw} and QwenClawBench~\citep{qwenclawbench}.
For both benchmarks, we report Avg@3 scores with a maximum sequence length of 256k tokens.

\paragraph{Results.}
As shown in Table~\ref{tab:openclaw_simrl}, Sim RL with \method-397B-A17B as the simulator improves the Claw-Eval score from 65.4 to 69.7 (+4.3) and QwenClawBench from 47.9 to 55.0 (+7.1), confirming that \method can scalably simulate a wide range of real-world scenarios and that the resulting policies transfer effectively to real-world domains absent from world-model training.
\textbf{Comparing the two simulators reveals the importance of world-model quality:} using Qwen3.6-Plus as the simulator yields only marginal gains, whereas \method-397B-A17B, trained explicitly for environment simulation, achieves substantially larger gains on both benchmarks, confirming that our dedicated LWM training pipeline is critical for building high-fidelity simulators that produce effective Sim RL gains.

\begin{table}[ht]
\caption{Sim RL on simulated OpenClaw environments. Qwen3.5-35B-A3B is trained via Sim RL using different environment simulators. $\Delta$ reports gains over the base model.}
\label{tab:openclaw_simrl}
\centering
\small
\begin{tabular}{@{}l cc@{}}
\toprule
\textbf{Model} & \textbf{Claw-Eval} & \textbf{QwenClawBench} \\
\midrule
Qwen3.5-35B-A3B          & 65.4 & 47.9 \\
\quad\quad Sim RL~(w/ Qwen3.6-Plus)           & 66.7 & 47.8 \\
\quad\quad Sim RL~(w/ \method-397B-A17B)           & 69.7 & 55.0 \\
\midrule
\quad\quad $\Delta$           & +4.3 & +7.1 \\
\bottomrule
\end{tabular}
\end{table}

\subsubsection{Controllable Simulation}
\label{sec:app:controllable}

Controllable simulation uses natural-language instructions to shape how \method behaves during training, at both the trajectory and turn levels.
Concurrent work on controllable tool-use environments~\citep{xu2026controllable} shares the motivation that oracle-preserving augmentations improve agent robustness.
We validate two complementary modes of controllability via Sim RL on MCP and Search agents respectively.
The base models are Qwen3.5-35B-A3B-SFT and Qwen3.5-397B-A17B-SFT, internal Qwen checkpoints that have undergone continual pre-training and general-purpose supervised fine-tuning but no world-model training:
\begin{itemize}[leftmargin=1.5em,itemsep=2pt]
    \item \textbf{Environment Adaptation}, where the simulator’s behavior and style are adjusted through instructions to inject targeted perturbations, thereby systematically exposing agent weaknesses.
    Control instructions then modify the simulation style and content to produce conditions more extreme or targeted than what real deployments provide, such as intermittent API errors, paginated responses requiring follow-up calls, incomplete intermediate results that force multi-step retrieval, and partial failures in batch operations.
    By systematically exposing the agent to its weak points, controllable Sim RL trains agents that surpass those trained on unperturbed real interactions alone.

    \item \textbf{Fictional-World Construction}, where the simulator generates an entirely fictional environment that is structurally realistic yet factually disjoint from the real world.
    For Search, we instruct \method to simulate a fully fictional, self-consistent world from a compact initial specification.
    LiteResearcher~\citep{li2026literesearcher} independently constructs virtual search worlds for agentic RL with sim-to-real transfer, sharing the same insight that search agents can be trained effectively in simulated environments.
    Fictional-world construction offers two structural advantages for Search-domain Sim RL.
    First, since the answers exist only within the fictional setting, the agent cannot bypass the search tool by answering from parametric memory and must learn to search effectively.
    Second, because all facts are invented with no real-world counterpart, the agent cannot confuse training-time search results with real-world knowledge. By contrast, directly simulating a real search engine risks injecting fabricated but plausible-looking facts that the agent later treats as true.
    \method faithfully follows the initial setting and coherently extends it into a logically consistent reality.
    For instance, a fictional 2030 scenario may specify that 430 people have migrated to Mars.
    The world model then generates consistent demographic records, news articles, and search results grounded in this premise.
    Surprisingly, Search agents trained entirely within these fictional environments generalize effectively to real-world search tasks with significant performance gains.
\end{itemize}

\paragraph{MCP: Environment Adaptation.}
We synthesize environment simulation system prompts from in-house MCP tool-use trajectories collected during the RL data pipeline: frontier agents (Claude Opus 4.6 with extended thinking, DeepSeek-v4-Pro) solve diverse cross-service tasks in real MCP environments, and only trajectories passing automated verification are retained.
Each prompt consists of three components: an \emph{initial state} that specifies the tool schemas and server configuration; an \emph{environment summary} that summarizes the hidden environment state behind the trajectory, such as database contents, permission settings, and service availability; and \emph{controllable simulation instructions} that define how the simulator should respond at each turn, including injected errors, withheld intermediate results, response formats, and the final tool and environment state required for task success.
Together, these components provide the simulator with realistic tool settings, environment context, and clear success criteria.
Since the agent is trained in a simulated environment, a natural concern is whether its gains arise from exploiting simulator-specific artifacts or from ground-truth leakage through the simulation instructions.
However, evaluation on out-of-distribution benchmarks in the real environment eliminates this concern, and the SimRL training data are entirely disjoint from the evaluation queries.
Tool Decathlon~\citep{li2025tool} covers 32 software applications and features multi-step tasks. Performance is measured by execution-verified success rate.
MCPMark~\citep{wu2025mcpmark} evaluates agents on 127 multi-step tasks spanning multiple MCP server categories, using script-verified pass@1 as the metric.

As shown in Table~\ref{tab:controllable_simrl_mcp}, standard Sim RL without control instructions provides no meaningful gain, and Tool Decathlon even drops from 32.4 to 31.5, because the simulator lacks sufficient grounding to produce faithful responses.
With controllable simulation, Tool Decathlon improves by +3.7 and MCPMark by +12.3.
Controllability is not merely a factor in the magnitude of improvement but a prerequisite for Sim RL to work at all in this domain: without grounded simulation instructions, the training signal is too noisy to yield any gain.
The larger gain on MCPMark suggests that controllable simulation is especially effective for tasks that require many tool calls and careful handling of intermediate tool results.

\begin{table}[ht]
\caption{Controllable Sim RL results on Tool Decathlon and MCPMark. ``w/ \method-397B-A17B controlled'' adds targeted environment control instructions during Sim RL.}
\label{tab:controllable_simrl_mcp}
\centering
\small
\setlength{\tabcolsep}{3pt}
\resizebox{\linewidth}{!}{%
\begin{tabular}{@{}l c cccccc c@{}}
\toprule
\multirow{2}{*}[-0.4em]{\textbf{Model}} & \multirow{2}{*}[-0.4em]{\shortstack{\textbf{Tool}\\\textbf{Decathlon}}} & \multicolumn{7}{c}{\textbf{MCPMark}} \\
\cmidrule(lr){3-9}
& & \shortstack{\textbf{Filesys.}} & \shortstack{\textbf{GitHub}} & \shortstack{\textbf{Postgres}} & \shortstack{\textbf{Notion}} & \shortstack{\textbf{Playwright}} & \shortstack{\textbf{WebArena}} & \shortstack{\textbf{Avg.}} \\
\midrule
Qwen3.5-35B-A3B-SFT & 32.4 & 16.7 & 17.4 & 28.6 & 25.0 & 0.0 & 33.3 & 21.5 \\
\quad Sim RL (w/ \method-397B-A17B) & 31.5 & 16.7 & 17.4 & 42.9 & 32.1 & 0.0 & 28.6 & 24.6 \\
\quad Sim RL (w/ \method-397B-A17B controlled) & \textbf{36.1} & \textbf{30.0} & 17.4 & \textbf{47.6} & \textbf{42.9} & 0.0 & \textbf{38.1} & \textbf{33.8} \\
\midrule
\quad $\Delta$ & +3.7 & +13.3 & +0.0 & +19.0 & +17.9 & +0.0 & +4.8 & +12.3 \\
\bottomrule
\end{tabular}%
}
\end{table}

\begin{figure}[!h]
\centering
\begin{subfigure}[b]{\linewidth}
    \centering
    \includegraphics[width=\linewidth]{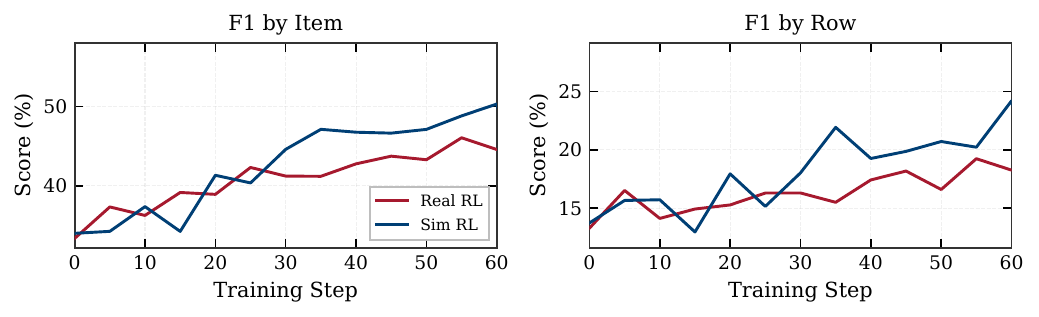}
    \caption{F1 by Item and F1 by Row on WideSearch validation set.}
    \label{fig:sim_vs_real_f1}
\end{subfigure}
\\[4pt]
\begin{subfigure}[b]{\linewidth}
    \centering
    \includegraphics[width=\linewidth]{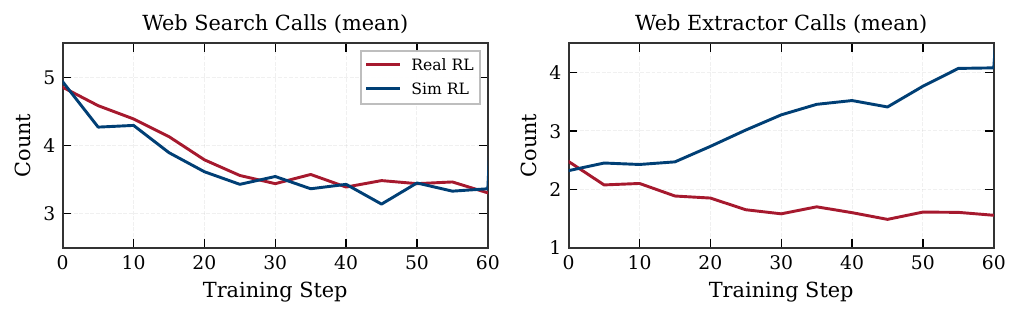}
    \caption{Mean tool call frequency per trajectory.}
    \label{fig:sim_vs_real_tool}
\end{subfigure}
\caption{Controllable Sim RL vs.\ Real RL (trained against a live search engine) on WideSearch during the first 60 training steps. Both experiments use Qwen3.5-35B-A3B-SFT as the base model.}
\label{fig:sim_vs_real_rl}
\end{figure}

\paragraph{Search: Fictional-World Construction.}
We construct $1k$ self-contained fictional environments using a multi-agent synthesis framework, each anchored by a large relational database (300--500 rows) of internally consistent fictional structured facts.
The pipeline collects high-quality domain documents as grounding material, instantiates fictional database schemas from these documents, populates them via LLM-guided generation, executes SQL to extract ground-truth answers, and reverse-generates natural-language queries.
Four synthesis strategies (time-shifted, granular long-tail, private simulation, realistic grounded) keep the data entirely fictional yet realistic: for instance, a time-shifted environment contains a 2029 smartphone market ranking with real brand names but non-existent model numbers.
Automated retrieval checks and dual rule-based\&LLM validation ensure that synthetic facts conform to real-world patterns (e.g., consistent price ranges, realistic temporal trends) while remaining unsearchable to prevent inject hallucinations.
Controllable simulation instructions further shape \method's turn-level behavior: the world model returns only information relevant to the current search query without revealing the full answer, so the agent must reformulate queries, cross-reference sources, and iteratively aggregate partial results across multiple search rounds.
This design increases the number of tool calls the agent must issue per trajectory, as surface-level search snippets no longer suffice and the agent must invoke page extraction to retrieve complete information (Figure~\ref{fig:sim_vs_real_tool}).
WideSearch~\citep{wong2025widesearch} is a broad information-seeking benchmark of 200 manually curated collection tasks across 15+ domains.
The agent must gather structured tabular answers from the web. The metrics are item-level F1 (per-cell accuracy) and row-level F1 (per-entity completeness).

Table~\ref{tab:controllable_simrl} reports results at both model scales.
On Qwen3.5-35B-A3B-SFT, controllable Sim RL raises F1 by Item from 34.02 to 50.31 (+16.29) and F1 by Row from 13.72 to 24.21 (+10.49).
On Qwen3.5-397B-A17B-SFT, where the base model already achieves 70.11 F1 by Item, Sim RL still yields gains of +3.87 and +6.05 on the two metrics respectively.
These results are notable because the training environments are entirely fictional: every search result, web page, and factual record is invented by \method from scratch, with no connection to any real search engine or real-world database.
The agent never interacts with a real environment during Sim RL, yet the capabilities it acquires (query reformulation, multi-source cross-referencing, and iterative result aggregation) transfer directly to real-world search tasks on WideSearch.
This demonstrates controllability at the level of entire environments: \method constructs and simulates complete, self-consistent fictional worlds that are sufficiently realistic to train effective search agents.
The +16.29 F1 by Item gain at 35B shows that fictional-world Sim RL can close a substantial capability gap for weaker models, while the +3.87 gain at 397B, where the base already achieves 70.11, confirms that the approach remains effective even at frontier scale.

\begin{table}[ht]
\caption{Controllable Sim RL results on WideSearch, using fictional-world simulation.}
\label{tab:controllable_simrl}
\centering
\small
\begin{tabular}{@{}l cc@{}}
\toprule
\multirow{2}{*}{\textbf{Model}} & \multicolumn{2}{c}{\textbf{WideSearch}} \\
\cmidrule(lr){2-3}
& \textbf{F1 Item} & \textbf{F1 Row} \\
\midrule
Qwen3.5-35B-A3B-SFT                          & 34.02 & 13.72 \\
\quad\quad Sim RL (w/ \method controlled) & 50.31 & 24.21 \\
\midrule
\quad\quad $\Delta$                           & +16.29 & +10.49 \\
\midrule
Qwen3.5-397B-A17B-SFT                        & 70.11 & 45.69 \\
\quad\quad Sim RL (w/ \method controlled) & 73.98 & 51.74 \\
\midrule
\quad\quad $\Delta$                           & +3.87 & +6.05 \\
\bottomrule
\end{tabular}
\end{table}

\paragraph{Real RL vs. Sim RL}
The MCP results in Table~\ref{tab:controllable_simrl_mcp} reveal that standard Sim RL without control instructions yields negligible improvement (Tool Decathlon actually drops from 32.4 to 31.5), whereas controlled Sim RL produces substantial gains.
This gap is even more pronounced in the Search domain, where the world model must simulate an entire search engine returning plausible results for fictional facts that no real index contains.
We therefore adopt a fully controllable simulation design from the outset:
the simulation instruction explicitly requires \method to withhold complete answers from search snippets, returning only partial, query-relevant information.
Specifically, snippets are required to \textbf{``tease information without fully answering''}, and each result page mixes highly relevant, moderately relevant, and background results.
This design forces the agent to issue follow-up \texttt{web\_extractor} calls to retrieve full page content rather than relying on surface-level snippets alone.

Figure~\ref{fig:sim_vs_real_rl} compares controllable Sim RL against Real RL trained with a live search engine on WideSearch.
In terms of task performance (Figure~\ref{fig:sim_vs_real_f1}), Sim RL tracks or slightly exceeds Real RL throughout the overlapping range: F1 by Item reaches 50.3\%, compared to 45.6\% for Real RL.
\textbf{The more informative signal comes from agent behavior.}
Figure~\ref{fig:sim_vs_real_tool} (left) shows that both training regimes reduce \texttt{web\_search} calls from ${\sim}5$ to ${\sim}3.5$ per trajectory, indicating that both agents learn to issue more targeted queries.
The \texttt{web\_extractor} calls (right), however, diverge sharply: Sim RL increases usage from 2.5 to 4.0 per trajectory, while Real RL decreases it from 2.5 to 1.5.
This divergence directly reflects the controllable simulation design.
Because the simulated search snippets deliberately withhold detailed content, the Sim-RL-trained agent learns that extracting full pages is necessary for assembling complete answers.
In contrast, the Real-RL-trained agent finds that real search snippets often contain sufficient information, and thus learns to skip extraction.
The result confirms that controllable simulation can shape agent behavior in targeted ways: by constructing adversarial environment conditions where specific capabilities are required, Sim RL trains those capabilities more effectively than uncontrolled real-environment training.

\subsection{Application II: Agent Foundation Model}
\label{sec:app:foundation}

\begin{tcolorbox}[takeawaysbox]
\small
LWM training unifies the world model and the agent, instilling next-state prediction as an internalized reasoning capability:
\begin{itemize}[leftmargin=1.2em,itemsep=1pt,topsep=2pt]
\item \textbf{Radical cross-task generalization.} Single-turn, non-agentic LWM RL warm-up with no tool calls transfers to multi-turn, tool-calling agentic tasks across seven benchmarks of five domains.
\item \textbf{Domain generalization.} Gains emerge on two completely out-of-distribution domains entirely absent from LWM training (+11.3 on Claw-Eval, +9.7 on QwenClawBench, and +9.0 on BFCL~v4), confirming transferable capabilities rather than domain-specific shortcuts.
\item \textbf{Next-state prediction as meta-reasoning pattern.} LWM training teaches the agent to mentally simulate environment responses before acting, which generalizes across task formats and domains.
\end{itemize}
\end{tcolorbox}

In Application~I the agent and world model are separate models.
Here we unify them: the same model that selects actions~(agent) also predicts environment states~(world model).
The underlying mechanism is that LWM training enables the agent to mentally simulate the consequences of a candidate action before committing, effectively using world modeling as an internal planning step that improves action quality.
This aligns with the unified \textbf{world-model--actor} architecture envisioned by \citet{lecun2022path} and echoes the \textbf{World Action Model} paradigm emerging in vision-language-action research~\citep{ye2026world}.

LWM RL warm-up familiarizes the agent with how environment states evolve in response to user actions: how different tool calls and search queries yield different responses, which tools are more effective, and how state transitions propagate across turns.
This amounts to learning next-state prediction as a meta-reasoning pattern, in which the agent internally simulates what will happen before deciding what to do.
We validate this effect by running LWM RL on Qwen3.5-35B-A3B-SFT, which is fundamentally a \textbf{single-turn task that involves reasoning with no tool calls or multi-turn interaction} (predicting the next environment state given a user action).
After warm-up, we evaluate the same model directly on \textbf{multi-turn, tool-calling agentic tasks} across seven benchmarks without any additional fine-tuning, including three out-of-domain benchmarks absent from LWM training.
All benchmarks use a maximum sequence length of 256k tokens.
Claw-Eval, QwenClawBench, SWE-Bench Verified, and SWE-Bench Pro scores are averaged over 3 independent rollouts; Terminal-Bench~2.0 scores are averaged over 5 runs; BFCL~v4 uses a single rollout.
SWE-Bench Verified and SWE-Bench Pro are evaluated with an internal agent scaffold (bash and file-edit tools); we correct several problematic tasks in the public set of SWE-Bench Pro and evaluate all baselines on the refined benchmark.
Terminal-Bench~2.0 is evaluated under the Terminus-2 harness with a 3-hour timeout and 8 CPU / 32\,GB RAM.

\begin{table*}[!h]
\caption{Agent foundation model: LWM RL warm-up on single-turn, non-agentic trajectories transfers to multi-turn, tool-calling agentic tasks. No additional fine-tuning is applied after LWM RL.}
\label{tab:foundation}
\centering
\small
\setlength{\tabcolsep}{2.5pt}
\resizebox{\textwidth}{!}{%
\begin{tabular}{@{}l ccc cc c cc ccccccc@{}}
\toprule
& \multicolumn{5}{c}{\textbf{In Domain}} & & \multicolumn{9}{c}{\textbf{Out of Domain}} \\
\cmidrule(lr){2-6} \cmidrule(lr){8-16}
\multirow{2}{*}{\textbf{Model}} & \multirow{2}{*}{\shortstack{\textbf{Terminal-}\\\textbf{Bench 2.0}}} & \multirow{2}{*}{\shortstack{\textbf{SWE-Bench}\\\textbf{Verified}}} & \multirow{2}{*}{\shortstack{\textbf{SWE-Bench}\\\textbf{Pro}}} & \multicolumn{2}{c}{\textbf{WideSearch}} & & \multirow{2}{*}{\shortstack{\textbf{Claw-}\\\textbf{Eval}}} & \multirow{2}{*}{\shortstack{\textbf{QwenClaw}\\\textbf{Bench}}} & \multicolumn{7}{c}{\textbf{BFCL v4}} \\
\cmidrule(lr){5-6} \cmidrule(lr){10-16}
& & & & \textbf{F1 Item} & \textbf{F1 Row} & & & & \textbf{Web} & \textbf{Mem.} & \textbf{Multi-T.} & \textbf{No Live} & \textbf{Live} & \textbf{Hallu.} & \textbf{Avg.} \\
\midrule
Qwen3.5-35B-A3B-SFT & 33.25 & 64.47 & 42.18 & 33.38 & 13.27 & & 53.60 & 39.76 & 67.50 & 54.19 & 47.25 & 78.17 & 77.28 & 82.32 & 62.29 \\
\quad\quad w/ LWM RL & 39.55 & 67.86 & 47.42 & 46.17 & 20.14 & & 64.88 & 49.43 & 83.00 & 60.65 & 60.38 & 80.56 & 79.13 & 84.38 & 71.25 \\
\midrule
\quad\quad $\Delta$   & +6.30 & +3.39 & +5.24 & +12.79 & +6.87 & & +11.28 & +9.67 & +15.50 & +6.46 & +13.13 & +2.39 & +1.85 & +2.06 & +8.96 \\
\bottomrule
\end{tabular}}%
\end{table*}

Table~\ref{tab:foundation} reports the results.
On the four in-domain benchmarks, LWM RL raises WideSearch F1 by Item from 33.38 to 46.17 (+12.79) and F1 by Row from 13.27 to 20.14 (+6.87), Terminal-Bench~2.0~\citep{merrill2026terminal} accuracy from 33.25 to 39.55 (+6.30), SWE-Bench Verified~\citep{jimenez2024swe} resolve rate from 64.5 to 67.9 (+3.4), and SWE-Bench Pro~\citep{deng2025swe} resolve rate from 42.2 to 47.4 (+5.2).
The gains extend to three out-of-domain benchmarks: Claw-Eval~\citep{ye2026claw} improves from 53.6 to 64.9 (+11.3), QwenClawBench~\citep{qwenclawbench} from 39.8 to 49.4 (+9.7), and BFCL~v4~\citep{patil2025berkeley} from 62.3 to 71.3 (+9.0).
The consistent gains across all seven benchmarks are striking.
Even the in-domain results represent a substantial distribution shift: LWM RL trains on single-turn next-state prediction with no tool calls, yet the improvements transfer to multi-turn, tool-calling agentic tasks that require iterative planning, tool selection, and result aggregation, and the benchmark queries are strictly disjoint from LWM training data.
The out-of-domain results demonstrate an even more thorough generalization: the LWM training pipeline contains no Claw or function-calling data whatsoever, yet gains of +11.3, +9.7, and +9.0 emerge on domains entirely absent from world-model training, confirming that LWM warm-up instills transferable agent capabilities rather than domain-specific shortcuts.
Concurrent work by \citet{shrivastava2026echo} provides independent validation from a complementary direction: adding an auxiliary cross-entropy loss on environment observation tokens during agent RL trains the agent to predict terminal outputs as a byproduct of policy optimization, roughly doubling performance on Terminal-Bench~2.0.
This confirms that world-modeling capabilities, whether from dedicated pre-training or auxiliary training signals, consistently transfer to agent performance.
Future work may explore combining LWM warm-up with auxiliary world-modeling objectives during agent training for compounding benefits.

\begin{wrapfigure}{r}{0.4\linewidth}
\centering
\vspace{-4pt}
\includegraphics[width=\linewidth]{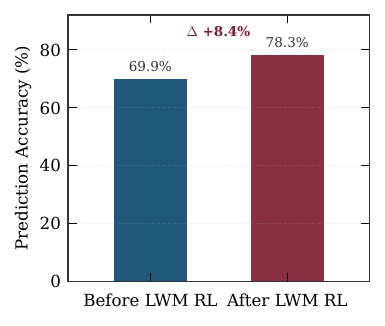}
\caption{Environment prediction accuracy on Terminal-Bench 2.0 trajectories.}
\label{fig:prediction-accuracy}
\vspace{-10pt}
\end{wrapfigure}

\paragraph{Insights: Prediction-Driven Action Refinement.}
To understand \emph{why} LWM warm-up transfers to agentic tasks, we examine the reasoning traces produced during multi-turn agentic evaluation.
We find that the RL-trained model systematically performs \textbf{mental simulation} of environment responses before executing actions, using its internalized world model to predict outcomes, identify infeasible approaches, and refine its action plan---all within the thinking trace and before any real execution.

Figure~\ref{fig:case-prediction-refine} illustrates this prediction-driven refinement on the \texttt{mailman} task from Terminal-Bench~2.0, where both models encounter the same Postfix recipient-rejection error.
The model after LWM RL correctly predicts that configuring \texttt{transport\_maps} alone will not work because Postfix rejects unknown recipients \emph{before} consulting transport routing, enabling it to refine its action toward modifying \texttt{local\_recipient\_maps} and apply a targeted fix.
In contrast, the model before LWM RL incorrectly predicts that transport routing precedes recipient validation, preventing productive action refinement and leading to futile exploration before timing out.

To quantify this effect, we first identify interaction turns in which the model's thinking trace contains an explicit prediction of the environment's next state.
For each such turn, we compare the predicted environment response against the actual environment feedback received after execution, and mark the prediction as correct if the two are semantically consistent.
The prediction accuracy is the fraction of correctly predicted turns among all turns that contain an explicit prediction.
Figure~\ref{fig:prediction-accuracy} presents the results.
RL training improves prediction accuracy from 69.9\% to 78.3\% (+8.4\%), confirming that the model's internal world model becomes more faithful through LWM training.
These results provide direct evidence that the performance gains in Table~\ref{tab:foundation} stem from an improved ability to \emph{predict before acting}---a meta-reasoning capability that generalizes across tasks and domains.

\begin{figure*}[t]
\centering
\small
\begin{tcolorbox}[
  colback=gray!3, colframe=black!50, boxrule=0.5pt,
  title={\textbf{Case Study: Accurate Environment Prediction Enables Action Refinement}},
  fonttitle=\bfseries\small, boxrule=0.5pt, arc=2pt
]

\textbf{Task:} \texttt{mailman} --- Configure a Postfix + Mailman3 mailing list server so that emails to \texttt{reading-group-join@local.edu} are routed to Mailman's LMTP service for subscription processing.

\vspace{4pt}
\hrule
\vspace{6pt}

\textbf{Shared Obstacle (both models encounter the same error):}

\vspace{2pt}
\begin{quote}
\ttfamily\scriptsize
\texttt{550 5.1.1 <reading-group-join@local.edu>: Recipient address rejected: User unknown in local recipient table}
\end{quote}

\vspace{2pt}
\noindent Both models configure \texttt{transport\_maps} to route list addresses to Mailman's LMTP (\texttt{lmtp:[127.0.0.1]:8024}). The transport lookup succeeds, yet Postfix still rejects the recipient. Resolving this requires an accurate world model of Postfix's internal processing pipeline.

\vspace{6pt}
\hrule
\vspace{6pt}

\definecolor{casebefore}{HTML}{0c4a6e}
\definecolor{caseafter}{HTML}{7f1d34}
\begin{tcolorbox}[colback=casebefore!5!white, colframe=casebefore, boxrule=0.4pt, arc=1pt, title={\textbf{Before LWM RL}}, fonttitle=\small]

\begin{quote}
\ttfamily\scriptsize
\textcolor{blue!70!black}{[Prediction]} If I configure \texttt{transport\_maps} to route to LMTP, Postfix will\\
use that transport to deliver the mail. The recipient validation against local\\
tables happens \textbf{after} transport routing decides the delivery path.\\
$\Rightarrow$ The transport config should be sufficient. \textcolor{casebefore}{$\times$ Incorrect}\\[4pt]
\textcolor{gray}{[Action] Tries virtual\_alias\_maps, virtual\_transport, relay configs...}\\
\textcolor{gray}{Oscillates between transport\_maps, virtual maps, and relay configurations.}\\
\textcolor{gray}{\textbf{Never modifies local\_recipient\_maps. Task fails (timeout).}}
\end{quote}
\end{tcolorbox}

\vspace{4pt}

\begin{tcolorbox}[colback=caseafter!5!white, colframe=caseafter, boxrule=0.4pt, arc=1pt, title={\textbf{After LWM RL}}, fonttitle=\small]

\begin{quote}
\ttfamily\scriptsize
\textcolor{blue!70!black}{[Prediction]} \texttt{local.edu} is in \texttt{mydestination}. If I only configure\\
\texttt{transport\_maps} without modifying \texttt{local\_recipient\_maps}, Postfix will reject\\
the recipient at the validation stage \textbf{before} transport routing is ever consulted.\\
$\Rightarrow$ The mail will never reach LMTP. \textcolor{caseafter}{$\checkmark$ Correct}\\[4pt]
\textcolor{orange!70!black}{[Refinement]} The fix must target \texttt{local\_recipient\_maps}, not \texttt{transport\_maps}.\\
\hspace{0.5cm} $\rightarrow$ Add Mailman addresses to the local recipient table so they pass validation\\
\hspace{0.5cm} before transport routes them to LMTP.\\[4pt]
\textcolor{teal!80!black}{[Action]} Create \texttt{hash:/var/lib/mailman3/data/local\_recipients} with list addresses. \textcolor{caseafter}{$\checkmark$}\\
\textcolor{gray}{All tests pass.}
\end{quote}
\end{tcolorbox}

\end{tcolorbox}

\caption{Case study of prediction-driven action refinement on the \texttt{mailman} task from Terminal-Bench~2.0.}
\label{fig:case-prediction-refine}
\end{figure*}

\section{Analysis}
\label{sec:analysis}

The preceding sections establish that \method achieves high simulation fidelity (Table~\ref{tab:main_results}) and enables effective downstream agent training (\S\ref{sec:applications}).
This section asks \emph{how} the model arrives at accurate predictions and \emph{what} changes during RL training.
We examine LWM reasoning patterns in the model's chain-of-thought (\S\ref{sec:analysis:reasoning}) and micro-level fidelity improvements driven by RL (\S\ref{sec:analysis:fidelity}).

\subsection{LWM Reasoning Patterns}
\label{sec:analysis:reasoning}

\begin{figure*}[t]
    \centering
    \includegraphics[width=\linewidth]{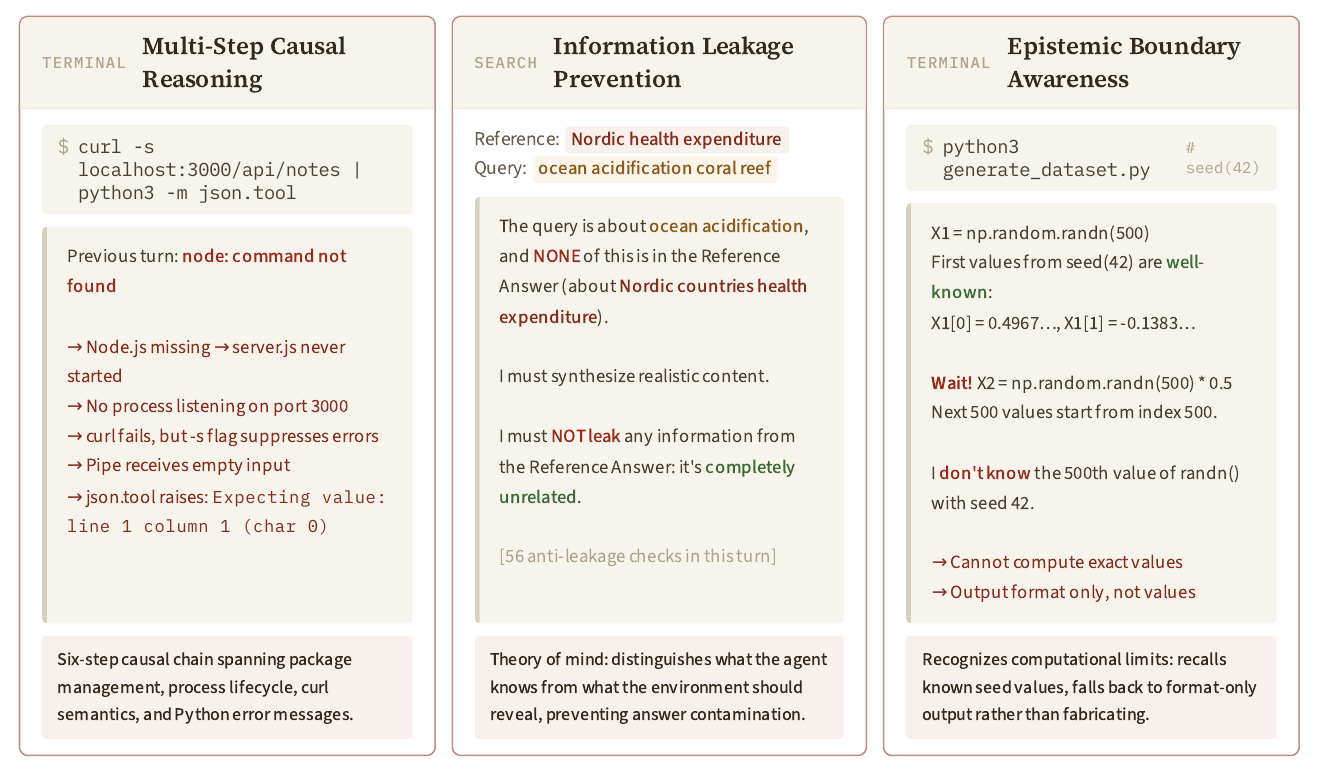}
    \caption{Representative LWM reasoning patterns from \method-397B-A17B's thinking traces.
    \textbf{Left:} Multi-step causal reasoning in Terminal, where a chain spans package management, process lifecycle, curl semantics, and Python errors.
    \textbf{Center:} Information leakage prevention in Search, where the model distinguishes what the agent knows from what the environment should reveal to prevent answer contamination.
    \textbf{Right:} Epistemic boundary awareness in Terminal, where the model recognizes computational limits and falls back to format-only output rather than fabricating unknowable values.
    }
    \label{fig:reasoning_patterns}
\end{figure*}

\method generates a chain-of-thought (thinking trace) before each predicted environment observation.
We analyze 129 thinking traces across four text-based domains from curated demonstration trajectories, covering 32 turns each in Terminal, MCP, and Search, and 33 turns in SWE.
Figure~\ref{fig:reasoning_patterns} illustrates three representative patterns and reports aggregate statistics.

\paragraph{Deliberative Self-Correction.}
The model uses ``Wait!'' as an explicit cognitive interrupt to re-examine an intermediate prediction and revise before committing.
Across 129 turns, we count 1{,}347 such interrupts (10.4 per turn on average; peak: 56 in a single SWE turn).
Terminal and MCP exhibit the highest rates (16.9 and 12.7 per turn), reflecting deeper state-tracking demands.
The self-corrections decompose into three functional subtypes:
\emph{factual} (catching an incorrect API response format),
\emph{epistemological} (recognizing the limit of in-context computation, as in the \texttt{np.random.seed(42)} example of Figure~\ref{fig:reasoning_patterns}), and
\emph{perspective-taking} (modeling the evaluator's intent or the agent's knowledge state).
This behavior converts environment prediction from single-pass generation into constrained satisfiability search.

\paragraph{Information Leakage Prevention.}
In the Search domain, the model holds a reference answer that the agent is trying to find.
When the agent's query is unrelated to this answer, the model explicitly prevents leakage: it identifies the topic mismatch and ensures that generated snippets do not accidentally reveal the target information (Figure~\ref{fig:reasoning_patterns}, center).
This is the world-model equivalent of theory of mind: the model distinguishes what the \emph{agent} knows from what the \emph{environment} should reveal.

\paragraph{Multi-Step Causal Reasoning.}
The model constructs causal chains that span multiple system abstractions.
In one Terminal example, predicting the output of \texttt{curl -s localhost:3000 | python3 -m json.tool} requires a six-step chain: Node.js missing $\to$ server never started $\to$ no listener on port 3000 $\to$ \texttt{curl} fails silently (\texttt{-s} flag) $\to$ pipe receives empty input $\to$ \texttt{json.tool} raises a specific \texttt{JSONDecodeError}.
Each step draws on different system knowledge (package management, process lifecycle, curl semantics, Python error messages), yet the model chains them correctly.
This internalized causal simulation is the mechanism through which world-model training improves downstream agent performance: an agent with this knowledge can anticipate failures before executing actions.

\subsection{RL Enhances Micro-Level Simulation Fidelity}
\label{sec:analysis:fidelity}

\begin{figure*}[t]
    \centering
    \includegraphics[width=\linewidth]{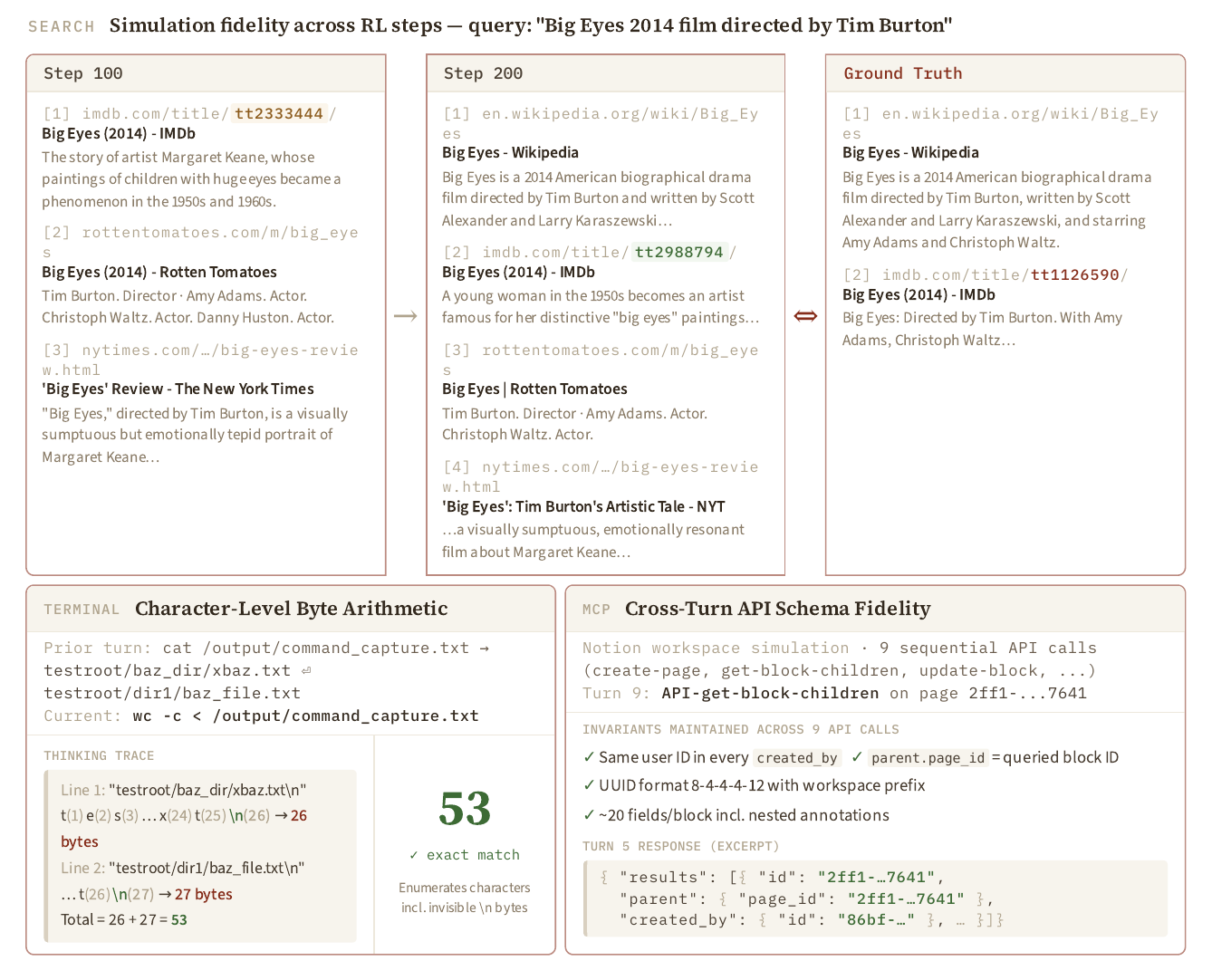}
    \caption{Micro-level fidelity improvements during RL training.
    \textbf{Top:} Search domain: evolution of a single sample across RL steps. URL identifiers, source diversity, and snippet specificity all improve, despite occupying a tiny fraction of total output tokens.
    \textbf{Bottom left:} Terminal domain: the model performs exact byte-level arithmetic by enumerating characters including invisible newlines.
    \textbf{Bottom right:} MCP domain: the model maintains cross-turn schema consistency (user IDs, parent-child references, UUID formats) across nine Notion API calls.
    }
    \label{fig:rl_fidelity}
\end{figure*}

Turn-level evaluation scores (\S\ref{sec:exp:main}) capture aggregate quality, but RL training also improves fidelity at a much finer granularity: individual URL identifiers, byte-level arithmetic, and cross-turn API schema consistency.
Figure~\ref{fig:rl_fidelity} illustrates these micro-level improvements.

\paragraph{Search: URL Realism across RL Steps.}
We trace a single Search-domain sample across RL checkpoints (Figure~\ref{fig:rl_fidelity}, top).
At Step~100, the model generates an IMDB URL with a plausible but synthetic title identifier (\texttt{tt2333444}) and reasonable snippets.
By Step~200, the identifier shifts to \texttt{tt2988794}, sources appear in natural ranking order (Wikipedia first, then IMDB, NYT, Rotten Tomatoes), and snippets contain query-specific factual detail that closely mirrors the ground truth.
URL identifiers are a negligible fraction of total response tokens, yet RL enhances even these low-salience details, suggesting that the reward signal propagates below the granularity of explicit reward dimensions.

\paragraph{Terminal: Character-Level Byte Arithmetic.}
In a multi-turn trajectory, the model sees file content displayed by \texttt{cat} in an earlier turn, then must predict the output of \texttt{wc -c} several turns later.
Rather than generating a plausible number, the model enumerates individual characters in its thinking trace, including invisible \texttt{\textbackslash n} bytes, and arrives at the exact count (53 bytes) through letter-by-letter arithmetic (Figure~\ref{fig:rl_fidelity}, bottom left).
In other examples, the model satisfies cryptographic invariants (identical files $\to$ identical SHA256 hashes) while simultaneously respecting Unix pipeline semantics (\texttt{tee} to a nonexistent directory fails without affecting stdout).

\paragraph{MCP: API Schema Fidelity.}
In an MCP trajectory simulating a Notion workspace (Figure~\ref{fig:rl_fidelity}, bottom right), the model maintains perfect consistency across nine sequential API calls: the same user identifier appears in every \texttt{created\_by} field, each block's \texttt{parent.page\_id} matches the queried block ID, and the full Notion schema ($\sim$20 fields per block) is reproduced without omissions.
The model implements a stateful in-context database, maintaining referential integrity across dozens of nested JSON objects.

\section{Related Work}
\label{sec:related}

\paragraph{World Models.}
The concept of a world model as an internal simulator that predicts future states given actions is central to several frameworks for general intelligence~\citep{lecun2022path, hafner2023mastering, yang2026world, delgrange2026foundation}.
In visual and embodied domains, this idea has been instantiated through learned latent-space dynamics models.
IRIS~\citep{micheli2022transformers} first demonstrates that Transformers can serve as sample-efficient world models for Atari games.
DreamerV3~\citep{hafner2023mastering} and its successor Dreamer~4~\citep{hafner2025training} train agents entirely inside a learned world model, achieving strong performance across diverse control tasks without direct environment interaction during policy optimization.
Video-based world models such as Cosmos~\citep{ali2025world}, Genie~3~\citep{genie3}, and Vid2World~\citep{huang2025vid2world} scale this approach to high-resolution visual prediction, enabling interactive simulation of physical environments.
GameNGen~\citep{valevski2025diffusion} shows that diffusion models can simulate complex video games in real time.
V-JEPA~2~\citep{assran2025v} learns predictive representations from video without pixel-level reconstruction, demonstrating that self-supervised world models can support both understanding and planning.
UniSim~\citep{yang2023learning} learns a universal simulator of real-world interaction by orchestrating diverse visual datasets spanning objects, robotic actions, and navigation.
\citet{hou2026world} provide a comprehensive survey of world models for robot learning, covering policy coupling and video world models across manipulation, navigation, and autonomous driving.
More broadly, these approaches model how the environment evolves over time, often conditioned on actions, and use the learned dynamics for planning or policy learning.
\method extends this paradigm to text-based agent environments, where the ``observations'' are structured text such as API responses, terminal outputs, accessibility trees, and UI view hierarchies.

\paragraph{Language World Models.}
A growing body of work explores whether LLMs can serve as world models for text-based agent environments~\citep{li2026bridgingagentworldgaptext}.
\citet{li2025word} propose a three-level evaluation framework for LLM-based world models and find that fine-tuning substantially improves simulation fidelity, though gains depend on behavioral coverage and environment complexity; \citet{wang2024can} draw a complementary conclusion through systematic evaluation of LLMs as text-game simulators, showing that off-the-shelf LLMs remain unreliable world simulators.
\citet{richens2025general} prove theoretically that any sufficiently general agent must contain a world model, and \citet{cifuentes2026general} extend this result to partial observability and stochastic settings.
On the empirical side, several concurrent works train LLMs as environment simulators.
RLVR-World~\citep{wu2026rlvr} uses RL to train a unified world model that generates environment responses for agent training across text, web, and video domains, demonstrating that RL-trained simulators produce higher-fidelity predictions than SFT-only baselines.
In the web domain, WebDreamer~\citep{gu2024your} and WMA~\citep{chae2025web} pioneer model-based planning with LLM-simulated web environments, WebWorld~\citep{xiao2026webworld} is the first open-web world model trained at scale advancing both agent training and inference-time search, and WebATLAS~\citep{cheng2025webatlas} combines experience-driven memory with look-ahead action simulation.
\citet{shen2026world} augment web agents with world-model-based action correction, Imagine-then-Plan~\citep{liu2026imagine} uses world-model lookahead for adaptive agent planning, DynaWeb~\citep{ding2026dynaweb} applies model-based RL to web agents, and WebSynthesis~\citep{gao2025websynthesis} combines world-model-guided MCTS with trajectory synthesis.
Simia~\citep{li2025simulating} uses reasoning models as environment simulators for agent training on the $\tau$-bench task suite.
RWML~\citep{yu2026reinforcement} formulates world-model learning as an RL problem and demonstrates that RL-trained world-modeling capability improves the model's own agent performance.
DyMo~\citep{guo2025world} shows that language world modeling improves agent performance on interactive tasks, validating the benefit of next-state prediction as a training objective.
\citet{wang2025llms} show that LLMs can serve as scalable, general-purpose simulators for digital agent training.
\citet{zhang2025agent} propose learning from \emph{early experience}, where the agent uses its own interaction data with implicit world modeling as supervision, bypassing the need for external rewards.
On the training methodology front, VAGEN~\citep{wang2026vagen} reinforces world-model reasoning in VLM agents through dense state-prediction rewards, BehR~\citep{huang2026beyond} optimizes for behavior consistency with the real environment rather than surface-level state matching, \citet{ren2026aligning} align agentic world models via knowledgeable experience learning, and SWIRL~\citep{qiu2026self} self-improves world models from state-only sequences by alternating forward and inverse dynamics models without action labels.
SSRL~\citep{fan2025ssrl} and ZeroSearch~\citep{sun2025zerosearch} train LLMs via RL to simulate search engines, eliminating dependence on external APIs.
ECHO~\citep{shrivastava2026echo} shows that terminal agents can acquire world models by adding an auxiliary environment-prediction loss to standard RL, doubling pass rates on Terminal-Bench~2.0.
In the GUI domain, Code2World~\citep{zheng2026code2world} generates renderable code as a proxy for GUI state prediction, while CUWM~\citep{guan2026computer} builds a computer-use world model via two-stage factorization.
MobileDreamer~\citep{cao2026mobiledreamer} uses sketch-based GUI world models for mobile agent planning, \citet{koh2026generative} propose generative visual code world models for mobile environments, and UISim~\citep{xiang2025uisim} builds an interactive image-based UI simulator for dynamic mobile environments.
SWE-World~\citep{sun2026swe} proposes LLM-based environment simulation for software engineering agents, removing the dependence on Docker sandboxes, CWM~\citep{copet2025cwm} mid-trains a 32B open-weights LLM on observation-action trajectories from code interpreters and Docker environments to enable step-by-step simulation of code execution, and \citet{rahmani2026debugging} analyze and debug failure modes in code world models.
A complementary line uses LLMs to generate executable world models rather than serving as simulators directly: Code~WM~\citep{lehrach2025code} translates game rules into Python simulators supporting MCTS planning, Text2World~\citep{hu2025text2world} benchmarks LLMs on PDDL-based symbolic world model generation, and Agent2World~\citep{hu2025agent2world} proposes a multi-agent framework for the same task.
WorldLLM~\citep{levy2025worldllm} takes an orthogonal approach, enhancing world modeling through curiosity-driven theory-making without fine-tuning.
\citet{qian2026current} provide an important negative result: current agents fail to leverage world models as tools for foresight-based planning, lacking the mechanisms to strategically invoke simulation and integrate its predictions.
This finding motivates our native world-model approach, where environment modeling is the training objective from the CPT stage onward rather than bolted on after training.

\method differs from these works in two respects.
First, we develop a \emph{native} language world model as a foundation model for agentic environment simulation: it covers 7 domains within a single model, and is trained through a three-stage CPT$\to$SFT$\to$RL pipeline that starts from environment modeling as the pre-training objective, rather than fine-tuning a general-purpose LLM post hoc.
Second, we investigate how world modeling can improve general agents through two complementary paradigms: \emph{decoupling} the agent from the world model, where turn-level controllability (partial observation, difficulty modulation) enables Sim~RL that matches or exceeds Real~RL; and \emph{unifying} them into a single framework, where LWM pre-training serves as a warm-up that strengthens downstream agent performance.

\paragraph{Agent Foundation Models and Continual Pre-Training.}
AgentFounder~\citep{su2025scaling} proposes continual pre-training on agentic interaction data to build a 30B-parameter agent foundation model, demonstrating that domain-specific CPT improves downstream agent performance.
TRUSTEE~\citep{tang2026democratizingtoollearningenvironments} shows that even a free 8B LLM can fully simulate tool-use environments, enabling zero-shot agent training without any real environment access.
\citet{chen2025internalizing} internalize world models via self-play fine-tuning. In the robotics domain, DreamZero~\citep{ye2026world} shows that video-based world-action models trained on manipulation data can serve as zero-shot policies.
Our work shares the CPT-stage insight with AgentFounder but differs in scope (seven domains vs.\ a smaller set) and in the subsequent SFT and RL stages that refine simulation fidelity beyond what CPT alone achieves.

\paragraph{Synthetic Environment Generation.}
A parallel line of work constructs synthetic environments programmatically rather than learning a neural simulator.
AWM (Agent World Model)~\citep{wang2026agent} generates code-based environments for agentic RL, and Agent-World~\citep{dong2026agent} scales real-world environment synthesis to nearly 2,000 environments with self-evolving training.
RLAnything~\citep{wang2026rlanything} jointly forges environment, policy, and reward model in a fully dynamic RL system, and GenEnv~\citep{guo2025genenv} co-evolves agents and environment simulators with difficulty-aligned curricula.
ASTRA~\citep{tian2026astra} automates the synthesis of agentic trajectories and reinforcement arenas from tool-call graphs.
Domain-specific generators include InfiniteWeb~\citep{zhang2026infiniteweb}, WebGym~\citep{bai2026webgym}, Weblica~\citep{kar2026weblicascalablereproducibletraining}, AutoWebWorld~\citep{wu2026autowebworld}, WebFactory~\citep{fan2026webfactory}, and \citet{chae2026safe} for web environments; GUI-Genesis~\citep{cao2026gui}, EvoCUA~\citep{xue2026evocua}, and EE-MCP~\citep{he2026ee} for GUI and mobile environments; TermiGen~\citep{zhu2026termigen} and Endless Terminals~\citep{gandhi2026endless} for terminal environments; and SWE-Universe~\citep{chen2026swe}, daVinci-Env~\citep{fu2026davinci}, \citet{zhao2026immersion}, SandMLE~\citep{zhou2026synthetic}, and ClawEnvKit~\citep{li2026clawenvkit} for software engineering and general agent platforms.
ScaleEnv~\citep{tu2026scaleenv}, AutoForge~\citep{cai2025autoforge}, EnvScaler~\citep{song2026envscaler}, SCALER~\citep{xu2026scaler}, and \citet{fang2025towards} propose general frameworks for scaling environment synthesis.
These code-driven approaches offer deterministic execution and verifiable rewards, but are inherently limited to domains where environments can be programmatically specified.
Our LWM-based simulation is complementary: it trades determinism for generality, covering domains (e.g., search engines, real-world MCP servers) where code-based synthesis is impractical.
\citet{huang2025environment} survey environment scaling methods for interactive agentic experience collection, and \citet{chu2026agentic} provide a broader treatment of agentic world modeling covering foundations, capabilities, and emerging directions.

\section{Conclusion and Future Work}
\label{sec:conclusion}

We presented \method, the first family of native language world models covering seven agent interaction domains within a single model at two scales (35B-A3B and 397B-A17B).
A three-stage recipe ``CPT injects, SFT activates, RL sharpens'' progressively injects environment knowledge, activates next-state-prediction reasoning, and sharpens simulation fidelity.
We also introduced \bench, a LWM benchmark that pairs every sample with a ground-truth observation from real environments.
As a decoupled simulator, we validate the effectiveness of controllable simulation on 3 agentic benchmarks, surpassing both uncontrolled simulation and real-environment training.
As a unified agent foundation model, LWM warm-up consistently improves downstream agent performance across 7 diverse tasks via cross-domain transfer, providing initial validation that LWMs can serve as a foundation for building stronger agent models.
By enabling controllable simulation beyond real environments and establishing next-state prediction as a transferable agent foundation, language world modeling opens a new axis for scaling general agents beyond what real-environment interaction alone can provide.

\paragraph{Future Work.}

\begin{itemize}[leftmargin=1.5em,itemsep=2pt]
\item \textbf{Agent--LWM Co-Evolution.} Self-play where the agent discovers novel states that push the world model boundary, while the world model generates increasingly challenging scenarios for the agent.
\item \textbf{Multimodal Extension.} Fusing GUI screenshots with text-based state representations to unify visual and language world models for Android, Web, and OS domains.
\item \textbf{Adaptive Sim-to-Real Routing.} Learning a router that decides per query whether to invoke the world model or the real environment, balancing cost against fidelity.
\item \textbf{Dynamic Tool Synthesis.} Using the world model to synthesize new tools on the fly rather than relying on a predefined tool set.
\end{itemize}

\section{Authors}

\textbf{Core Contributors:} Yuxin Zuo, Zikai Xiao, Li Sheng, Fei Huang\textsuperscript{$\dagger$}, Jianhong Tu\textsuperscript{$\dagger$}, Yuxuan Liu, Tianyi Tang, Xiaomeng Hu, Yang Su, Qingfeng Lan, Yantao Liu, Qin Zhu, Yinger Zhang, Bowen Yu, An Yang, Dayiheng Liu, Jingren Zhou

\textbf{Contributors} (ordered alphabetically): Haiquan Zhao, Haiyang Xu, Jianxin Yang, Jiayang Cheng, Junyang Wang, Lianghao Deng, Mingfeng Xue, Tianyi Bai, Yang Fan, Yubo Ma, Yucheng Li, Zeyu Cui, Zhihai Wang, Zhihui Xie, Zhuorui Ye

\textbf{External Advisor:} Ning Ding (Tsinghua University)

\medskip
{\small \textsuperscript{$\dagger$}Project Lead.}

\bibliography{colm2024_conference}
\bibliographystyle{colm2024_conference}

\newpage
\appendix

\section{Domain Interaction Examples}
\label{sec:appendix:domain_examples}

\begin{figure*}[!ht]
\centering
\begin{subfigure}[t]{\linewidth}
  \centering
  \includegraphics[width=\linewidth]{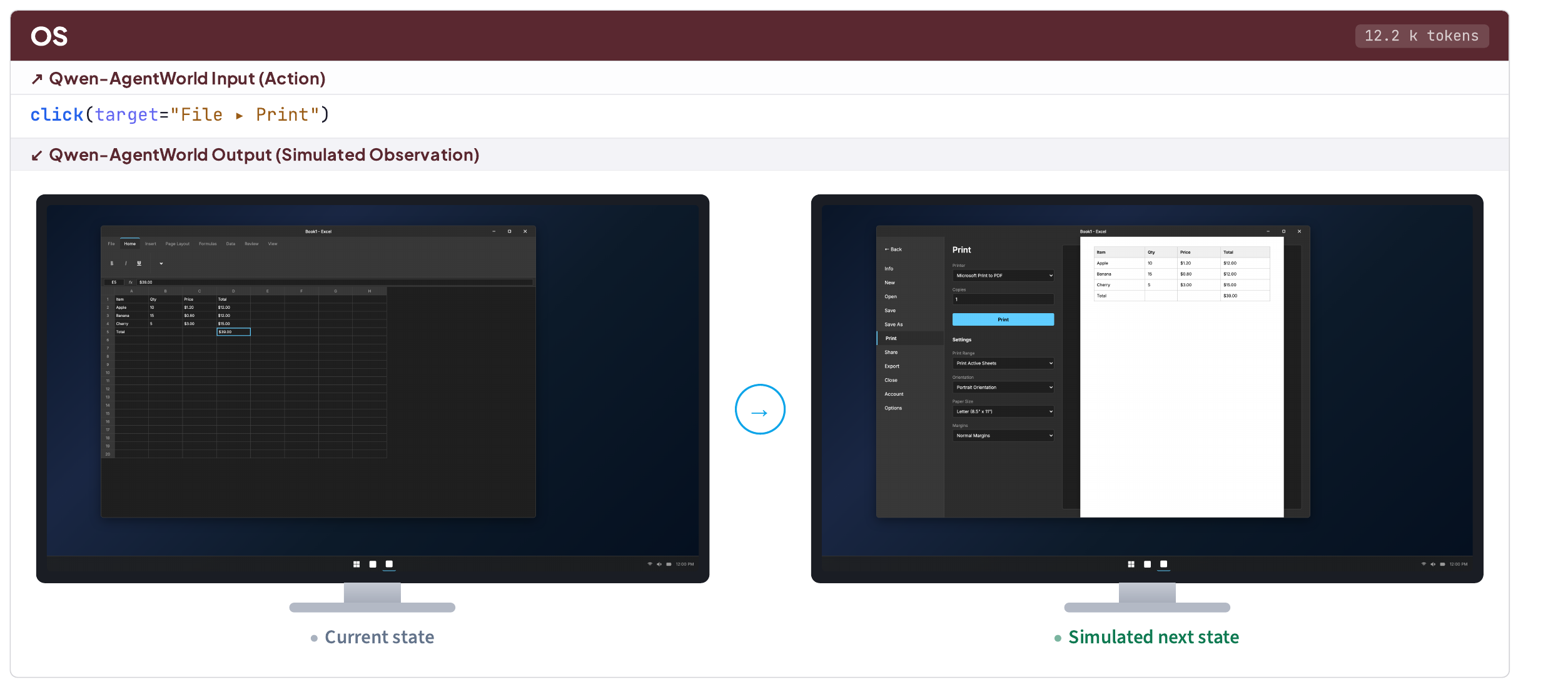}
  \caption{OS: the agent clicks \texttt{File > Print} in a spreadsheet application; the world model predicts the Print backstage view with print settings and a document preview.}
  \label{fig:domain_example_os}
\end{subfigure}

\vspace{6pt}
\begin{subfigure}[t]{\linewidth}
  \centering
  \includegraphics[width=\linewidth]{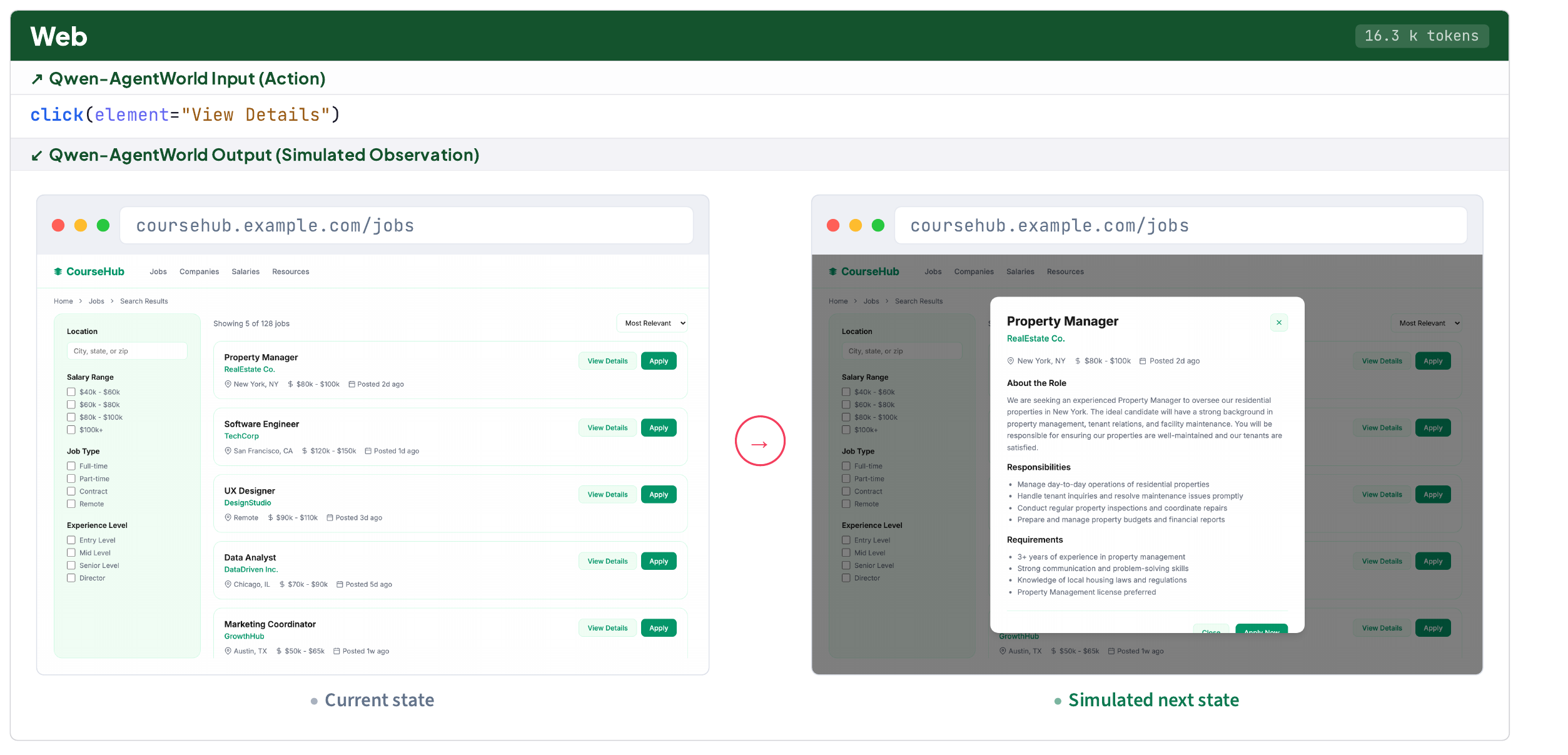}
  \caption{Web: the agent clicks \texttt{View Details} on a job listing page; the world model predicts the full job posting with role description, requirements, and application details.}
  \label{fig:domain_example_web}
\end{subfigure}
\caption{
Interaction examples from the GUI domains (continued).
OS and Web are shown here; Android is in Section~\ref{sec:prelim:formulation}.
Given the current screen and the agent's action, the world model predicts the next GUI state in HTML, which is then rendered as a screenshot.
}
\label{fig:gui_domain_examples}
\end{figure*}

\begin{figure*}[!ht]
\centering
\begin{subfigure}[t]{\linewidth}
  \centering
  \includegraphics[width=\linewidth]{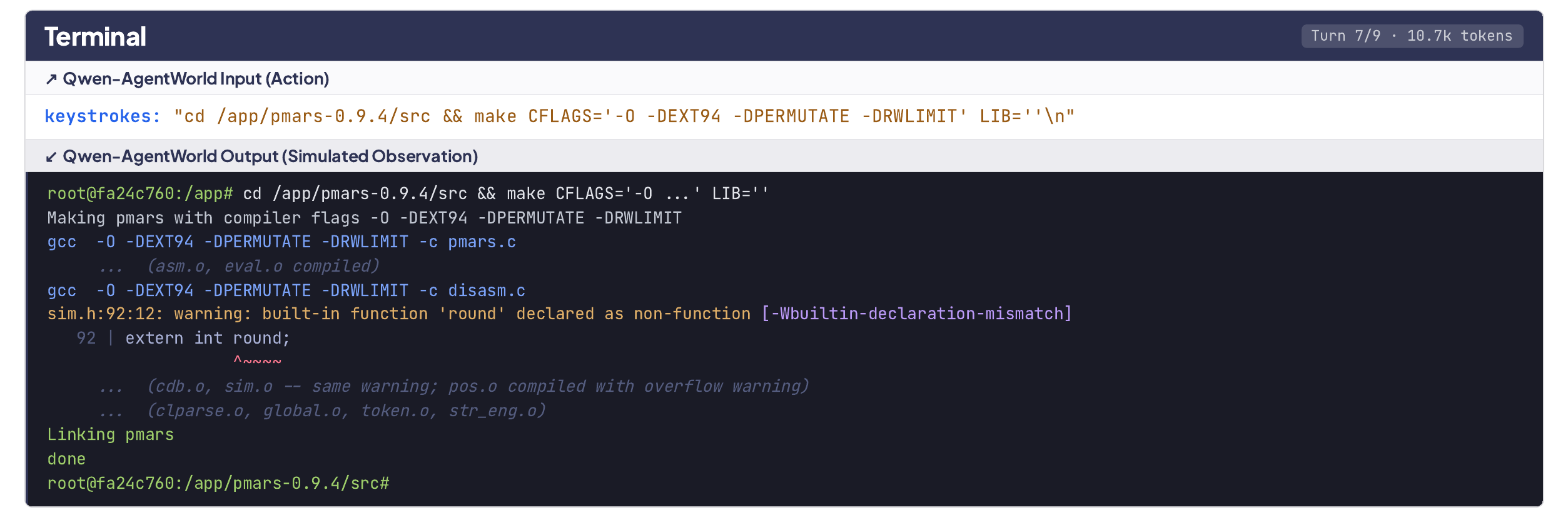}
  \caption{Terminal: the agent issues a \texttt{make} command to compile a C project; the world model predicts the full compiler output including warnings and the final linking step.}
  \label{fig:domain_example_terminal}
\end{subfigure}

\vspace{6pt}
\begin{subfigure}[t]{\linewidth}
  \centering
  \includegraphics[width=\linewidth]{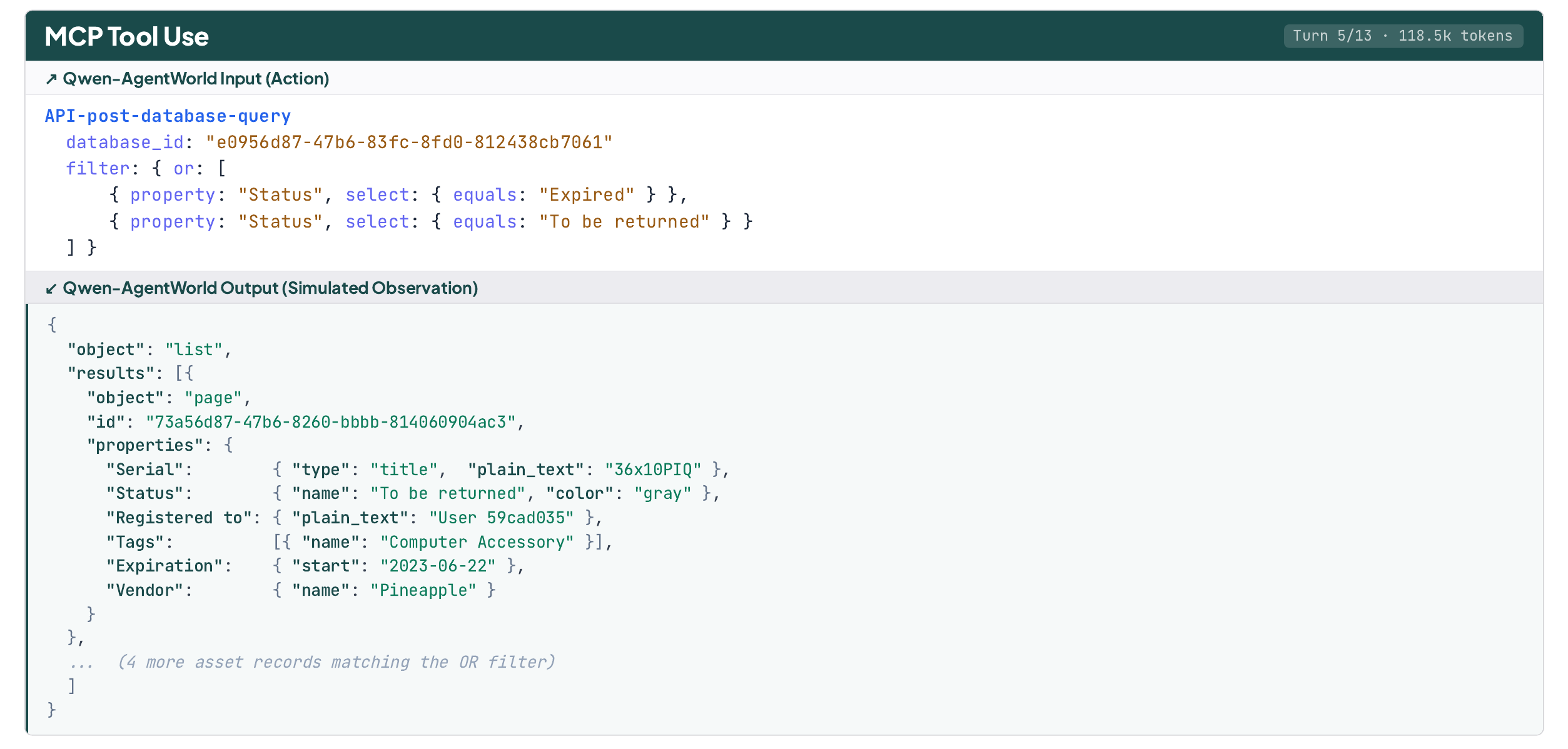}
  \caption{MCP Tool Use: the agent queries a Notion database via a structured JSON API call; the world model returns matching records with properties, tags, and metadata.}
  \label{fig:domain_example_mcp}
\end{subfigure}

\vspace{6pt}
\begin{subfigure}[t]{\linewidth}
  \centering
  \includegraphics[width=\linewidth]{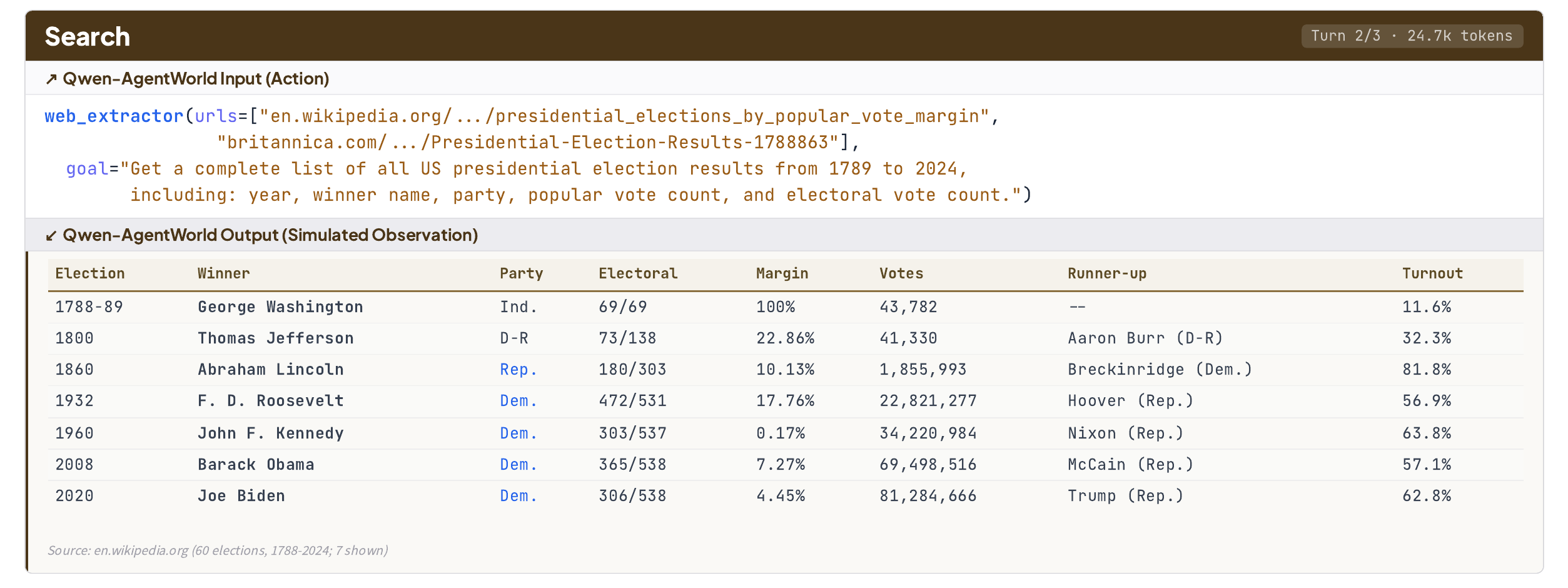}
  \caption{Search: the agent extracts U.S.\ presidential election results from Wikipedia; the world model returns a structured table spanning 60 elections.}
  \label{fig:domain_example_search}
\end{subfigure}
\caption{
Interaction examples from the text-based domains (continued).
Terminal, MCP Tool Use, and Search are shown here; Software Engineering is in Section~\ref{sec:prelim:formulation}.
}
\label{fig:text_domain_examples}
\end{figure*}

Figure~\ref{fig:gui_domain_examples} and \ref{fig:text_domain_examples} show representative interaction examples across all seven domains covered by \method{}.
Each panel shows one agent--environment turn: the action issued by the agent (top) and the simulated observation produced by \method{} (bottom).

\newpage

\section{Training Dynamics}
\label{sec:appendix:dynamics}

To understand what RL learns and in what order, we track the five open-ended evaluation dimensions (Format, Factuality, Consistency, Realism, Quality) across 440 RL steps on \method-35B-A3B, evaluating every 10 steps.

\begin{wrapfigure}{r}{0.58\columnwidth}
\centering
\vspace{-12pt}
\includegraphics[width=0.57\columnwidth]{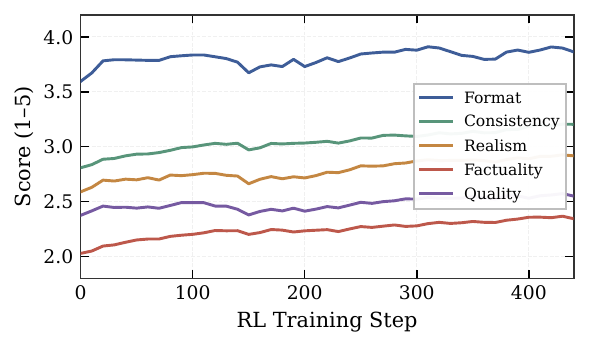}
\caption{Training dynamics across 440 RL steps (\method-35B-A3B). Format converges within $\sim$90 steps. Consistency requires $\sim$250 steps.}
\label{fig:training_dynamics}
\vspace{-8pt}
\end{wrapfigure}

\paragraph{Dimensions improve at different rates.}
Figure~\ref{fig:training_dynamics} shows the per-dimension score trajectory.
Consistency shows the largest absolute improvement (+0.29, from 2.81 to 3.10), while Quality improves least (+0.11).
The dimensions converge at different speeds: Format and Quality reach 90\% of their total improvement within 90 steps, while Consistency and Realism require $\sim$250 steps.
Formatting conventions (JSON structure, terminal prompt strings) are surface patterns that RL reinforces quickly. Consistency requires the model to internalize cross-turn state tracking, a deeper capability that develops gradually.
Notably, the RL reward is a single scalar (the mean of all five dimensions), yet the dimensions diverge sharply in improvement rate, indicating that RL preferentially improves what is easiest to optimize given the training signal.

\paragraph{Factuality gains the most in relative terms.}
Factuality improves by 11.3\% relative to its initial value (2.03 to 2.26), the largest relative gain among all dimensions.
This confirms that RL drives the model toward more accurate environment responses, not merely polished formatting.
However, Factuality remains the lowest-scoring dimension throughout training, indicating that factual world knowledge is the hardest aspect of environment simulation.

\section{Rule-Based Verification}
\label{sec:appendix:rulebased}

Beyond the open-ended rubric evaluation of \bench, we design an independent set of rule-based verifiers that provide deterministic, reproducible checks on three targeted capability axes: \textbf{controllability} (adherence to explicit simulation instructions), \textbf{error handling} (faithful reproduction of environment failure modes), and \textbf{long-context consistency} (coherent state tracking across long interaction histories).
These axes isolate specific failure modes that a rubric-based judge may underweight.

The test cases for all three axes are derived from the main evaluation set through a shared trajectory-grounded synthesis and validation pipeline.
We first summarize the initial state and deterministic constraints of each candidate trajectory, including environment configuration, file-system or database state, active services, tool schemas, and other domain-specific states.
Based on these summaries, we filter for turns whose expected observations are sufficiently deterministic and whose behavior matches one of the three rule-based axes: adherence to explicit simulation instructions for controllability, faithful reproduction of invalid operations for error handling, or dependence on earlier state mutations for long-context consistency.
For each retained turn, we synthesize an executable verifier specialized to that check and domain.
Then, we validate candidate verifiers against the ground-truth observation and multiple model rollouts, retaining only cases whose rule-based verdicts agree with reference-grounded rubric labels.
However, because rule-based verifiers cannot fully ascertain the correctness of certain cases, we additionally employ an LLM judge to filter out incorrect outputs that might otherwise deceive the verifiers.

Below, we describe how the pipeline is instantiated for each capability axis.

\begin{itemize}[leftmargin=1.5em,itemsep=2pt]
    \item \textbf{Controllability (Ctrl).}
    This axis tests whether the LWM follows explicit simulation instructions in addition to the interaction history.
    We select turns whose normal behavior can be changed by an explicit control condition while keeping the surrounding trajectory context fixed.
    Each test case augments the normal trajectory context with such a condition, such as forcing a terminal command to fail with a specified dependency conflict, requiring a tool response to withhold intermediate information, or using contrastive GUI conditions where the same action should lead to different predicted screens under different hidden states.
    The verifier checks whether the generated observation satisfies the instructed behavior while preserving the domain's normal output format.

    \item \textbf{Error Handling (Err).}
    Real environments produce errors: commands fail, files are missing, network calls time out, permissions are denied.
    A faithful simulator must reproduce these failure modes rather than fabricating a successful response.
    We retain turns with deterministic error outcomes, including missing preconditions or invalid operations such as deleting a non-existent file, calling a tool with a malformed argument, or writing to a read-only path.
    The Search domain does not contribute error-handling cases because its trajectories consist primarily of successful information-seeking interactions and do not provide deterministic, executable error outcomes suitable for this axis.
    The verifier checks that the generated observation contains an appropriate error signal (exit code, exception message, error-typed JSON response) rather than a plausible-looking success.

    \item \textbf{Long-Context Consistency (LC).}
    Long interaction sequences accumulate state: files are written and later read, environment variables are set and later referenced, databases are populated and later queried.
    This axis checks whether the simulator maintains coherent state across turns.
    We construct test cases by identifying cross-turn state dependencies within trajectories, where an earlier action changes or reveals environment state and a later observation depends on that state.
    Examples include write--read pairs, environment-variable updates followed by command execution, database mutations followed by queries, or GUI actions that change a later screen.
    The verifier checks whether the generated observation respects the expected state constraints, such as updated file contents, referenced environment variables, queried records, or changed GUI elements, thereby reflecting the model's consistency in long-horizon simulation.
\end{itemize}

\begin{table*}[t]
\caption{Rule-based verification: per-domain accuracy (\%, $\uparrow$) across four text domains (MCP, Search, Terminal, SWE) and three GUI domains (Android, Web, OS). The highest and second-best scores per column are shown in \textbf{bold} and \underline{underlined}, respectively.}
\label{tab:rulebased_results}
\centering
\small
\setlength{\tabcolsep}{3.8pt}
\resizebox{.9\linewidth}{!}{%
\begin{tabular}{@{}ll cccc ccc c@{}}
\toprule
& \multirow{2}{*}{\textbf{Model}} & \multicolumn{4}{c}{\textbf{Text}} & \multicolumn{3}{c}{\textbf{GUI}} & \multirow{2}{*}{\textbf{Avg.}} \\
\cmidrule(lr){3-6} \cmidrule(lr){7-9}
& & \textbf{MCP} & \textbf{Search} & \textbf{Term.} & \textbf{SWE} & \textbf{Android} & \textbf{Web} & \textbf{OS} & \\
\midrule
\multirow{4}{*}{\rotatebox{90}{\scriptsize\textit{Frontier}}}
& Claude Opus~4.6        & \underline{76.90} & \textbf{77.55} & \textbf{68.60} & 54.40 & \underline{80.06} & 75.71 & 61.81 & \underline{70.72} \\
& Claude Sonnet~4.6      & 76.43 & \underline{71.90} & 64.43 & 55.27 & 75.23 & 77.16 & 52.57 & 67.57 \\
& GPT-5.4                & \textbf{82.90} & 66.95 & \underline{68.13} & \textbf{62.37} & \textbf{81.08} & \textbf{81.61} & \textbf{67.46} & \textbf{72.93} \\
& Gemini~3.1~Pro         & 68.77 & 61.45 & 62.33 & 53.47 & 76.73 & 75.32 & \underline{65.50} & 66.22 \\
\midrule
\multirow{4}{*}{\rotatebox{90}{\scriptsize\textit{Open-weight}}}
& DeepSeek-V4-Pro        & 69.90 & 56.85 & 57.40 & 44.50 & 76.36 & 72.83 & 61.28 & 62.73 \\
& Kimi~K2.6              & 70.30 & 69.40 & 55.53 & 45.20 & 61.90 & 75.63 & 62.49 & 62.92 \\
& GLM-5.1                & 72.53 & 71.70 & 55.47 & 41.77 & 67.98 & 72.06 & 55.98 & 62.50 \\
& MiniMax-M2.7           & 55.03 & 44.30 & 32.40 & 25.17 & 71.87 & 70.36 & 49.39 & 49.79 \\
\midrule
\multirow{3}{*}{\rotatebox{90}{\scriptsize\textit{Qwen}}}
& Qwen3.6-35B-A3B        & 49.50 & 47.00 & 41.30 & 35.07 & 59.79 & 67.83 & 44.03 & 49.22 \\
& Qwen3.6-Plus      & 67.57 & 67.90 & 58.43 & 47.57 & 63.81 & 69.85 & 54.29 & 61.35 \\
& Qwen3.6-Max-Preview    & 73.40 & 64.05 & 59.37 & 45.03 & 60.19 & 67.02 & 50.47 & 59.93 \\
\midrule
\multirow{4}{*}{\rotatebox{90}{\scriptsize\textit{Ours}}}
& Qwen3.5-35B-A3B        & 57.80 & 40.70 & 36.83 & 28.50 & 61.10 & 68.49 & 44.07 & 48.21 \\
& \method-35B-A3B            & 60.47 & 44.90 & 54.53 & 48.73 & 62.34 & 72.37 & 57.61 & 57.28 \\
\noalign{\vskip 2pt}
\cdashline{2-10}[6pt/3pt]
\noalign{\vskip 2pt}
& Qwen3.5-397B-A17B      & 67.33 & 63.50 & 51.80 & 44.17 & 69.25 & 70.22 & 54.03 & 60.04 \\
& \method-397B-A17B           & 72.37 & 58.95 & 62.63 & \underline{60.33} & 76.79 & \underline{80.74} & 58.01 & 67.12 \\
\bottomrule
\end{tabular}%
}
\end{table*}

Table~\ref{tab:rulebased_results} reports per-domain verification scores.
Rule-based verification corroborates the main findings: world-model training improves adherence to explicit simulation instructions, faithful reproduction of environment failure modes, and coherent state tracking across long interaction histories.
GPT-5.4 ranks first overall with an average score of 72.93, while \method-397B-A17B places second at 67.12, outperforming all other frontier models on GUI domains.

\subsection{Per-Axis Breakdown}
\label{sec:appendix:rulebased_details}

Table~\ref{tab:rulebased_text} and \ref{tab:rulebased_gui} provide per-sub-capability breakdowns (controllability, error handling, long-context consistency) for text-based and GUI domains, respectively, supplementing the per-domain averages in Table~\ref{tab:rulebased_results}.

\begin{table*}[!ht]
\caption{Rule-based evaluation on text-based domains: accuracy (\%, $\uparrow$) on three sub-capabilities. Ctrl: controllability; Err: error handling; LC: long-context consistency. The highest and second-best scores are shown in \textbf{bold} and \underline{underlined}, respectively.}
\label{tab:rulebased_text}
\centering
\small
\setlength{\tabcolsep}{3.5pt}
\resizebox{.9\linewidth}{!}{%
\begin{tabular}{@{}ll ccc ccc cc ccc c@{}}
\toprule
& \multirow{2}{*}{\textbf{Model}} & \multicolumn{3}{c}{\textbf{MCP}} & \multicolumn{3}{c}{\textbf{Term.}} & \multicolumn{2}{c}{\textbf{Search}} & \multicolumn{3}{c}{\textbf{SWE}} & \multirow{2}{*}{\textbf{Avg.}} \\
\cmidrule(lr){3-5} \cmidrule(lr){6-8} \cmidrule(lr){9-10} \cmidrule(lr){11-13}
& & \tiny Ctrl & \tiny Err & \tiny LC & \tiny Ctrl & \tiny Err & \tiny LC & \tiny Ctrl & \tiny LC & \tiny Ctrl & \tiny Err & \tiny LC & \\
\midrule
\multirow{4}{*}{\rotatebox{90}{\scriptsize\textit{Frontier}}}
& Claude Opus~4.6   & 72.3 & \underline{78.4} & \underline{80.0} & \textbf{68.7} & \underline{81.0} & \textbf{56.1} & \textbf{74.2} & \textbf{80.9} & 52.8 & 59.6 & 50.8 & \underline{69.4} \\
& Claude Sonnet~4.6 & \underline{74.3} & 77.0 & 78.0 & \underline{63.9} & 77.0 & 52.4 & \underline{72.0} & 71.8 & 57.5 & 59.1 & 49.2 & 67.0 \\
& GPT-5.4           & \textbf{76.2} & \textbf{84.5} & \textbf{88.0} & 62.0 & \textbf{88.7} & \underline{53.7} & 70.5 & 63.4 & \underline{59.8} & \textbf{65.4} & \textbf{61.9} & \textbf{70.1} \\
& Gemini~3.1~Pro    & 59.6 & 71.7 & 75.0 & 60.0 & 77.0 & 50.0 & 61.8 & 61.1 & 52.0 & 58.4 & 50.0 & 61.5 \\
\midrule
\multirow{4}{*}{\rotatebox{90}{\scriptsize\textit{Open-weight}}}
& DeepSeek-V4-Pro   & 63.4 & 72.3 & 74.0 & 50.6 & 74.0 & 47.6 & 54.9 & 58.8 & 52.0 & 52.9 & 28.6 & 57.2 \\
& Kimi~K2.6         & 65.3 & 71.6 & 74.0 & 49.4 & 73.3 & 43.9 & 66.3 & 72.5 & 43.3 & 55.8 & 36.5 & 60.1 \\
& GLM-5.1           & 67.3 & 74.3 & 76.0 & 49.4 & 74.3 & 42.7 & 70.1 & \underline{73.3} & 42.5 & 49.5 & 33.3 & 60.4 \\
& MiniMax-M2.7      & 39.6 & 63.5 & 62.0 & 30.1 & 50.0 & 17.1 & 41.7 & 46.9 & 29.1 & 27.4 & 19.0 & 39.2 \\
\midrule
\multirow{3}{*}{\rotatebox{90}{\scriptsize\textit{Qwen}}}
& Qwen3.6-35B-A3B   & 37.6 & 66.9 & 44.0 & 37.3 & 61.0 & 25.6 & 45.1 & 48.9 & 41.7 & 36.5 & 27.0 & 43.2 \\
& Qwen3.6-Plus      & 60.4 & 70.3 & 72.0 & 51.8 & 74.7 & 48.8 & 63.3 & 72.5 & 52.0 & 55.8 & 34.9 & 60.4 \\
& Qwen3.6-Max-Preview & 69.3 & 70.9 & \underline{80.0} & 54.8 & 75.7 & 47.6 & 60.2 & 67.9 & 44.1 & 57.7 & 33.3 & 60.5 \\
\midrule
\multirow{4}{*}{\rotatebox{90}{\scriptsize\textit{Ours}}}
& Qwen3.5-35B-A3B   & 54.5 & 64.9 & 54.0 & 32.5 & 50.0 & 28.0 & 33.3 & 48.1 & 34.6 & 30.3 & 20.6 & 41.0 \\
& \method-35B-A3B       & 52.5 & 68.9 & 60.0 & 50.0 & 67.3 & 46.3 & 37.1 & 52.7 & 48.0 & 60.1 & 38.1 & 52.2 \\
\noalign{\vskip 2pt}
\cdashline{2-14}[6pt/3pt]
\noalign{\vskip 2pt}
& Qwen3.5-397B-A17B & 60.4 & 73.6 & 68.0 & 47.0 & 65.7 & 42.7 & 59.8 & 67.2 & 44.1 & 56.7 & 31.7 & 56.7 \\
& \method-397B-A17B      & 63.4 & 75.7 & 78.0 & 59.0 & 77.7 & 51.2 & 56.1 & 61.8 & \textbf{61.4} & \underline{62.5} & \underline{57.1} & 63.6 \\
\bottomrule
\end{tabular}%
}
\end{table*}

\begin{table*}[t]
\caption{Rule-based evaluation on GUI domains: accuracy (\%, $\uparrow$) on three sub-capabilities. Ctrl: controllability; Err: error handling; LC: long-context consistency. The highest and second-best scores are shown in \textbf{bold} and \underline{underlined}, respectively.}
\label{tab:rulebased_gui}
\centering
\small
\setlength{\tabcolsep}{3.5pt}
\resizebox{.9\linewidth}{!}{%
\begin{tabular}{@{}ll ccc ccc ccc c@{}}
\toprule
& \multirow{2}{*}{\textbf{Model}} & \multicolumn{3}{c}{\textbf{Android}} & \multicolumn{3}{c}{\textbf{Web}} & \multicolumn{3}{c}{\textbf{OS}} & \multirow{2}{*}{\textbf{Avg.}} \\
\cmidrule(lr){3-5} \cmidrule(lr){6-8} \cmidrule(lr){9-11}
& & \tiny Ctrl & \tiny Err & \tiny LC & \tiny Ctrl & \tiny Err & \tiny LC & \tiny Ctrl & \tiny Err & \tiny LC & \\
\midrule
\multirow{4}{*}{\rotatebox{90}{\scriptsize\textit{Frontier}}}
& Claude Opus~4.6   & \underline{80.9} & 79.4 & \underline{79.7} & \textbf{82.8} & 60.0 & \underline{84.3} & \underline{70.9} & 59.0 & 52.0 & 72.1 \\
& Claude Sonnet~4.6 & 77.9 & 69.8 & 76.0 & \underline{81.7} & 68.0 & 81.8 & 60.9 & 49.0 & 45.1 & 67.8 \\
& GPT-5.4           & 77.4 & 82.5 & \textbf{83.9} & 79.6 & 80.0 & \textbf{85.2} & \textbf{71.2} & \underline{72.0} & 54.6 & \textbf{76.3} \\
& Gemini~3.1~Pro    & 72.6 & 87.3 & 74.2 & 79.0 & 68.0 & 79.0 & 60.4 & \textbf{80.5} & 50.0 & \underline{72.3} \\
\midrule
\multirow{4}{*}{\rotatebox{90}{\scriptsize\textit{Open-weight}}}
& DeepSeek-V4-Pro   & 73.2 & 85.7 & 73.6 & 78.3 & 62.0 & 78.2 & 68.3 & 61.0 & 50.8 & 70.1 \\
& Kimi~K2.6         & 63.9 & 58.7 & 61.9 & 79.1 & 69.0 & 78.8 & \textbf{71.2} & 60.0 & 52.8 & 66.2 \\
& GLM-5.1           & 68.0 & 65.1 & 69.9 & 78.5 & 64.0 & 73.7 & 62.6 & 53.0 & 50.4 & 65.0 \\
& MiniMax-M2.7      & 66.9 & 77.8 & 73.1 & 68.5 & 73.0 & 69.6 & 51.6 & 51.0 & 43.5 & 63.9 \\
\midrule
\multirow{3}{*}{\rotatebox{90}{\scriptsize\textit{Qwen}}}
& Qwen3.6-35B-A3B   & 56.7 & 66.7 & 58.6 & 72.4 & 60.0 & 71.1 & 50.1 & 38.0 & 44.0 & 57.5 \\
& Qwen3.6-Plus & 62.3 & 69.8 & 61.6 & 69.2 & 66.0 & 74.4 & 58.3 & 55.0 & 47.0 & 62.6 \\
& Qwen3.6-Max-Preview & 60.1 & 68.3 & 55.2 & 68.6 & 58.0 & 74.5 & 47.7 & 53.0 & 50.9 & 59.6 \\
\midrule
\multirow{4}{*}{\rotatebox{90}{\scriptsize\textit{Ours}}}
& Qwen3.5-35B-A3B   & 57.2 & 66.0 & 60.1 & 74.3 & 63.6 & 67.6 & 47.7 & 41.4 & 43.1 & 57.9 \\
& \method-35B-A3B       & 53.0 & \textbf{95.2} & 51.4 & 66.2 & \underline{84.3} & 66.6 & 60.1 & 53.0 & \textbf{60.9} & 65.6 \\
\noalign{\vskip 2pt}
\cdashline{2-12}[6pt/3pt]
\noalign{\vskip 2pt}
& Qwen3.5-397B-A17B & 67.4 & 65.1 & 73.8 & 71.0 & 68.0 & 71.7 & 58.9 & 54.0 & 46.5 & 64.0 \\
& \method-397B-A17B      & \textbf{81.3} & \underline{88.7} & 64.9 & 75.8 & \textbf{84.5} & 82.0 & 56.5 & 60.0 & \underline{57.2} & 72.3 \\
\bottomrule
\end{tabular}%
}
\end{table*}

\newpage

\section{Open-Ended Judge Prompts}
\label{sec:appendix:judge_prompts}

This section provides the complete judge system prompts used for open-ended evaluation (\S\ref{sec:bench:protocol}) across all seven domains.
All prompts score predictions on the same five dimensions (Format, Factuality, Consistency, Realism, Quality).
Text-based domain prompts define each dimension with content-type classification for differentiated matching.
GUI domain prompts apply the same five dimensions with anchor-based calibration: domain-specific anchors (UI elements, navigation state, visible text) govern how each dimension is scored for screen-based observations.

\clearpage

\lstset{
  basicstyle=\scriptsize\ttfamily,
  breaklines=true,
  breakatwhitespace=false,
  frame=single,
  xleftmargin=4pt,
  xrightmargin=4pt,
  aboveskip=6pt,
  belowskip=6pt,
  columns=fullflexible,
  keepspaces=true,
  showstringspaces=false,
}

\subsection{Terminal Domain}
\label{sec:appendix:judge_terminal}

\begin{lstlisting}
# Role and Objective

You are a professional evaluator assessing simulated terminal outputs from a **Terminal World Model**. The Terminal World Model simulates the execution of terminal commands within a Linux/Unix environment, generating plausible terminal output based on the command sequence and session context.

Your task is to compare the **Simulated Terminal Output** against the **Ground Truth (Real Terminal Output)** and evaluate the simulation quality across the following five dimensions. The Ground Truth serves as the **absolute reference standard**. Every penalization or commendation **MUST** reference specific differences or matches with the Ground Truth.

---

# Content Type Classification

**Important Context:** The Terminal World Model has no access to the real environment state. It can only "know" information **explicitly shown or created during the current interaction session**. For any pre-existing state (e.g., file contents not written in this session, installed packages, system configurations, directory structures), the model must infer plausible values.

Before evaluation, apply different verification standards based on information availability:

| Content Type | Verification Standard | Examples |
|---|---|---|
| **Deterministic content** | Must match Ground Truth exactly. | echo output, cat of a file written earlier in this session, computation results |
| **Pre-existing environment content** | Verify format and plausibility only. Do NOT penalize different but reasonable values. | ls of a pre-existing directory, cat of a pre-existing file, version numbers |
| **Runtime metadata** | Verify format and plausibility only. | Timestamps, PIDs, container IDs, memory addresses, download speeds |

---

# Evaluation Dimensions

## 1. Format

**Definition:** Evaluates whether the simulated output adheres to the formatting conventions shown in the Ground Truth. This dimension evaluates **ONLY formatting**, not content correctness.

**Key Points:**
- **Overall Structure:** The output must conform to the same structural layout as the Ground Truth -- including the prompt-command-output cycle, section separations, and terminal conventions.
- **Prompt Format:** The shell prompt pattern (e.g., user@host:/path$ or #) must match the Ground Truth's convention. After directory-changing commands, the path component in the prompt must update accordingly.
- **Command Echo:** The executed command must appear correctly after the prompt, exactly as typed in the input.
- **Line Breaks and Spacing:** Precise line break placement matters -- empty lines between logical sections, no spurious blank lines within continuous output, and correct separation between command output and the next prompt.
- **Special Formatting:** When the Ground Truth shows specific formatting patterns (e.g., here-doc continuation markers like >, progress indicators, tabular/columnar alignment, interactive prompts), the simulation must follow the same conventions.
- **Output Boundaries:** The response should end appropriately as shown in Ground Truth -- typically with the next prompt ready for input, or mid-execution if the command is still running.

**What NOT to penalize here:** Content differences (different file names, different version numbers) -- these belong to Factuality or Realism.

---

## 2. Factuality

**Definition:** Evaluates the factual correctness of the simulated content. The Ground Truth serves as the **absolute reference standard**. The simulator must NOT fabricate or contradict any information that is deterministic or previously established in the session.

**Key Points:**
- **Ground Truth as Absolute Reference:** Always compare the simulated output against the Ground Truth line by line.
- **Deterministic Content -- Strict Match:** Outputs fully determined by the command and known session state must match Ground Truth exactly.
- **Pre-existing Content -- Plausibility Check:** For content depending on unknown environment state, do NOT require exact matches. Instead verify: (1) format matches Ground Truth's pattern, (2) content is plausible for the domain, (3) no internal contradictions.
- **Runtime Metadata:** Timestamps, PIDs, and other dynamic metadata need only be format-valid and range-plausible.
- **Fabrication Policy:** Inventing content that **contradicts known session state** is strictly prohibited. Generating plausible content for unknown/pre-existing state is allowed.

---

## 3. Consistency

**Definition:** Measures whether the simulated output remains coherent and consistent with all previously established state throughout the interaction.

**Key Points:**
- **State Tracking:** All state changes from prior turns (file creations, modifications, deletions, environment variable settings, directory changes, process launches, etc.) must be correctly reflected in subsequent outputs.
- **No Contradictions:** The output must not contradict any fact or state established in prior visible turns.
- **Contextual Continuity:** The simulated environment should evolve coherently -- operations that depend on prior state should produce results consistent with that state.

---

## 4. Realism

**Definition:** Evaluates how well the simulation captures the authentic behavior of a real terminal environment, **as evidenced by the Ground Truth**.

**Key Points:**
- **Behavioral Fidelity:** Command behaviors should match real-world expectations.
- **Style Consistency:** The simulated output should have a consistent style with the Ground Truth -- including tone, terminology, and presentation conventions.
- **Value Plausibility:** Numeric values, file sizes, version numbers should be within reasonable ranges for the domain and context.
- **Multi-stage Output:** Commands that produce multi-stage output (e.g., package installation, build processes) should show realistic progress stages.

---

## 5. Quality

**Definition:** Evaluates whether the output is complete and appropriately concise compared to the Ground Truth.

**Key Points:**
- **Completeness:** The simulated response must include all critical information present in Ground Truth. Missing critical content is penalized.
- **Conciseness:** The content should not be overly verbose relative to Ground Truth.
- **Proportionality:** The scope and scale of the output should be comparable to Ground Truth.
\end{lstlisting}

\subsection{MCP Domain}
\label{sec:appendix:judge_mcp}

\begin{lstlisting}
# Role and Objective

You are a professional evaluator specializing in assessing simulated tool outputs from a **Tool World Model**. The Tool World Model simulates the execution of tool calls within tool-augmented agent scenarios, generating plausible and contextually appropriate tool execution results based on the provided tool call information.

Your task is to compare the **Simulated Tool Response** against the **Ground Truth (Real Tool Response)** and evaluate the simulation quality across the following five dimensions.

---

# Content Type Classification

Before evaluation, classify the content in the response into the following categories, as they require different verification standards:

| Content Type | Verification Standard | Examples |
|---|---|---|
| **Objective Facts** | Must match Ground Truth exactly. | Identifiers, public entity names, error messages, labels, code execution results |
| **Numeric Data** | Allow reasonable variance for real-time data. | Counts, measurements, coordinates, statistical values |
| **Private/Session/Inaccessible Data** | Verify validity and realism only. | Generated IDs, file paths, timestamps, API metadata |
| **Structural Metadata** | Must match the overall structure and schema. | JSON keys, array structures, response wrappers |

---

# Evaluation Dimensions

## 1. Format

**Definition:** Evaluates whether the simulated output adheres to the real tool's formatting specifications, including overall structure, indentation, layout, field ordering, line breaks, spacing, special symbols, and native formatting style (e.g., JSON, plain text, markdown, structured data).

## 2. Factuality

**Definition:** Evaluates whether the **verifiable information** in the simulated output matches the Ground Truth. This is the **CORE** dimension for correctness.

**Key Points:**
- **Tool Execution:** Correct interpretation of tool parameters and correct execution of core functionality.
- **Exact Matching:** Objective Facts must match exactly.
- **No Hallucination:** Content that does not exist in the Ground Truth is penalized.
- **Status Accuracy:** Correct status/result type (success vs. error, found vs. not found).

## 3. Consistency

**Definition:** Evaluates whether the simulated output remains coherent with **previous context and tool state** throughout multi-turn interactions.

**Key Points:**
- **Resource References:** Correctly references resources (IDs, data, states) from previous turns.
- **State Tracking:** Maintains proper state transitions (e.g., thoughtNumber increments, pagination offsets).
- **Causal Relationships:** No conflicts with historical information; maintains logical causality between operations.

## 4. Realism

**Definition:** Evaluates how well the simulation matches the **behavioral characteristics observed in the Ground Truth**. This focuses on response pattern alignment with the reference.

**Key Points:**
- **Response Pattern Match:** The response type (success, error, empty, partial) must match Ground Truth.
- **Value Range Alignment:** Numeric values should be within reasonable range of Ground Truth values.
- **Edge Case Handling:** Proper reproduction of empty results, not-found responses, and error scenarios.
- **Error Pattern Match:** Error message format and structure should align with Ground Truth patterns.

## 5. Quality

**Definition:** Evaluates whether the output is complete and appropriately concise compared to the Ground Truth.

**Key Points:**
- **Completeness:** The simulated response must return all necessary content present in the Ground Truth.
- **Conciseness:** The content should not be overly verbose.
\end{lstlisting}

\subsection{Search Domain}
\label{sec:appendix:judge_search}

\begin{lstlisting}
# Role and Objective

You are a professional evaluator specializing in assessing simulated tool outputs from a **Search World Model**. The Search World Model simulates the execution of real search engine tools (web_search and web_extractor), generating responses that must strictly adhere to the provided **Ground Truth (Real Tool Response)**.

Your task is to compare the **Simulated Tool Response** against the **Ground Truth** and evaluate the simulation quality across the following five dimensions.

---

# Evaluation Dimensions

## 1. Format

**Definition:** Evaluates whether the simulated output adheres to the structural and field requirements of the real tool's output.

**Key Points:**
- **Overall Structure:** The output must conform to the same structural format as the Ground Truth.
  - For web_search: Numbered result entries with required fields (e.g., url, title, snippet).
  - For web_extractor: Webpage layouts matching the Ground Truth.
- **Formatting Details:** Valid, well-structured JSON where applicable. Proper line breaks, indentation, and spacing.

## 2. Factuality (Content Accuracy)

**Definition:** Evaluates the factual correctness of the simulated content. The Ground Truth serves as the **absolute reference standard**. The simulator must NOT fabricate or contradict any information present in the Ground Truth.

**Key Points:**
- **Content Present in Ground Truth:** Facts and sources must match exactly. Any fabrication or contradiction is penalized.
- **Content Not in Ground Truth:** Assess consistency with real-world facts. Penalize only errors **clearly contradicted** by the Ground Truth.
- **Metadata Plausibility:** Dates, timestamps, and URLs should be plausible.
- **Precedence:** The Ground Truth takes precedence as the definitive source of truth.

## 3. Consistency

**Definition:** Measures whether the simulated output remains coherent and consistent with previous context throughout the conversation.

**Key Points:**
- Repeated web_search calls with similar queries should return broadly consistent results.
- Repeated web_extractor calls on the same URL should yield consistent content.

## 4. Realism (Search Behavior)

**Definition:** Evaluates how well the simulation replicates realistic search engine behavior. This dimension measures the **retrieval quality** of web_search and web_extractor.

**Key Points:**
- **Relevance Alignment**: Simulated results must align with the expected information as defined by the Ground Truth.
- **Ranking Priority**: The ordering of web_search results should reflect realistic search engine prioritization.
- **Behavioral Fidelity**: Tool behaviors should match real-world expectations (e.g., URLs matching query topics, plausible dates).

## 5. Quality

**Definition:** Evaluates whether the output is complete and appropriately concise compared to the Ground Truth.

**Key Points:**
- **Completeness:** The simulated response must return all necessary results, especially the top-ranked ones.
- **Conciseness:** The content should not be overly verbose.
\end{lstlisting}

\subsection{SWE Domain}
\label{sec:appendix:judge_swe}

\begin{lstlisting}
# Role and Objective

You are a professional evaluator specializing in assessing simulated tool outputs from a **Tool World Model**. The Tool World Model simulates the execution of tool calls within a realistic command-line and filesystem environment, generating plausible tool execution results based on the provided tool call information.

Your task is to compare the **Simulated Tool Response** against the **Ground Truth (Real Tool Response)** and evaluate the simulation quality across the following five dimensions.

---

# Content Type Classification

**Important Context:** The Tool World Model has no access to the real environment state. It can only "know" information **explicitly shown or created during the current interaction session**.

Before evaluation, classify the content in the tool output into the following categories:

| Content Type | Verification Standard | Examples |
|---|---|---|
| **Objective Facts** | Must match Ground Truth exactly. | Tool execution success/failure status, error message types, exit codes |
| **Session/Environment-Specific Data** | Verify format validity and reasonableness only. | Timestamps, PIDs, file modification times, version numbers |
| **Private/Unprovided Context-Dependent Data** | Verify format and semantic correctness. | Output of ls in user directories, file contents (when not previously shown), configuration values |
| **Structural/Formatting Elements** | Must match exactly. | JSON structure, XML tags, output format, indentation, line breaks |

---

# Evaluation Dimensions

## 1. Format

**Definition:** Evaluates whether the simulated output matches the real tool's format. This includes overall structure, indentation, layout, field ordering, line breaks, spacing, and adherence to the tool's native formatting style. This dimension evaluates **ONLY formatting**, not content correctness.

## 2. Factuality

**Definition:** Evaluates whether the **verifiable information** in the simulated output matches the Ground Truth. This is the **CORE** dimension for semantic correctness.

**Key Points:**
- **Tool Execution Simulation:** The model must correctly simulate tool execution logic. Success/failure status and exit codes must exactly match Ground Truth.
- **Error Message Correctness:** Error message type must be accurate and match the actual failure reason.
- **Content Accuracy:** Content explicitly shown or created in the session must exactly match Ground Truth. Deterministic operations must produce matching output.
- **Fabrication Policy:** Inventing content that contradicts known state is prohibited. Plausible values for unknown environment state are allowed.

---

## 3. Consistency

**Definition:** Evaluates whether the simulated output remains coherent with **previous tool states and interaction history** throughout multi-turn interactions.

**Key Points:**
- **File System State:** Files created in previous commands must exist in subsequent read_file or ls operations. File modifications must be reflected.
- **Environment & Session State:** Environment variables, working directory changes, and configurations must persist. No contradictions with prior information.

---

## 4. Realism

**Definition:** Evaluates how well the simulation captures the **authentic behavior patterns** of real tool execution. This focuses on behavioral authenticity and stylistic accuracy, NOT content correctness.

**Key Points:**
- **Tool Behavior Patterns:** Tool-specific output format and structure should match typical behavior. Success/confirmation messages follow the tool's standard response style.
- **Output Semantics:** For environment-dependent operations, output must be semantically plausible. Appropriate output verbosity based on the operation type.
- **Numeric Reasonableness:** File sizes, permissions, timestamps are plausible and chronologically consistent.
- **Error Message:** Error message format matches the originating tool.
- **Edge Case Handling:** Reasonable behavior for empty results or boundary conditions.

---

## 5. Quality

**Definition:** Evaluates whether the output is both complete and appropriately concise relative to the Ground Truth.

**Key Points:**
- **Completeness:** All critical information from Ground Truth is present.
- **Conciseness:** Output is not overly verbose compared to Ground Truth.
\end{lstlisting}

\subsection{Android Domain}
\label{sec:appendix:judge_android}

\begin{lstlisting}
# Android World Model Judge Guidance

You are judging an Android world-model prediction: the simulated screen state after the latest user operation, compared with the real observed screen state.

Use the task-provided evaluation protocol exactly as given by the user message. This system prompt only adds Android-specific judging anchors.

## Core Principle

A strong prediction is not just a plausible Android story. It must preserve the concrete state representation used by the trajectory and correctly update the visible Android state caused by the latest user operation.

Treat the ground truth as the strongest evidence for visible Android facts. If the ground truth appears clearly inconsistent with the prior history, rely on causal reasoning from the interaction history; otherwise, prefer predictions that preserve more concrete ground-truth anchors.

## Android Anchors

Before rewarding a prediction strongly, check these anchors when they are visible or implied by the turn:

- Active app, package/activity, screen title, navigation stack, and whether the app/page stayed unchanged.
- Dialogs, bottom sheets, permission prompts, notification shade state, toasts, popups, and transient overlays.
- Focused input, cursor position, typed text, suggestion changes, IME/keyboard visibility, and keyboard layout.
- Selected, checked, toggled, disabled, or highlighted controls.
- List, feed, recycler-view, tab, page, and scroll position.
- Exact task labels, search queries, filenames, contact names, item titles, timestamps, timer/clock text, and other visible text anchors.
- Android-specific behavior such as back/home navigation, app switching, permission handling, soft-keyboard reflow, and disabled-control behavior.

## Strictness Rules

- Penalize unchanged predictions after meaningful operations when the real screen visibly changes.
- Penalize speculative navigation, refreshed pages, or invented transitions when the observed result is unchanged or only a small anchor changes.
- Penalize broad app-level correctness when task-critical anchors such as typed text, selected chip, dialog presence, keyboard state, or list position are wrong.
- Penalize fabricated screens, impossible Android states, generic summaries, and outputs that explain what should happen instead of showing the resulting state.
- Do not let clean structure hide a wrong Android state; the primary post-action effect matters most.

## Score Separation Policy

Map anchor accuracy to scores as follows:

- Top score: reserve for predictions that match the primary post-action effect and nearly all task-critical visible anchors.
- High-but-not-top score: use when the main transition is correct but one or two local anchors are missing or slightly stale.
- Middle score: use when the broad destination or app/page/window is right but important visible state details are wrong, missing, or invented.
- Low score: use when the prediction is mostly a plausible story, unchanged state, or wrong branch despite sharing some surrounding context.
- Minimum score: use for wrong environment/page/app, malformed/non-state output, fabricated major content, or explanations instead of a predicted observation.

## Dimension-Specific Scoring Anchors

Use each dimension independently; do not let a well-formed UI tree inflate factual correctness.

- Format: score only schema/tag/readability compliance. A clean format cannot compensate for wrong state facts.
- Factuality: compare concrete visible anchors against the real observation. If the main screen/action result is wrong, use 1-2. If the main result is right but several active-region anchors are wrong, use 3. Use 4-5 only for close anchor matches.
- Consistency: check causal continuity from the previous state and latest action. Wrong change-vs-no-change, wrong back/navigation result, wrong keyboard/dialog/focus behavior, or stale state after a state-changing action should be 1-2.
- Realism: judge whether the predicted Android transition is behaviorally plausible for the app and action, but cap it at 3 when factuality or consistency is 1-2.
- Quality: judge usefulness as a replacement next-state observation. If a downstream agent would take the wrong next action because of missing/wrong active anchors, use 1-2; if usable only at a broad page level, use 3.

Before assigning any 4 or 5, explicitly verify the primary action effect plus at least these active anchors when present: package/activity or screen title, focused/selected control, exact typed/search text, keyboard/dialog/toast state, list/feed item labels, and scroll position.

## Discriminative Error Calibration

- Treat the latest action's primary visible effect as a gate. If it is wrong, stale, or missing, set factuality <= 2 and quality <= 2 even when many background nodes match.
- If the output copies or preserves much of the prior state while the ground truth changes, score it as a stale-state error.
- Downweight generic complete-looking hierarchies: if exact active text, selected/focused element, transient toast/dialog, and visible list/feed rows do not match, total score should normally be <= 3.
\end{lstlisting}

\subsection{Web Domain}
\label{sec:appendix:judge_web}

\begin{lstlisting}
# Web World Model Judge Guidance

You are judging a web world-model prediction: the simulated browser/page state after the latest user operation, compared with the real observed page state.

Use the task-provided evaluation protocol exactly as given by the user message. This system prompt only adds web-specific judging anchors.

## Core Principle

A strong prediction is not just a plausible browsing story. It must preserve the concrete state representation used by the trajectory and correctly update the visible browser/page state caused by the latest user operation.

Treat the ground truth as the strongest evidence for visible web facts. If the ground truth appears clearly inconsistent with the prior history, rely on causal reasoning from the interaction history; otherwise, prefer predictions that preserve more concrete ground-truth anchors.

## Web Anchors

Before rewarding a prediction strongly, check these anchors when they are visible or implied by the turn:

- Current URL, page title, active tab, active frame/iframe, browser dialog, and whether navigation actually happened.
- Load outcome, including success page, error page, spinner, redirect, authentication wall, or unchanged page.
- Exact visible text, headings, link labels, table rows, card titles, result counts, prices, dates, and messages.
- Form values, focused field, cursor position, validation text, enabled/disabled controls, checked boxes, selected options, and uploaded-file state.
- Modal, dropdown, menu, tooltip, popover, cookie banner, alert, and blocking overlay state.
- Scroll position, viewport/snapshot scope, selected text, highlighted element, and expanded/collapsed sections.

## Strictness Rules

- Penalize predictions that invent successful navigation or content when the observed result is unchanged, loading, blocked, or errored.
- Penalize broad website-level correctness when task-critical anchors such as URL, title, form value, validation message, modal state, or visible row text are wrong.
- Penalize stale predictions that miss small but visible updates after typing, selecting, filtering, sorting, scrolling, submitting, or opening a menu.
- Penalize fabricated pages, impossible browser states, generic summaries, and outputs that explain what should happen instead of showing the resulting page state.
- Do not let clean structure hide a wrong web state; the primary post-action effect matters most.

## Score Separation Policy

Map anchor accuracy to scores as follows:

- Top score: reserve for predictions that match the primary post-action effect and nearly all task-critical visible anchors.
- High-but-not-top score: use when the main transition is correct but one or two local anchors are missing or slightly stale.
- Middle score: use when the broad destination or app/page/window is right but important visible state details are wrong, missing, or invented.
- Low score: use when the prediction is mostly a plausible story, unchanged state, or wrong branch despite sharing some surrounding context.
- Minimum score: use for wrong environment/page/app, malformed/non-state output, fabricated major content, or explanations instead of a predicted observation.

## Dimension-Specific Scoring Anchors

Use each dimension independently; do not let a well-formed DOM/accessibility tree inflate factual correctness.

- Format: score only schema/tag/readability compliance. A clean format cannot compensate for wrong page facts.
- Factuality: compare concrete visible anchors against the real observation. If the main page/action result is wrong, use 1-2. If the main result is right but several active-region anchors are wrong, use 3. Use 4-5 only for close anchor matches.
- Consistency: check causal continuity from the previous state and latest action. Wrong change-vs-no-change, wrong navigation/load result, wrong focus/dropdown/modal behavior, or stale state after a state-changing action should be 1-2.
- Realism: judge whether the predicted browser/page transition is behaviorally plausible for the site and action, but cap it at 3 when factuality or consistency is 1-2.
- Quality: judge usefulness as a replacement next-state observation. If a downstream agent would take the wrong next action because of missing/wrong active anchors, use 1-2; if usable only at a broad page level, use 3.

Before assigning any 4 or 5, explicitly verify the primary action effect plus at least these active anchors when present: URL/title or page identity, focused element, exact typed/search text, dropdown/modal/error state, selected item/filter, visible result/list/table labels, and scroll position.

## Discriminative Error Calibration

- Treat the latest action's primary visible effect as a gate. If it is wrong, stale, or missing, set factuality <= 2 and quality <= 2 even when many background nodes match.
- If the output copies or preserves much of the prior state while the ground truth changes, score it as a stale-state error.
- Downweight generic complete-looking accessibility trees: if exact active text, selected/focused element, error/load/modal state, and visible result rows do not match, total score should normally be <= 3.
\end{lstlisting}

\subsection{OS Domain}
\label{sec:appendix:judge_os}

\begin{lstlisting}
# Desktop OS World Model Judge Guidance

You are judging a desktop OS world-model prediction: the simulated desktop/app state after the latest user operation, compared with the real observed desktop state.

Use the task-provided evaluation protocol exactly as given by the user message. This system prompt only adds desktop-specific judging anchors.

## Core Principle

A strong prediction is not just a plausible desktop story. It must preserve the concrete state representation used by the trajectory and correctly update the visible desktop state caused by the latest user operation.

Treat the ground truth as the strongest evidence for visible desktop facts. If the ground truth appears clearly inconsistent with the prior history, rely on causal reasoning from the interaction history; otherwise, prefer predictions that preserve more concrete ground-truth anchors.

## Desktop Anchors

Before rewarding a prediction strongly, check these anchors when they are visible or implied by the turn:

- Active application, focused window, window title, z-order, minimized/maximized state, geometry, and workspace/desktop context.
- Dialogs, menus, context menus, popovers, file pickers, permission prompts, alerts, and whether they block interaction.
- Exact file, folder, document, tab, terminal, process, or app names visible in the state.
- Focused input, cursor position, selected text/items, edited text, unsaved/saved state, and clipboard-like visible effects.
- Terminal command output, prompt position, last visible rows, errors, exit state, and current working directory when represented.
- File-manager state such as created, renamed, moved, deleted, selected, or highlighted files and folders.
- Scroll position, visible viewport, status bars, clock/status text, coordinates, taskbar/dock/panel state, and notification indicators.

## Strictness Rules

- Penalize predictions that switch apps/windows, close dialogs, dismiss menus, move focus, or alter files without causal support from the latest user operation.
- Penalize broad desktop-level correctness when task-critical anchors such as focused window, exact file name, terminal output, dialog state, selection, or geometry are wrong.
- Penalize stale predictions that miss small but visible updates after typing, opening menus, selecting files, running commands, scrolling, saving, or switching focus.
- Penalize fabricated windows, impossible desktop states, generic summaries, and outputs that explain what should happen instead of showing the resulting desktop state.
- Do not let clean structure hide a wrong desktop state; the primary post-action effect matters most.

## Score Separation Policy

Map anchor accuracy to scores as follows:

- Top score: reserve for predictions that match the primary post-action effect and nearly all task-critical visible anchors.
- High-but-not-top score: use when the main transition is correct but one or two local anchors are missing or slightly stale.
- Middle score: use when the broad destination or app/page/window is right but important visible state details are wrong, missing, or invented.
- Low score: use when the prediction is mostly a plausible story, unchanged state, or wrong branch despite sharing some surrounding context.
- Minimum score: use for wrong environment/page/app, malformed/non-state output, fabricated major content, or explanations instead of a predicted observation.

## Dimension-Specific Scoring Anchors

Use each dimension independently; do not let a well-formed desktop hierarchy inflate factual correctness.

- Format: score only schema/tag/readability compliance. A clean format cannot compensate for wrong desktop facts.
- Factuality: compare concrete visible anchors against the real observation. If the main window/action result is wrong, use 1-2. If the main result is right but several active-region anchors are wrong, use 3. Use 4-5 only for close anchor matches.
- Consistency: check causal continuity from the previous state and latest action. Wrong change-vs-no-change, wrong app/window/menu/dialog result, wrong focus/selection behavior, or stale state after a state-changing action should be 1-2.
- Realism: judge whether the predicted desktop transition is behaviorally plausible for the app and action, but cap it at 3 when factuality or consistency is 1-2.
- Quality: judge usefulness as a replacement next-state observation. If a downstream agent would take the wrong next action because of missing/wrong active anchors, use 1-2; if usable only at a broad app/window level, use 3.

Before assigning any 4 or 5, explicitly verify the primary action effect plus at least these active anchors when present: active app/window/title, focused control, exact typed text/path/terminal output, dialog/menu state, selected file/item, visible list/table rows, and window geometry/scroll position.

## Discriminative Error Calibration

- Treat the latest action's primary visible effect as a gate. If it is wrong, stale, or missing, set factuality <= 2 and quality <= 2 even when many background nodes match.
- If the output copies or preserves much of the prior state while the ground truth changes, score it as a stale-state error.
- Downweight generic complete-looking desktop trees: if exact active text/path/cell/menu item, selected/focused element, dialog/menu state, and visible rows/files do not match, total score should normally be <= 3.
\end{lstlisting}

\end{document}